\documentclass[10pt,journal,compsoc]{IEEEtran}
\ifCLASSOPTIONcompsoc
  \usepackage[nocompress]{cite}
\else
  \usepackage{cite}
\fi
\ifCLASSINFOpdf
  \usepackage[pdftex]{graphicx}
\else
  \usepackage[dvips]{graphicx}
\fi
\usepackage{amsmath}
\usepackage{amssymb}
\usepackage{amsthm}
\usepackage{color}
\theoremstyle{plain}
\newtheorem{Theorem}{Theorem}
\newtheorem{Lemma}{Lemma}

\usepackage{array}
\usepackage{multirow}
\ifCLASSOPTIONcompsoc
  \usepackage[caption=false,font=footnotesize,labelfont=sf,textfont=sf]{subfig}
\else
  \usepackage[caption=false,font=footnotesize]{subfig}
\fi
\begin{document}
\title{A Theoretical View of Linear Backpropagation and Its Convergence}

\author{Ziang~Li*,
        Yiwen~Guo*,
        Haodi~Liu,
        and~Changshui~Zhang,~\IEEEmembership{Fellow,~IEEE}% <-this % stops a space
\IEEEcompsocitemizethanks{
\IEEEcompsocthanksitem Z. Li, H. Liu, and C. Zhang are with the Institute for Artificial Intelligence, Tsinghua University (THUAI), State Key Lab of Intelligent Technologies and Systems, Beijing National Research Center for Information Science and Technology (BNRist), Department of Automation, Tsinghua University, Beijing 100084, China.\protect\\
E-mail: liza19@mails.tsinghua.edu.cn, liuhd21@mails.tsinghua.edu.cn, zcs@mail.tsinghua.edu.cn.
\IEEEcompsocthanksitem Y. Guo is an independent researcher, Beijing 100000, China.\protect\\
E-mail: guoyiwen89@gmail.com. 
\IEEEcompsocthanksitem *These two authors contribute equally. 
\IEEEcompsocthanksitem Correspondence to: Y. Guo and C. Zhang.}
\thanks{}}

\IEEEtitleabstractindextext{%
\begin{abstract}
Backpropagation (BP) is widely used for calculating gradients in deep neural networks (DNNs). Applied often along with stochastic gradient descent (SGD) or its variants, BP is considered as a de-facto choice in a variety of machine learning tasks including DNN training and adversarial attack/defense. Recently, a linear variant of BP named LinBP was introduced for generating more transferable adversarial examples for performing black-box attacks, by Guo \emph{et al.}~\cite{guo2020backpropagating}. Although it has been shown empirically effective in black-box attacks, theoretical studies and convergence analyses of such a method is lacking. This paper serves as a complement and somewhat an extension to Guo \emph{et al.}'s paper, by providing theoretical analyses on LinBP in neural-network-involved learning tasks, including adversarial attack and model training. We demonstrate that, somewhat surprisingly, LinBP can lead to faster convergence in these tasks in the same hyper-parameter settings, compared to BP. We confirm our theoretical results with extensive experiments. Code for reproducing our experimental results is available at https://github.com/lzalza/LinBP.
\end{abstract}

}

\maketitle

\IEEEdisplaynontitleabstractindextext

% \fi
\IEEEpeerreviewmaketitle

\IEEEraisesectionheading{\section{Introduction}\label{sec:introduction}}

\IEEEPARstart{O}{ver} the past decade, the surge of research on deep neural networks (DNNs) has been witnessed. Powered with large-scale training, DNN-based models have achieved state-of-art performance in a variety of real-world applications. Tremendous amount of research has been conducted to improve the architecture~\cite{simonyan2015very,he2016deep,huang2017densely,vaswani2017attention,dosovitskiy2020image} and the optimization~\cite{kingma2014adam,loshchilov2018decoupled,loshchilov2016sgdr,reddi2019convergence,sutskever2013importance} of DNNs. 

The optimization of DNNs often involves stochastic gradient descent (SGD)~\cite{bottou2010large} (that minimizes some training loss) and backpropagation (BP)~\cite{lecun1988theoretical} (for computing gradients of the loss function). 
In the work of Guo \emph{et al.} published at NeurIPS 2020~\cite{guo2020backpropagating}, a different way of computing ``gradients'' was proposed. The method, \emph{i.e.}, linear BP (LinBP), keeps the forward pass of BP unchanged while skips the partial derivative regarding some of the non-linear activations in the backward pass. It was shown empirically that LinBP led to improved results in generating transferable adversarial examples for performing \emph{black-box attacks}~\cite{papernot2017practical}, in comparison with using the original BP. This paper was written as a complement and somewhat an extension to Guo \emph{et al.}'s paper, in a way that we delve deep into the theoretical properties of LinBP. Moreover, we target two more applications of LinBP, \emph{i.e.}, \emph{white-box attack} and model parameter training, to study the optimization convergence of learning with LinBP.

In this paper, 
%we follow prior work~\cite{tian2017analytical,li2017convergence} and perform study in a teacher-student framework which uses ReLU activation functions and squared $l_2$ loss. 
we will provide convergence analyses for LinBP, and we will demonstrate that, in white-box adversarial attack scenarios, using LinBP can produce more deceptive adversarial examples (similar to the results in black-box scenarios as given in~\cite{guo2020backpropagating}), 
than using the standard BP in the same hyper-parameter settings. Plausible explanations of such fast convergence will be given. Similarly, we will also show theoretically that LinBP helps converge faster in training DNN models, if the same hyper-parameters are used for the standard BP and LinBP. Simulation experiments confirm our theoretical results, and extensive results will verify our findings in more general and practical settings using a variety of widely used DNNs, including VGG-16~\cite{simonyan2015very}, ResNet-50~\cite{he2016deep}, DenseNet-161~\cite{huang2017densely}, MobileNetV2~\cite{sandler2018mobilenetv2}, and WideResNet (WRN)~\cite{zagoruyko2016wide}.

The rest of this paper is organized as follows. In Section~\ref{sec:background}, we will revisit preliminary knowledge about network training and adversarial examples, to introduce LinBP and discuss its applications. In Section~\ref{sec:TA}, we will provide theoretical analyses on the convergence of LinBP in both the white-box attack and model training scenarios. In Section~\ref{sec:exp}, we will conduct extensive simulation and practical experiments to validate our theoretical findings and test LinBP with different DNNs. In Section~\ref{sec:conclusion}, we draw conclusions.

\section{Background and Preliminary Knowledge}\label{sec:background}
\subsection{Model Training}
Together with SGD or its variant, BP has long been adopted as a default method for computing gradients in training machine learning models. Consider a typical update rule of SGD, we have 
\begin{equation}
\label{SGD}
\mathbf{w}_{t+1} = \mathbf{w}_{t} - \eta \nabla_{\mathbf{w}_t}\sum_k\mathcal{L}(\mathbf{x}^{(k)}, y^{(k)}),
\end{equation}
where $\{\mathbf{x}^{(k)}, y^{(k)}\}$ contains the $k$-th training instance and its label, $\nabla_{\mathbf{w}}\sum_k\mathcal{L}(\mathbf{x}^{(k)}, y^{(k)})$ is the partial gradient of the loss function with respect to a learnable weight vector $\mathbf{w}$, 
and $\eta\in \mathbb R^{+}$ represents the learning rate. 

\subsection{White-box Adversarial Attack}
Another popular optimization task in deep learning is to generate adversarial examples, which is of particular interest in both the machine learning community and the security community. Instead of minimizing the prediction loss as in the objective of model training, this task aims to craft inputs that lead to arbitrary incorrect model predictions~\cite{goodfellow2015explaining}. The adversarial examples are expected to be perceptually indistinguishable from the benign ones. Adversarial examples can be untargeted or targeted, and in this paper we focus on the former. In the white-box setting, where it is assumed that the architecture and parameters of the victim model are both known to the adversary, we have the typical learning objectives for generating adversarial examples: 
\begin{equation}
\label{white-box-a}
\max_{\|\mathbf{r}\|_p\leq \epsilon}\mathcal{L}(\mathbf{x}+\mathbf{r},\mathbf{y})
\end{equation}
and
\begin{equation}
\label{white-box-b}
\begin{aligned}
\min  {\|\mathbf r\|_p},\  
\mathrm{s.t.,} \ \ \arg\max_j \text{prob}_j(\mathbf x+\mathbf r) \neq \arg\max_j \text{prob}_j(\mathbf x)
\end{aligned}
\end{equation}
where $\mathbf r$ is a perturbation vector whose $l_p$ norm is constrained (\emph{e.g.,} ${\|\mathbf r\|_p}<\epsilon$) to guarantee that the generated adversarial example $\mathbf x+\mathbf r$ is perceptually similar to the benign example $\mathbf x$, function $\text{prob}_j(\cdot)$ provides the prediction probability for the $j$-th class. For solving the optimization problems in Eq.~(\ref{white-box-a}) and~(\ref{white-box-b}), a series of methods have been proposed over the last decade. For instance, with $p=\infty$,~\emph{i.e.}, for performing $l_\infty$ attack, Goodfellow \emph{et al.}\cite{goodfellow2015explaining} proposed the fast gradient sign method (FGSM) which simply calculates the sign of the input gradient,~\emph{i.e.,} $\mathrm{sign}(\nabla_{\mathbf x} L(\mathbf x))$, and adopted $\mathbf r = \epsilon\cdot \mathrm{sign}(\nabla_{\mathbf x} L(\mathbf x))$ as the perturbation. To enhance the power of the attack, follow-up methods proposed iterative FGSM (I-FGSM)~\cite{kurakin2017adversarial} and PGD~\cite{madry2018deep} that took multiple steps of update and computed input gradients using BP in each of the steps. The iterative process of these methods is similar to that of model training, except that the update is performed in the input space instead of the parameter space, and normally some constraints are required for attacks. In transfer-based attacks, I-FGSM is widely adopted as a baseline method while it has been reported that PGD is capable of bypassing gradient masking~\cite{athalye2018obfuscated}. Other famous attacks include DeepFool~\cite{moosavi2016deepfool} and C\&W's attack~\cite{carlini2017towards}, just to name a few.

\subsection{Linear Backpropagation}
Inspired by the hypothesis made by Goodfellow \emph{et al.}~\cite{goodfellow2015explaining} that the linear nature of modern DNNs causes the transferability of adversarial samples, LinBP, a method improves the linearity of DNNs, was proposed. For a $d$-layer DNN model, the forward pass includes the computation of
\begin{equation}
\label{forward}
f_d(\mathbf{x}) = \mathbf{W}^T_{d}\sigma\left(\mathbf{W}^T_{d-1}\cdots\sigma(\mathbf{W}^T_1\mathbf{x})\right),
\end{equation}
where $\sigma(\cdot)$ is the activation function and it is often set as the ReLU function, $\mathbf{W}_1, \cdots ,\mathbf{W}_d$ are learnable weight matrices in the DNN model. In order to generate a white-box adversarial example on the basis of $\mathbf x$ with $f_d(\mathbf x)$, we shall further compute the prediction loss $\mathcal{L}(\mathbf{x},\mathbf{y})$ for $\mathbf x$ and then backpropagate the loss to compute the input gradient  $\nabla_{\mathbf{x}}\mathcal{L}(\mathbf{x},\mathbf{y})$, 
the default BP for computing the input gradient is formulated as
\begin{equation}
\label{backward}
\nabla_{\mathbf{x}}\mathcal{L}(\mathbf{x},\mathbf{y})=\mathbf{M}_1\cdots\mathbf{M}_{d-1}\mathbf{W_d}\frac{\partial \mathcal{L}(\mathbf{x},\mathbf{y})}{\partial f_d(\mathbf{x})},
\end{equation}
where $\mathbf{M}_i=\mathbf{W}_i\partial \sigma(f_i(\mathbf{x}))/\partial f_i(\mathbf{x})$. LinBP keeps the computation in the forward pass unchanged while removing the influence of the ReLU activation function (\emph{i.e.}, $\partial \sigma(f_i(\mathbf{x}))/\partial f_i(\mathbf{x})$) in the backward pass. That being said, LinBP computes:
\begin{equation}
\label{linbp_ori}
\tilde{\nabla}_{\mathbf{x}}\mathcal{L}(\mathbf{x},\mathbf{y})=\mathbf{W}_1\cdots\mathbf{W}_d\frac{\partial \mathcal{L}(\mathbf{x},\mathbf{y})}{\partial f_d(\mathbf{x})}.
\end{equation}
Since the partial derivative of the activation function has been removed, it is considered linear in the backward pass and thus called linear BP (LinBP).
In practice, LinBP may only remove some of the nonlinear derivatives,~\emph{e.g.,} only modifies those starting from the ($m+1$)-th layer and uses the following formulation:
\begin{equation}
\label{linbp}
\tilde{\nabla}_{\mathbf{x}}\mathcal{L}(\mathbf{x},\mathbf{y})=\mathbf{M}_1\cdots\mathbf{M}_m\mathbf{W}_{m+1}\cdots\mathbf{W}_{d}\frac{\partial \mathcal{L}(\mathbf{x},\mathbf{y})}{\partial f_d(\mathbf{x})}.
\end{equation}
Experimental results on CIFAR-10~\cite{krizhevsky2009learning} and ImageNet~\cite{russakovsky2015imagenet} tested the transferability of generated adversarial examples on different source models and found that LinBP achieved state-of-the-arts~\cite{guo2020backpropagating}.

In addition to generating adversarial examples, we can also adopt LinBP to compute the gradient with respect to model weights for training DNNs. Note that the gradient of the loss with respect to $\mathbf{W}_i$ in the standard BP is formulated as
\begin{equation}
\label{backward_w}
\begin{aligned}
\nabla_{\mathbf{W}_i}\mathcal{L}(\mathbf{x},\mathbf{y})=&\sigma\left(f_{i-1}(\mathbf{x})\right)\left(\frac{\partial \sigma(f_i(\mathbf{x}))}{\partial f_i(\mathbf{x})}\mathbf{M}_{i+1}\cdots\right.\\&\left.\mathbf{M}_{d-1}\mathbf{W}_d\frac{\partial\mathcal{L}(\mathbf{x},\mathbf{y})}{\partial f_d(\mathbf{x})}\right)^T,
\end{aligned}
\end{equation}
where we abuse the notations and define $\sigma(f_0(\mathbf{x}))=\mathbf{x}$ for simplicity. Just like in Eq.~(\ref{linbp_ori}), LinBP computes a linearized version of the gradient, which is formulated as 
\begin{equation}
\label{backward_w_linbp}
\tilde{\nabla}_{\mathbf{W}_i}\mathcal{L}(\mathbf{x},\mathbf{y})=\sigma\left(f_{i-1}(\mathbf{x})\right)\left(\mathbf{W}_{i+1}\cdots\mathbf{W}_d\frac{\partial\mathcal{L}(\mathbf{x},\mathbf{y})}{\partial f_d(\mathbf{x})}\right)^T.
\end{equation}

\section{Theoretical Analyses}\label{sec:TA}
If we consider skipping ReLU in the backward pass as finding an approximation to the gradient of a non-smooth objective function, then LinBP is somewhat related to the straight-through estimation~\cite{bengio2013estimating}. However, such a possible relation does not provide much insight of the convergence of LinBP, which is the focus of this paper.
In this section, we shall provide convergence analyses for LinBP, and compare it to the standard BP. There have been attempts for studying the training dynamics of neural networks~\cite{du2019gradient,arora2019fine,tian2017analytical,du2019provably}. We follow Tian's work~\cite{tian2017analytical} and consider a teacher-student framework with the ReLU activation function and squared $l_2$ loss, based on no deeper than two-layer networks. We shall start from theoretical analyses for white-box adversarial attacks and then consider model training.
\subsection{Theoretical Analyses for Adversarial Attack}\label{sec:theo_attack}
Compared to the two-layer teacher-student frameworks in Tian's work~\cite{tian2017analytical}, we here consider a more general model, which can be formulated as
\begin{equation}
\label{attack_forward}
g(\mathbf{W},\mathbf{V},\mathbf{x}) = \mathbf{V}\sigma(\mathbf{W}\mathbf{x}),
\end{equation}
where $\mathbf{x}\in\mathbb{R}^{d_1}$ is the input data vector, $\mathbf{W}\in\mathbb{R}^{d_2 \times d_1}$ and $\mathbf{V}\in\mathbb{R}^{d_3 \times d_2}$ are weight matrices, $\sigma(\cdot)$ is the ReLU function which compares its inputs to $0$ in an element-wise manner. In the teacher-student frameworks, we assume the teacher network possesses the optimal adversarial example $\mathbf{x}^\star$, and the student network aims to learn an adversarial example $\mathbf{x}$ from the teacher network, in which the loss function is set as the squared $l_2$ loss between the output of the student and the teacher network, \emph{i.e.},
\begin{equation}
\label{loss_attack}
\mathcal{L}(\mathbf{x}) = \frac{1}{2}\|g(\mathbf{W},\mathbf{V},\mathbf{x})-g(\mathbf{W},\mathbf{V},\mathbf{x}^\star)\|_2^2.
\end{equation}
We further define $D(\mathbf{W},\mathbf{x}) := {\rm diag} (\mathbf{Wx} > 0)$. From Eq.~(\ref{attack_forward}) and Eq.~(\ref{loss_attack}), we can easily obtain the analytic expression of the gradient with respect to $\mathbf{x}$ for the standard BP:
\begin{equation}
\label{gradient_x}
\begin{aligned}
\nabla_{\mathbf{x}}\mathcal{L}(\mathbf{x}) = &\mathbf{W}^TD(\mathbf{W},\mathbf{x})\mathbf{V}^T\mathbf{V}(D(\mathbf{W},\mathbf{x})\mathbf{Wx} \\&- D(\mathbf{W},\mathbf{x}^\star) \mathbf{W} \mathbf{x}^{\star}).
\end{aligned}
\end{equation}

Compared with Eq.~(\ref{gradient_x}), the gradient obtained from LinBP removes the derivatives of the ReLU function in the backward pass, \emph{i.e.,}
\begin{equation}
\label{gradient_x2}
\tilde{\nabla}_{\mathbf{x}}\mathcal{L}(\mathbf{x}) = \mathbf{W}^T\mathbf{V}^T\mathbf{V}\left(D(\mathbf{W},\mathbf{x})\mathbf{Wx} - D(\mathbf{W},\mathbf{x}^\star) \mathbf{W} \mathbf{x}^{\star}\right).
\end{equation}
Inspired by Theorem 1 in Tian's work~\cite{tian2017analytical}, we introduce the following lemma to give the analytic expression of the gradients in expectations in adversarial settings.
\begin{Lemma}
	\label{lemma1}
	$G(\mathbf{e},\mathbf{x}) := \mathbf{W}^TD(\mathbf{W},\mathbf{e})\mathbf{V}^T\mathbf{V}D(\mathbf{W},\mathbf{x})\mathbf{Wx}$, where $\mathbf{e} \in \mathbb{R}^{d_1}$ is a unit vector, $\mathbf{x} \in \mathbb{R}^{d_1}$ is the input data vector, $\mathbf{W}\in\mathbb{R}^{d_2 \times d_1}$ and $\mathbf{V}\in\mathbb{R}^{d_3 \times d_2}$ are weight matrices. If $\mathbf{W}$ and $\mathbf{V}$ are independent and both generated from the standard Gaussian distribution, we have $$\mathbb{E}\left(G(\mathbf{e},\mathbf{x})\right)=\frac{d_3d_2}{2\pi}[(\pi-\Theta)\mathbf{x}+\|\mathbf{x}\|\sin\Theta \mathbf{e}],$$ where $\Theta \in [0,\pi]$ is the angle between $\mathbf{e}$ and $\mathbf{x}$.
\end{Lemma}
The proof of Lemma~\ref{lemma1} can be found in the appendices. With Lemma~\ref{lemma1}, the expectation of Eq.~(\ref{gradient_x}) and Eq.~(\ref{gradient_x2}) can be formulated as
\begin{equation}
\label{ex4_main}
\begin{aligned}
\mathbb{E}[\nabla_{\mathbf{x}}\mathcal{L}(\mathbf{x})] &= G(\mathbf{x}/\|\mathbf{x}\|,\mathbf{x}) - G(\mathbf{x}/\|\mathbf{x}\|,\mathbf{x}^\star)\\
&=\frac{d_3d_2}{2}(\mathbf{x}-\mathbf{x}^\star)+\frac{d_3d_2}{2\pi}\left(\Theta \mathbf{x}^\star-\frac{\|\mathbf{x}^\star\|}{\|\mathbf{x}\|}\sin\Theta \mathbf{x}\right)
\end{aligned}
\end{equation}
and
\begin{equation}
\label{ex3_main}
\begin{aligned}
\mathbb{E}[\tilde{\nabla}_{\mathbf{x}}\mathcal{L}(\mathbf{x})] &= G(\mathbf{x}/\|\mathbf{x}\|,\mathbf{x}) - G(\mathbf{x}^\star/\|\mathbf{x}^\star\|,\mathbf{x}^\star)\\
&=\frac{d_3d_2}{2}(\mathbf{x}-\mathbf{x}^\star),
\end{aligned}
\end{equation}
respectively, where $\Theta \in [0,\pi]$ is the angle between $\mathbf{x}$ and $\mathbf{x^\star}$.
We here focus on $l_\infty$ attacks. Different iterative gradient-based $l_\infty$ attack methods may have slightly different update rules.  Here, for ease of unified analyses, we simplify their update rules as
\begin{equation}
\label{update_x}
\mathbf{x}^{(t+1)}=\text{Clip}(\mathbf{x}^{(t)} - \eta \nabla_{\mathbf{x}^{(t)}}\mathcal{L}(\mathbf{x}^{(t)})),
\end{equation}
where $\text{Clip}(\cdot) = \min(\mathbf x + \epsilon\mathbf 1,\max(\mathbf x - \epsilon\mathbf 1,\cdot))$ performs element-wise input clip to guarantee that the intermediate results always stay in the range fulfilling the constraint of $\|\mathbf x^{(t+1)} - \mathbf x \|_\infty\leq \epsilon$. We use the updates of the standard BP and LinBP, \emph{i.e.}, $\nabla_{\mathbf{x}}\mathcal{L}(\mathbf{x})$ and $\tilde{\nabla}_{\mathbf{x}}\mathcal{L}(\mathbf{x})$, to obtain $\{\mathbf{x}^{(t)}\}$ and $\{\tilde{\mathbf{x}}^{(t)}\}$, respectively. Based on all these results above, we can give the following theorem that provides convergence analysis of LinBP.
\begin{Theorem}
	\label{theorem1}
	For the two-layer teacher-student network formulated as in Eq.~(\ref{attack_forward}), in which the adversarial attack adopts Eq.~(\ref{loss_attack}) and Eq.~(\ref{update_x}) as the loss function and the update rule, respectively, we assume that $\mathbf{W}$ and $\mathbf{V}$ are independent and both generated from the standard Gaussian distribution, $\mathbf{x}^\star \sim N(\mu_1,\sigma_1^2)$, $\mathbf{x}^{(0)} \sim N(0,\sigma_2^2)$, and $\eta$ is reasonably small~\footnote{See Eq.~(\ref{eq:constraint_2}) for more details of the constraint.}. Let $\mathbf{x}^{(t)}$ and $\tilde{\mathbf{x}}^{(t)}$ be the adversarial examples generated in the $t$-th iteration of attack using BP and LinBP, respectively, then we have $$\mathbb{E}\|\mathbf{x}^\star - \tilde{\mathbf{x}}^{(t)}\|_1 \leq \mathbb{E}\|\mathbf{x}^\star - \mathbf{x}^{(t)}\|_1.$$
\end{Theorem}
The proof of the theorem is deferred to the appendices. Theorem~\ref{theorem1} shows that LinBP can produce adversarial examples closer to the optimal adversarial examples $\mathbf{x}^\star$ (when compared with the standard BP) in the same settings for any finite number of iteration steps, which means that LinBP can craft more powerful and destructive adversarial examples even in the white-box attacks. It is also straightforward to derive from our proof that LinBP produces a more powerful adversarial example, if starting optimization from the benign example with low prediction loss. The conclusion is somewhat surprising since popular white-box attack methods like FGSM (I-FGSM) and PGD mostly use the gradient obtained by the standard BP, yet we find LinBP may lead to stronger attacks with the same hyper-parameters. 

From our proof, it can further be derived that the norm of Eq.~(\ref{ex3_main}) is larger than the norm of Eq.~(\ref{ex4_main}), which means LinBP provides larger gradients in expectation. Also, the direction of the update obtained from LinBP is closer to the residual $\mathbf{x} - \mathbf{x}^\star$ in comparison to the standard BP. These may cause LinBP to produce more destructive adversarial examples and more powerful attack in white-box settings.
We conducted extensive experiments on deeper networks using common attack methods to verify our theoretical findings. 
The experimental results will be shown in Section~\ref{sec:exp_real}.

\subsection{Theoretical Analyses for Model Training}\label{sec:theo_train}
For model training, we adopt the same framework as in Section~\ref{sec:theo_attack} to analyze the performance of LinBP. By contrast, we mainly consider a one-layer teacher-student framework, since the student network may not converge to the teacher network with two-layer and deeper models. 
We will show in Section~\ref{sec:exp_real} that many of our theoretical results still hold in more complex network architectures. 

The one-layer network can be formulated as
\begin{equation}
\label{ori}
h(\mathbf{x},\mathbf{w}) = \sigma(\mathbf{x}^T\mathbf{w}),
\end{equation}
where $\mathbf{x}\in\mathbb{R}^d$ is the input vector, $\mathbf{w}\in\mathbb{R}^d$ is the weight vector, and $\sigma(\cdot)$ is the ReLU function. Given a set of training samples, we obtain $h(\mathbf X, \mathbf w)=\sigma(\mathbf X\mathbf w)$, where $\mathbf{X} = [\mathbf{x}_1^T;...;\mathbf{x}_N^T]$ is the input data matrix, $\mathbf{x}_k\in\mathbb{R}^d$ is the $k$-th training instance, for $k=1,...,N$. We assume the teacher network 
has the optimal weight, and the loss function for optimizing the one-layer network can be formulated as 
\begin{equation}
\label{loss}
\mathcal{L}(\mathbf{w}) = \frac{1}{2}\|h(\mathbf{X},\mathbf{w})-h(\mathbf{X},\mathbf{w}^\star)\|_2^2,
\end{equation}
where $\mathbf{w}$ and $\mathbf{w}^\star$ are the weight vectors for the student network and the teacher network, respectively. Therefore, we can derive the gradient with respect to $\mathbf{w}$ for the standard BP as:
\begin{equation}
\label{gradient}
\nabla_{\mathbf{w}}\mathcal{L}(\mathbf{w}) = \mathbf{X}^T\mathbf{D}(\mathbf{X},\mathbf{w})(\mathbf{D}(\mathbf{X},\mathbf{w})\mathbf{Xw} - \mathbf{D}(\mathbf{X},\mathbf{w}^\star) \mathbf{X} \mathbf{w}^{\star}).
\end{equation}
By contrast, the gradient with respect to $\mathbf{w}$ obtained from LinBP is formulated as
\begin{equation}
\label{gradient2}
\tilde{\nabla}_{\mathbf{w}}\mathcal{L}(\mathbf{w}) = \mathbf{X}^T(\mathbf{D}(\mathbf{X},\mathbf{w})\mathbf{Xw} - \mathbf{D}(\mathbf{X},\mathbf{w}^\star) \mathbf{X} \mathbf{w}^{\star}).
\end{equation}
While training the simple one-layer network with SGD, the update rule can be formulated as 
\begin{equation}
\label{update}
\mathbf{w}^{(t+1)}=\mathbf{w}^{(t)} - \eta \nabla_{\mathbf{w}^{(t)}}\mathcal{L}(\mathbf{w}^{(t)}),
\end{equation}
where we use $\nabla_{\mathbf{w}}\mathcal{L}(\mathbf{w})$ and $\tilde{\nabla}_{\mathbf{w}}\mathcal{L}(\mathbf{w})$ showed in Eq.~(\ref{gradient}) and Eq.~(\ref{gradient2}) to obtain $\{\mathbf{w}^{(t)}\}$ and $\{\tilde{\mathbf{w}}^{(t)}\}$ for the standard BP and LinBP, respectively. The following theorem is given for analyzing the training convergence, when LinBP is used. 
\begin{Theorem}
	\label{theorem2}
	For the one-layer teacher-student network formulated as in Eq.~(\ref{ori}), in which training adopts Eq.~(\ref{loss}) and Eq.~(\ref{update}) as the loss function and the update rule, respectively, we assume that $\mathbf{X}$ is generated from the standard Gaussian distribution, $\mathbf{w}^\star \sim N(\mu_1,\sigma_1^2)$, $\mathbf{w}^{(0)} \sim N(0,\sigma_2^2)$, and $\eta$ is reasonably small\footnote{See Eq.~(\ref{eq:constraint_1}) for a precious formulation of the constraint.}. 
	Let $\mathbf{w}^{(t)}$ and $\tilde{\mathbf{w}}^{(t)}$ be the weight vectors obtained in the $t$-th iteration of training using standard BP and LinBP, respectively, then we have $$\mathbb{E}\|\mathbf{w}^\star - \tilde{\mathbf{w}}^{(t)}\|_1 \leq \mathbb{E}\|\mathbf{w}^\star - \mathbf{w}^{(t)}\|_1.$$
\end{Theorem}
The proof of Theorem~\ref{theorem2} can also be found in the appendices. Theorem~\ref{theorem2} shows that LinBP may lead the obtained weight vector to get closer toward the optimal weight vector $\mathbf{w}^\star$ compared to the standard BP with the same learning rate and the same number of training iterations, which means LinBP can also lead to faster convergence in model training.

\subsection{Discussions about $\eta$ and $t$}
In Theorem~\ref{theorem1} and Theorem~\ref{theorem2}, we assume that the update step size in adversarial attacks and the learning rate in model training are reasonably small. Here we would like to discuss more about these assumptions. 

Precisely, we mean that $\eta$ should be small enough to satisfy the following constraints: for Theorem~\ref{theorem1}, it is required that
\begin{equation}
\label{eq:constraint_2}
\begin{aligned}
    |\sum\limits_{j=0}^{m-1}\frac{\eta d_3d_2}{2\pi}(1-&\frac{\eta d_3d_2}{2})^{m-1-j}\mathbf{p}_{ji}| \\
    &< |(1-\frac{\eta d_3d_2}{2})^{m}(\mathbf{x}_i^{\star}-\mathbf{x}^{(0)}_i)|,
\end{aligned}
\end{equation}
for $m=1,\ldots,t$, and $i=1,\ldots,d_1$, where $\mathbf{p}_j=\theta_j\mathbf{x}^\star-\frac{\|\mathbf{x}^\star\|}{\|\mathbf{x}^{(j)}\|}\sin\theta_j\mathbf{x}^{(j)}$ and $i$ indicates the $i$-th entry of a $d_1$-dimensional vector. For Theorem~\ref{theorem2}, it is required that
\begin{equation}
\label{eq:constraint_1}
|\sum\limits_{j=0}^{m-1}\frac{\eta N}{2\pi}(1-\frac{\eta N}{2})^{m-1-j}\mathbf{q}_{ji}| < |(1-\frac{\eta N}{2})^{m}(\mathbf{w}_i^{\star}-\mathbf{w}^{(0)}_i)|,
\end{equation}
for $m=1,\ldots,t$, and $i=1,\ldots,d$, where $\mathbf{q}_j=\theta_j\mathbf{w}^\star-\frac{\|\mathbf{w}^\star\|}{\|\mathbf{w}^{(j)}\|}\sin\theta_j\mathbf{w}^{(j)}$.
It is nontrivial to obtain analytic solutions for Eq.~(\ref{eq:constraint_2}) and Eq.~(\ref{eq:constraint_1}). However, we can know that they are both more likely to hold when $t$ is small. Since the maximum number of learning iterations for generating adversarial examples is often set to be small (at most several hundred) in commonly adopted methods like I-FGSM and PGD, the assumption in Eq.~\eqref{eq:constraint_2} is easier to be fulfilled than that in Eq.~\eqref{eq:constraint_1}, \emph{i.e.,} in the setting of model training which can take tens of thousands of iterations, indicating that the superiority of LinBP can be more obvious in performing adversarial attacks. %We will provide empirical discussions in Section~\ref{sec:exp}.

\section{Experimental Results}\label{sec:exp}
In this section, we provide experimental results on synthetic data (see Section~\ref{sec:exp_sim}) and real data (see Section~\ref{sec:exp_real}) to confirm our theorem and compare LinBP against BP in more practical settings for white-box adversarial attack/defense and model training, respectively. The practical experiments show that our theoretical results hold in a variety of different model architectures. All the experiments were performed on NVIDIA GeForce RTX 1080/2080 Ti and the code was implemented using PyTorch~\cite{paszke2019pytorch}.

\subsection{Simulation Experiments}
\label{sec:exp_sim}

\begin{figure}[htb]
	\centering
 	\vskip -0.0in
	\includegraphics[width=0.55\linewidth]{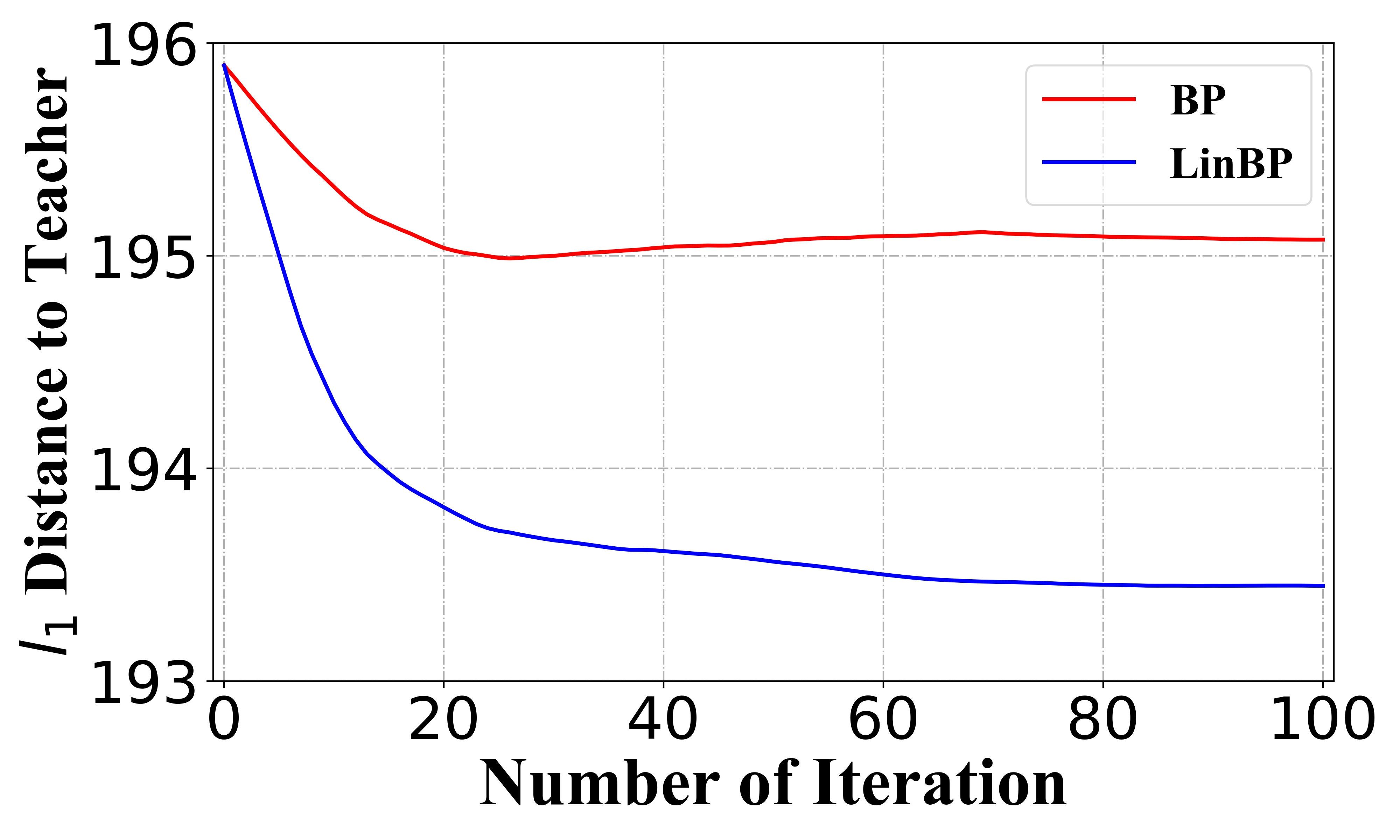}\vskip -0.1in
	\caption{LinBP leads to more powerful white-box adversarial examples which are closer to the optimal ones.}\label{fig:T2}
	\vskip -0.1in
\end{figure}
\begin{figure}[htb]
	\centering
	\subfloat[$l_2$ normalized updates]{\label{fig:attack_error2}	
		\includegraphics[width=0.45\linewidth]{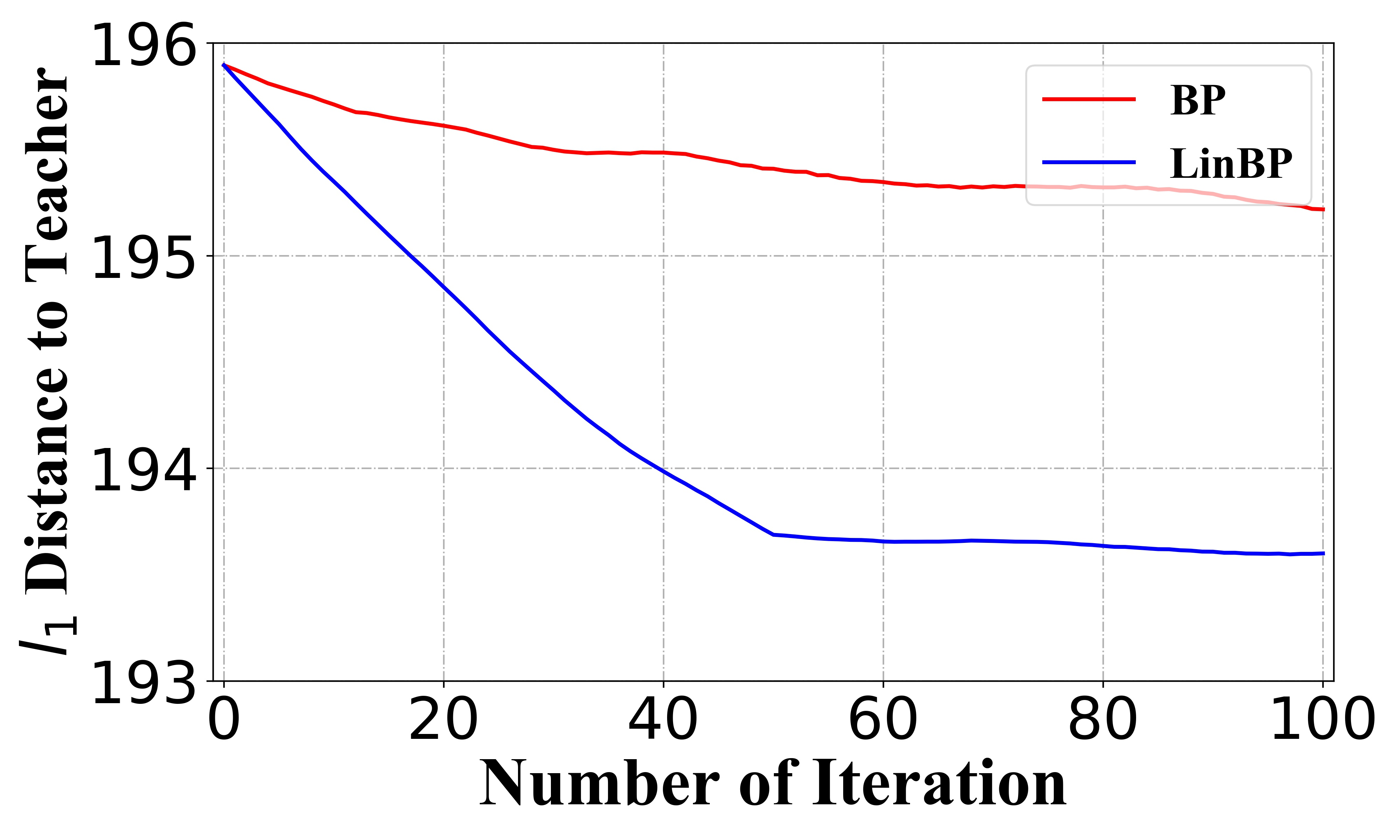}}
	\hskip 0.1in
	\subfloat[$l_\infty$ normalized updates]{\label{fig:attack_error3}	
		\includegraphics[width=0.45\linewidth]{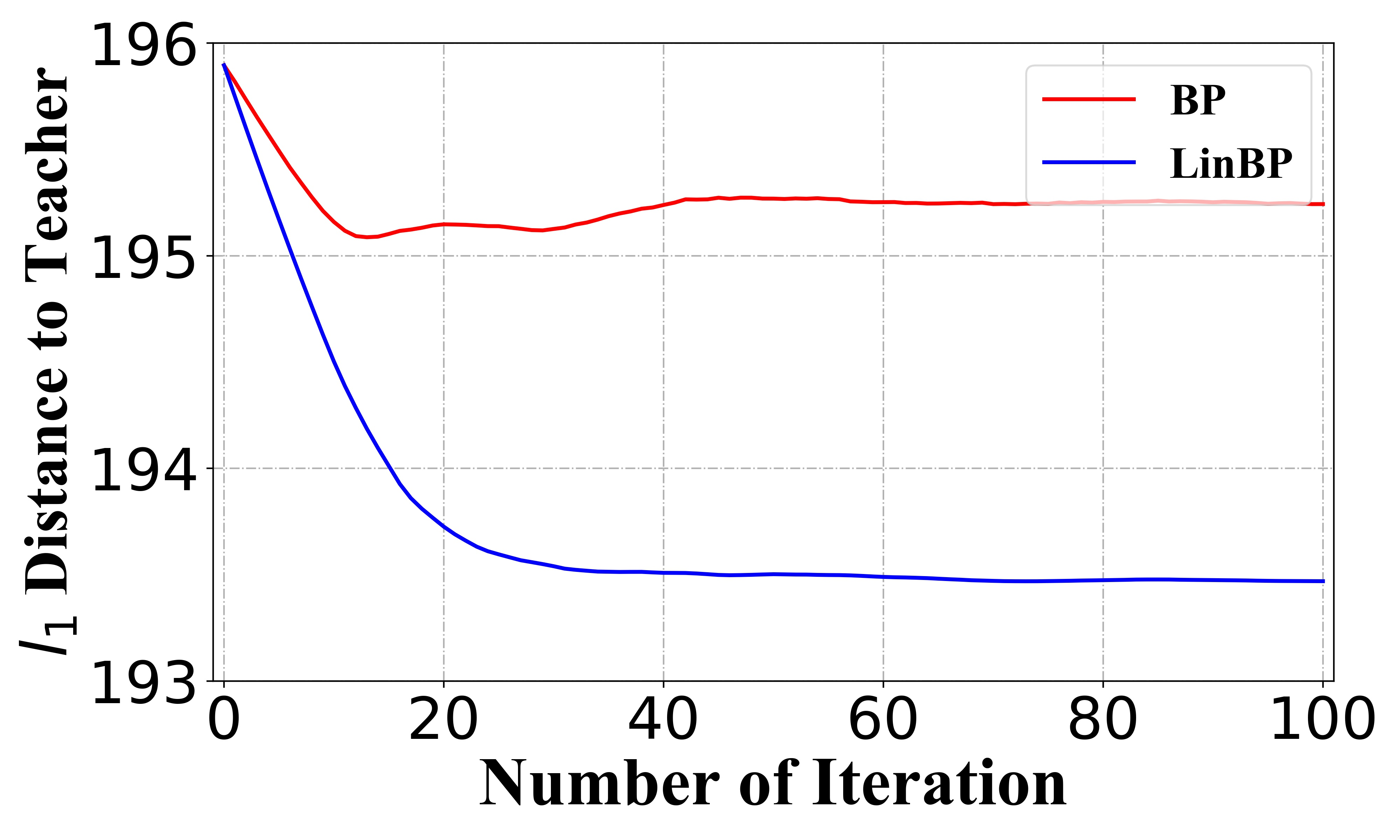}}
	\caption{Using normalized gradients, LinBP still leads to more powerful white-box adversarial examples.}\label{fig:T3}
	\vskip -0.1in
\end{figure}
\subsubsection{Adversarial attack} \label{sec:exp_attack}
We constructed a two-layer neural network and performed adversarial attack following the teacher-student framework described in Section~\ref{sec:theo_attack}. To be more specific, the victim model is formulated as Eq.~(\ref{attack_forward}), the loss function is given as Eq.~(\ref{loss_attack}), and Eq.~(\ref{update_x}) is the update rule to generate adversarial examples. Following the assumption in Theorem~\ref{theorem1}, the weight matrices $\mathbf{V}$ and $\mathbf{W}$ were independent and both generated from the standard Gaussian distribution. Similarly, the optimal adversarial example $\mathbf{x}^\star$ and the initial adversarial example $\mathbf{x}^{(0)}$ were obtained via sampling from Gaussian distributions, \emph{i.e.,} $\mathbf{x}^\star \sim N(\mu_1,\sigma_1^2)$ and $\mathbf{x}^{(0)} \sim N(0,\sigma_2^2)$, where $\mu_1$, $\sigma_1$, and $\sigma_2$ can in fact be arbitrary constants. Here we set $\mu_1 = 1.0$, $\sigma_1 = 2.0$, and $\sigma_2 = 1.0$. In our experiments, we set $d_1= 100$, $d_2 = 20$, $d_3 = 10$, $\eta=0.001$, and $\epsilon=0.25$. And the maximum number of the iteration step was set to $100$. We randomly sampled $10$ sets of parameters (including weight matrices, optimal adversarial examples, and initial adversarial examples) for different methods and evaluate the average performance over $10$ runs of the experiment. The average $l_1$ distance between the obtained student adversarial examples and the optimal adversarial examples from the teacher network is shown in Figure~\ref{fig:T2}. It shows that LinBP makes the obtained adversarial examples converge faster to the optimal ones, indicating that LinBP helps craft more powerful adversarial examples in the same hyper-parameter settings, which clearly confirms our Theorem~\ref{theorem1}. 

As we have mentioned in Section~\ref{sec:TA}, the norm of the update obtained from LinBP is larger than that obtained from the standard BP, which may cause faster convergence for LinBP. In this experiment, the $l_2$ norm and $l_\infty$ norm of the gradient obtained using LinBP are $\sim1.65\times$ and $\sim1.60\times$ larger than BP, respectively. For deeper discussions, we have conducted two more experiments. We used $l_2$ norm and $l_\infty$ norm to normalize the update obtained from LinBP and standard BP. $\eta$ was set as 0.05 and 0.005, respectively. The results are shown in Figure~\ref{fig:attack_error2} and Figure~\ref{fig:attack_error3}, where we can find the conclusion still holds with gradient normalization. %
\begin{figure}[htb]
	\centering \vskip -0.0in
	\subfloat[Training loss]{\label{fig:T1_loss}	
		\includegraphics[width=0.45\linewidth]{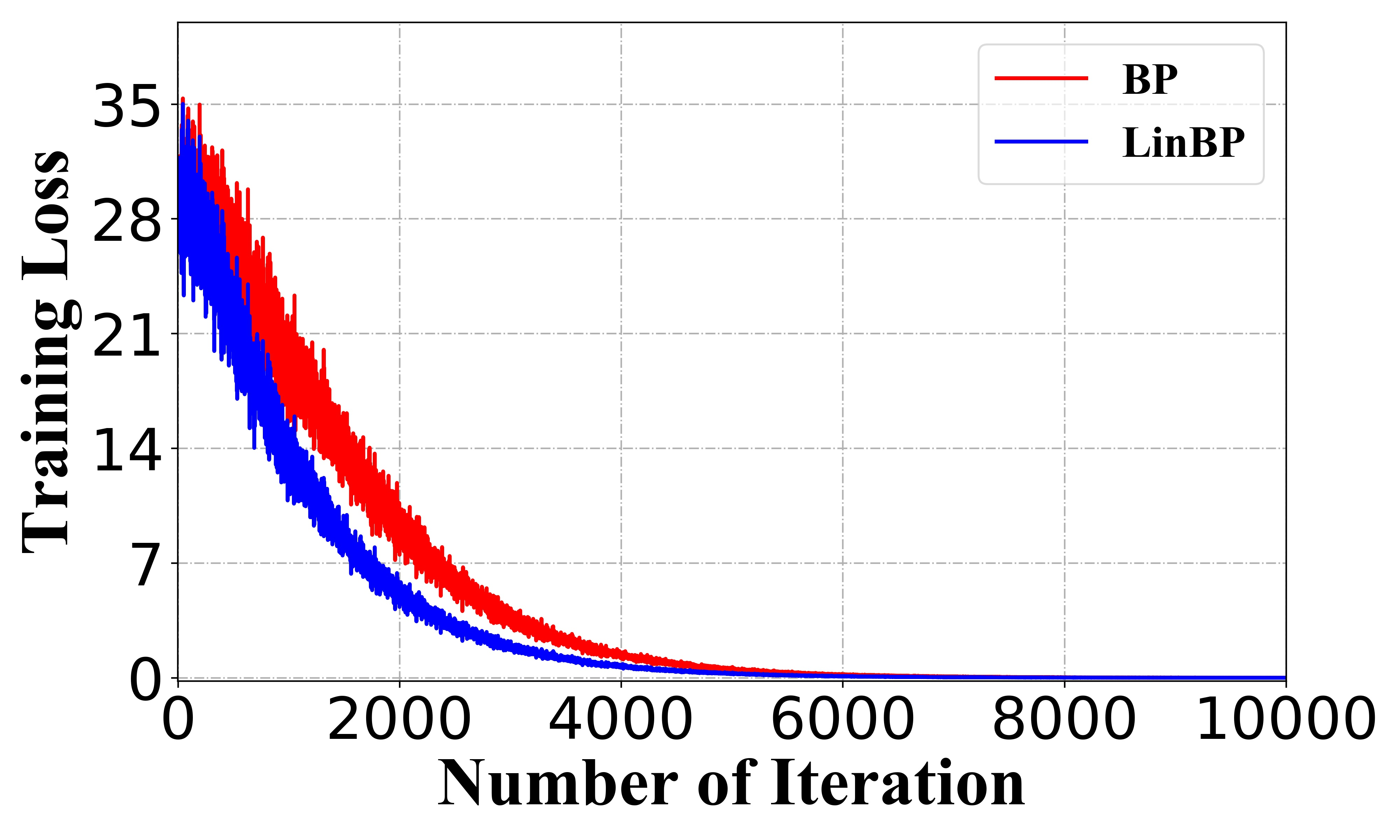}}
	\hskip 0.1in
	\subfloat[Approximation to the teacher]{\label{fig:T1_error}	
		\includegraphics[width=0.45\linewidth]{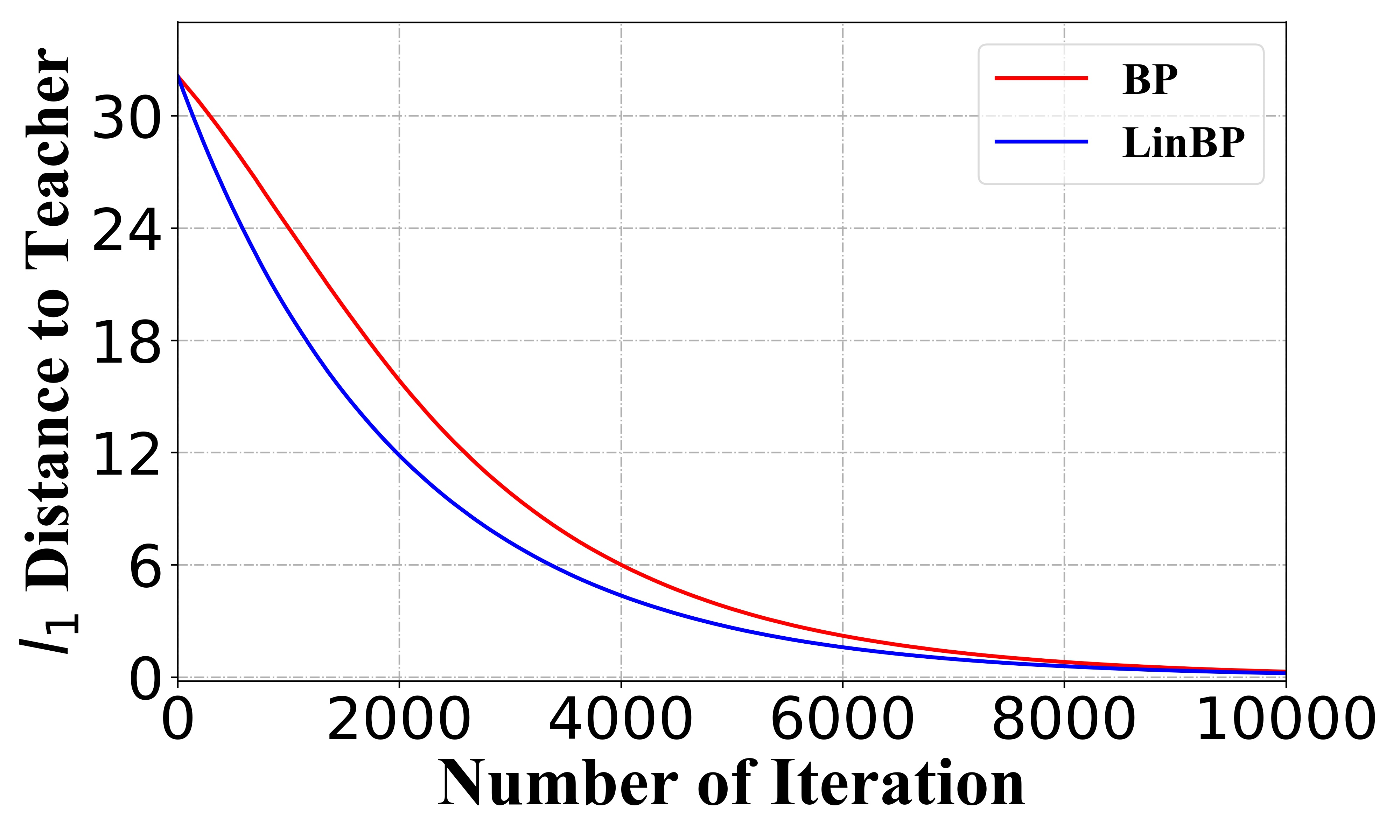}}
	\vskip -0.0in	
	\caption{LinBP leads to lower training loss and better approximation to the teacher model, especially at the early stage of training.}\label{fig:T1}
\end{figure}
\begin{figure}[htb]
	\centering \vskip -0.0in
	\subfloat[Training loss]{\label{fig:T1_loss_l2}	
		\includegraphics[width=0.45\linewidth]{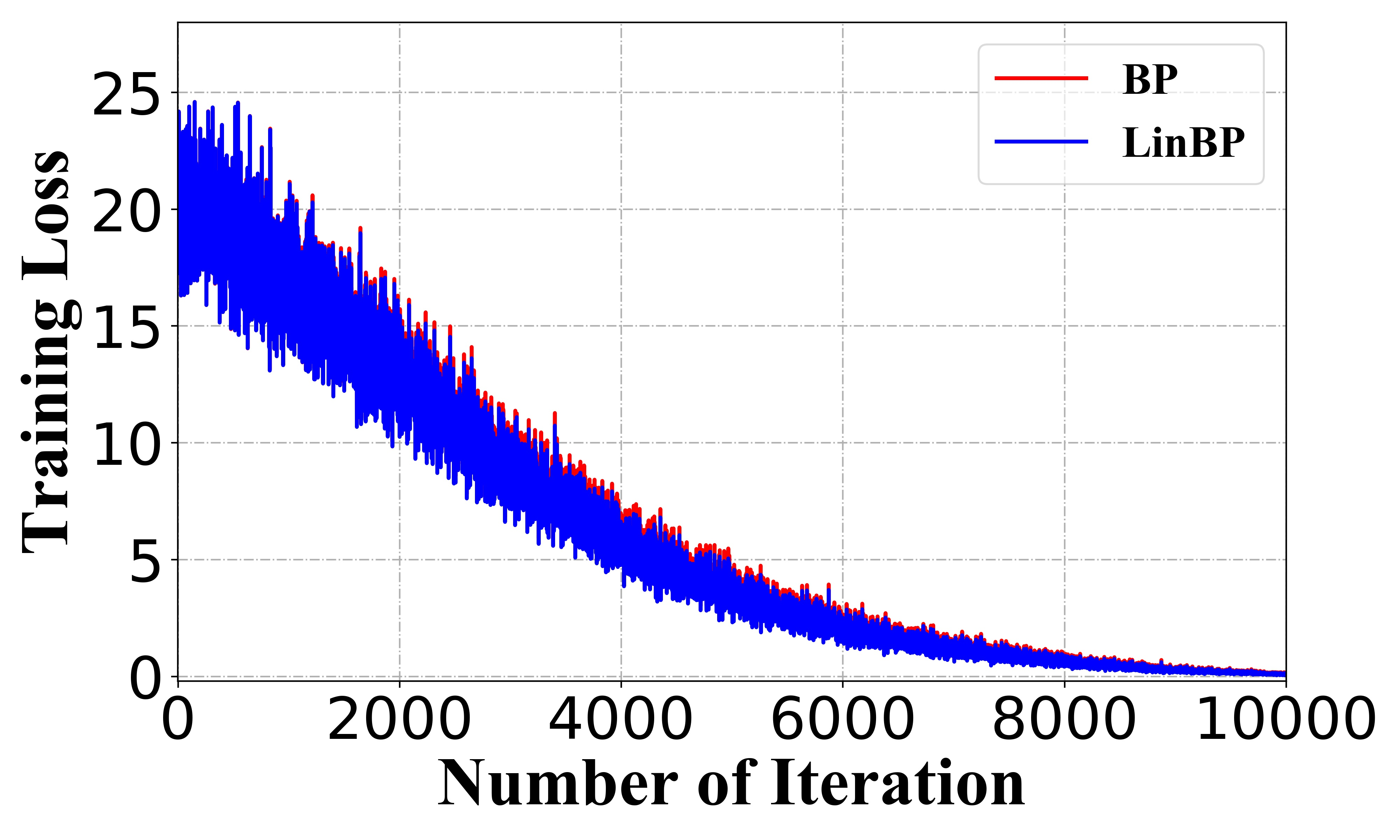}}
	\hskip 0.1in
	\subfloat[Approximation to the teacher]{\label{fig:T1_error_l2}	
		\includegraphics[width=0.45\linewidth]{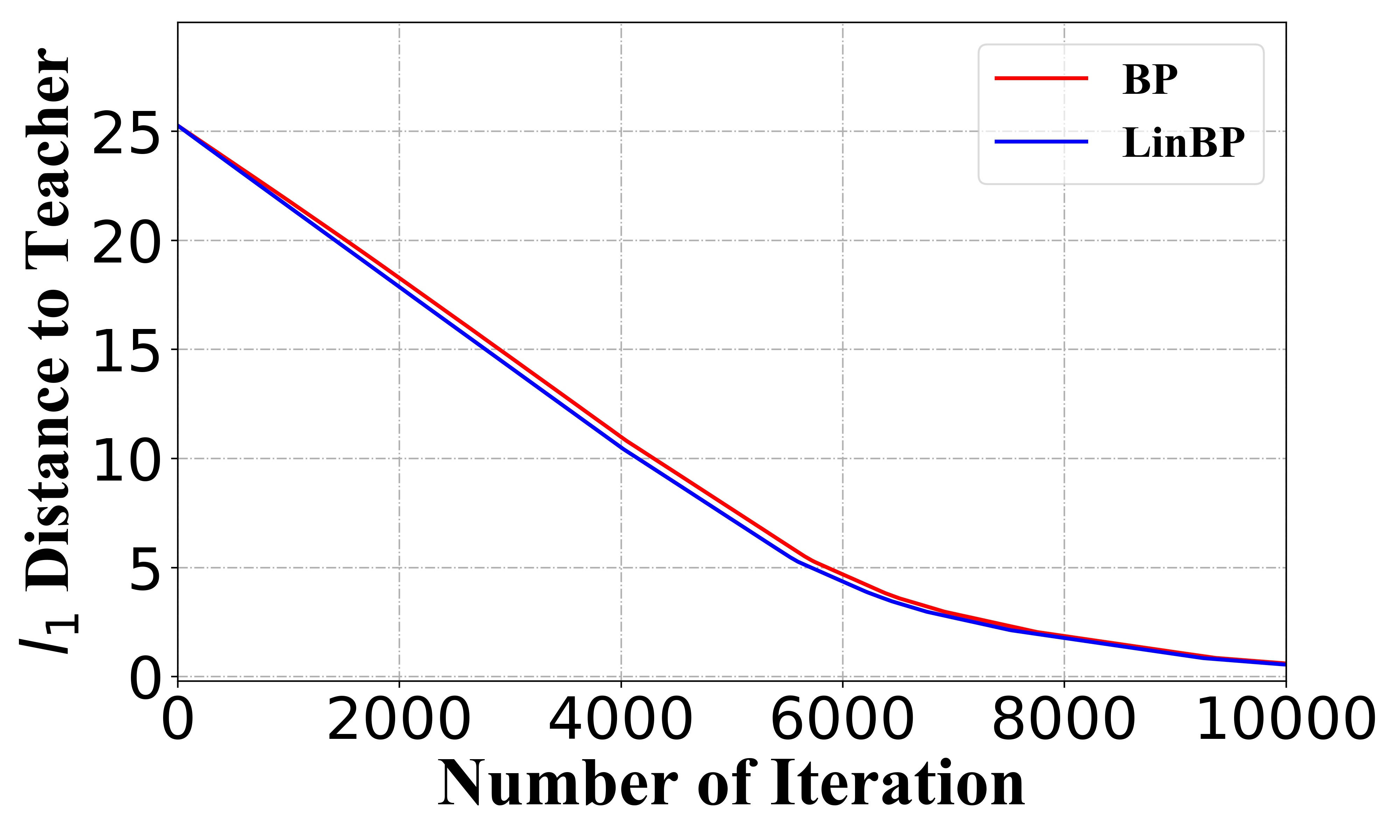}}
	\vskip -0.0in	
	\caption{LinBP leads to similar training loss and approximation to the teacher model with $l_2$ normalized gradients.}\label{fig:T1_l2}
\end{figure}
\subsubsection{Model training} \label{sec:exp_train}
As described in Section~\ref{sec:theo_train}, we constructed the one-layer neural network formulated as in Eq.~(\ref{ori}) and trained it following the student-teacher framework. Eq.~(\ref{loss}) is the loss function. Similar to the experiments in Section~\ref{sec:exp_attack}, the input data matrix $\mathbf X$ was generated from the standard Gaussian distribution, the optimal weight vector $\textbf{w}^\star$ in the teacher network and the initial weight vector $\textbf{w}^{(0)}$ in the student network were obtained via sampling from Gaussian distributions with $\mu_1=1.0$, $\mu_2=0.0$, $\sigma_1=3.0$, and $\sigma_2=1.0$. We used SGD for optimization and set $N=100$, $d=10$, and $\eta=0.001$ in our experiments. The maximal number of optimization iterations was set to $10000$ to ensure training convergence. We used the $l_1$ distance between the weight vector in the teacher model and that in the student model to evaluate their difference. Figure~\ref{fig:T1} illustrates the average results over 10 runs and compares the two methods. From the experimental results, we can observe that LinBP leads to more accurate approximation (\emph{i.e.}, smaller $l_1$ distance) to the teacher as well as lower training loss in comparison with BP, which clearly confirms our Theorem~\ref{theorem2}, indicating that LinBP can lead to faster convergence in the same hyper-parameter settings.

We found that the norm of gradient obtained from LinBP is larger than that obtained from the standard BP in the early stage of training, similar to the results in Section~\ref{sec:exp_attack}. To be specific, the $l_2$ norm of the gradient from LinBP is $\sim2.2\times$ larger than that from BP in the first 100 iterations, yet over all $10000$ iterations, it is only $\sim0.75\times$. We have also conducted the experiment where the gradient is normalized by its $l_2$ norm. $\eta$ was set to 0.0015. The results are shown in Figure~\ref{fig:T1_l2}. We see that the superiority of LinBP drops a lot (when compared with Figure~\ref{fig:T1}), which is different from the results obtained in adversarial attacks.

\subsection{More Practical Experiments}
\label{sec:exp_real}

{\renewcommand\arraystretch{1.3}
\begin{table*}[!t]
\vskip -0.0in
\caption{The highest success rate of white-box attacks using PGD with LinBP and BP in different settings over 5 restarts. $K$ is the maxiumum allowed number of update steps and $\epsilon$ is the perturbation budget. Higher success rate (\emph{i.e.}, lower prediction accuracy on the adversarial examples) indicates more powerful attack. "WRN(robust)" represents a robust WRN-28-10 model~\cite{zhang2021geometry} available on the RobustBench~\cite{croce2020robustbench}.}\vskip -0.1in
\label{table:cifar}
\begin{center}
    \begin{small}
        \setlength{\tabcolsep}{3mm}{\begin{tabular}{cccccccc}
			\hline
			Method &$K$& $\epsilon$ & VGG-16 & ResNet-50 & DenseNet-161 & MobileNetV2 & WRN (robust)\\
			\hline
			\multirow{6}{*}{BP}
			&10&8/255&72.28\%&66.41\%&56.89\%&92.94\% &33.14\%\\
			&20&8/255&87.07\%&80.82\%&73.33\%&98.19\% &38.11\%\\
                &100&8/255&96.54\%&92.53\%&86.46\%&99.88\%&38.94\%\\
			&10&16/255&77.25\%&71.52\%&62.79\%&94.90\% &33.78\%\\
			&20&16/255&94.88\%&91.18\%&87.38\%&99.44\% &44.25\%\\
                &100&16/255&99.73\%&99.81\%&98.35\%&\textbf{100.00\%}&59.38\%\\
			\hline
			\multirow{6}{*}{LinBP}
			&10&8/255&\textbf{95.52\%}&\textbf{85.83\%}&\textbf{61.87\%}&\textbf{99.80\%} &\textbf{33.48\%}\\
			&20&8/255&\textbf{98.50\%}&\textbf{93.11\%}&\textbf{77.72\%}&\textbf{99.95\%} &\textbf{39.55\%}\\

            &100&8/255&\textbf{99.25\%}&\textbf{95.82\%}&\textbf{89.03\%}&\textbf{99.99\%}&\textbf{40.44\%}\\
			&10&16/255&\textbf{98.78\%}&\textbf{92.68\%}&\textbf{69.04\%}&\textbf{99.95\%} &\textbf{34.21\%}\\
			&20&16/255&\textbf{99.99\%}&\textbf{99.51\%}&\textbf{91.18\%}&\textbf{100.00\%} &\textbf{45.22\%}\\

            &100&16/255&\textbf{100.00\%}&\textbf{99.97\%}&\textbf{99.07\%}&\textbf{100.00\%}&\textbf{60.78\%} \\
			\hline
	\end{tabular}}
    \end{small}
\end{center} \vskip -0.0in
\end{table*}}

{\renewcommand\arraystretch{1.3}
\begin{table*}[!t]
\vskip -0.0in
\caption{The highest success rate of white-box attacks using Auto-PGD with LinBP and BP in different settings over 5 restarts. $\epsilon$ is the perturbation budget. Higher success rate (\emph{i.e.}, lower prediction accuracy on the adversarial examples) indicates more powerful attack. "WRN(robust)" represents a robust WRN-28-10 model~\cite{zhang2021geometry} available on the RobustBench~\cite{croce2020robustbench}.
}\vskip -0.1in
\label{table:auto}
\begin{center}
    \begin{small}
        \setlength{\tabcolsep}{3mm}{\begin{tabular}{cccccccc}
			\hline
			Method & $\epsilon$ & VGG-16 & ResNet-50 & DenseNet-161 & MobileNetV2 & WRN (robust)\\
			\hline
			\multirow{2}{*}{BP}
			&8/255&98.06\%&95.77\%&91.91\%&99.97\% &39.46\%\\
			&16/255&99.97\%&99.98\%&99.58\%&\textbf{100.00\%}&60.94\%\\
			\hline
			\multirow{2}{*}{LinBP}
			&8/255&\textbf{99.65\%}&\textbf{97.60\%}&\textbf{93.35\%}&\textbf{100.00\%} &\textbf{41.20\%}\\
			&16/255&\textbf{100.00\%}&\textbf{99.99\%}&\textbf{99.81\%}&\textbf{100.00\%} &\textbf{62.99\%}\\
			\hline
	\end{tabular}}
    \end{small}
\end{center} \vskip -0.0in
\end{table*}}

% {\renewcommand\arraystretch{1.3}
% \begin{table*}[!t]
% \vskip -0.0in
% \caption{Success rate of white-box adversarial attacks using I-FGSM with LinBP and BP in different settings. Higher success rate (\emph{i.e.}, lower prediction accuracy on the adversarial examples) indicates more powerful attack.}\vskip -0.1in
% \label{table:cifar}
% \begin{center}
%     \begin{small}
%         \setlength{\tabcolsep}{3mm}{\begin{tabular}{ccccccc}
% 			\hline
% 			Method &$K$& $\epsilon$ & VGG-16 & ResNet-50 & DenseNet-161 & MobileNetV2\\
% 			\hline
% 			\multirow{4}{*}{BP}
% 			&5&0.05&71.37\%&66.97\%&58.25\%&91.88\%\\
% 			&5&0.1&71.37\%&66.97\%&58.25\%&91.88\%\\
% 			&10&0.05&90.99\%&86.25\%&82.94\%&98.56\%\\
% 			&10&0.1&93.25\%&90.76\%&87.53\%&98.87\%\\
% 			\hline
% 			\multirow{4}{*}{LinBP}
% 			&5&0.05&\textbf{97.80\%}&\textbf{90.23\%}&\textbf{82.56\%}&\textbf{99.85\%}\\
% 			&5&0.1&\textbf{97.80\%}&\textbf{90.23\%}&\textbf{82.56\%}&\textbf{99.85\%}\\
% 			&10&0.05&\textbf{99.90\%}&\textbf{98.67\%}&\textbf{96.31\%}&\textbf{100.0\%}\\
% 			&10&0.1&\textbf{100.00\%}&\textbf{99.72\%}&\textbf{99.81\%}&\textbf{100.0\%}\\
% 			\hline
% 	\end{tabular}}
%     \end{small}
% \end{center} \vskip -0.0in
% \end{table*}}

We also conducted experiments in more practical settings for adversarial attack and model training on MNIST~\cite{lecun1998gradient} and CIFAR-10~\cite{krizhevsky2009learning}, using DNNs with a variety of different architectures, including a simple MLP, LeNet-5~\cite{lecun1998gradient}, VGG-16~\cite{simonyan2015very}, ResNet-50~\cite{he2016deep}, DenseNet-161~\cite{huang2017densely}, MobileNetV2~\cite{sandler2018mobilenetv2}, and WRN (WRN)~\cite{zagoruyko2016wide}.  

\subsubsection{Adversarial attack on DNNs}
 
We performed white-box attacks on CIFAR-10 using different DNNs, including VGG-16, ResNet-50, DenseNet-161, MobileNetV2, and a robust WRN-28-10~\cite{zhang2021geometry} available on the RobustBench~\cite{croce2020robustbench}. 
CIFAR-10 test images that could be correctly classified by these models were used to generate adversarial examples, and the attack success rate was adopted to evaluate the performance of the attack. 
We evaluate the results of PGD~\cite{madry2018deep} as it is popular in white-box attacks, and we report the best attack performance over 5 restarts of PGD.
Note that, without restart, PGD and I-FGSM~\cite{kurakin2017adversarial} achieve similar performance.
The gradient steps were normalized by the $l_\infty$ norm in PGD by default. The update step size of the iterative attack $\eta$ was fixed to be $1/255$.
We tested different settings of the maximum allowed number of update steps $K$ and the perturbation budget $\epsilon$, including $K\in\{10, 20, 100\}$ and $\epsilon\in\{8/255, 16/255\}$.
The attack performance using LinBP and BP following all these settings is summarized in Table~\ref{table:cifar}. 

It is easy to see in Table~\ref{table:cifar} that LinBP gains consistently higher attack success rates with PGD, showing that it can help generate more powerful adversarial examples. For instance, with $K=10$ and $\epsilon =8/255$, using LinBP improves the success rate of attacking ResNet-50 by 19.42\% (85.83\% vs 66.41\%). 
With $K=100$, the success rates of attacking some (non-robust) models approach $\sim$100\%, and LinBP still helps achieve similar or higher attack performance than using BP.
As gradient normalization was by default adopted in PGD, the superiority of LinBP has nothing to do with the norm of gradient, just as in Section~\ref{sec:exp_attack}.

We have also tried using Auto-PGD~\cite{croce2020reliable}, which was developed as a stronger method for evaluating the adversarial robustness of models. 
Similar to the experiment with PGD, the best attack performance over 5 restarts are shown (see Table~\ref{table:auto}).
It can be observed that LinBP is still more effective than BP, even with Auto-PGD which has adaptive step sizes. To give more discussions on how LinBP enhances the performance in Auto-PGD, we carefully analyzed when the optimizer adjusted its step sizes with LinBP/BP. The results show that LinBP helps Auto-PGD halve the step size faster. More specifically, the average number of iterations before Auto-PGD halves its step size using LinBP is 22.95, 42.18, 58.59, and 72.11, compared with 25.85, 45.11, 61.98 and 77.45 for BP when attacking ResNet-50 with $\epsilon=8/255$. 
Since Auto-PGD adjusts its step size when it reaches ``convergence'' with the current step size, the results show that LinBP is beneficial to convergence even with such a method in practice.
% Our theoretical analyses show that LinBP improves the performance when the step size is fixed, thus it is helpful to fulfil the condition for halving step size in Auto-PGD. The experimental results confirm our inference.}
% We also find that the success rate for LinBP using PGD attack ($K=100$) is 95.72\%. The results show that LinBP may lead to a stronger attack.]

LinBP can be effective with other activation functions. To gain more insight, we further tested with the Swish/SiLU activation function. 
For simplicity of experiments, we replace the original ReLU activations in ResNet-50 with the Swish/SiLU activations. With $K=10$ and $\eta=1/255$, the PGD attack success rate for LinBP are 75.85\% and 87.97\% when $\epsilon=8/255$ and $\epsilon=16/255$, respectively, compared with 74.24\% and 82.65\% for BP. %We find LinBP still gain a little higher attack success rates.

\subsubsection{Training DNNs}
\begin{figure}[htb]
	\centering
	\vskip -0.0in
	\subfloat[Training loss]{\label{fig:MNIST_loss}	
		\includegraphics[width=0.45\linewidth]{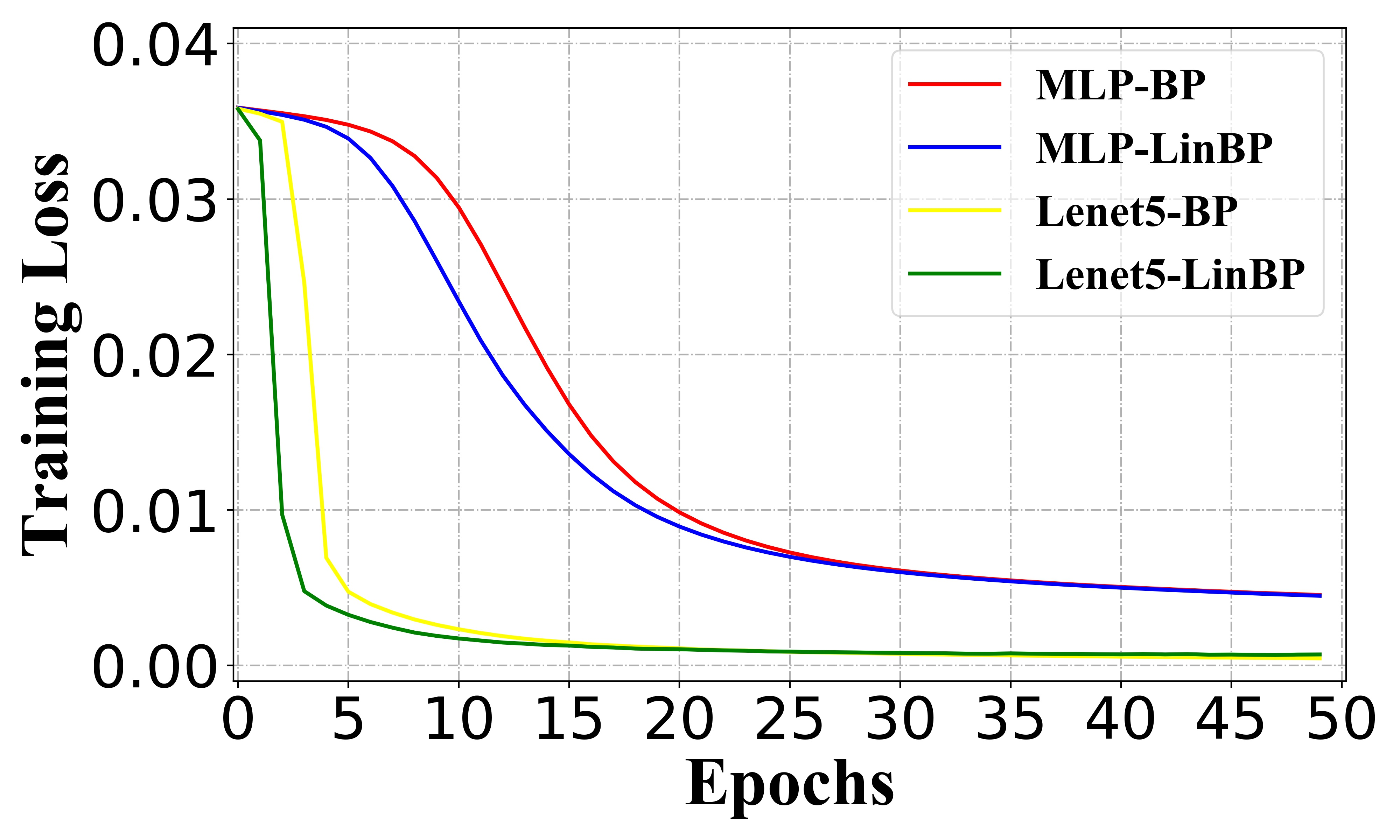}}
	\hskip 0.1in
	\subfloat[Training accuracy]{\label{fig:MNIST_acc}	
		\includegraphics[width=0.45\linewidth]{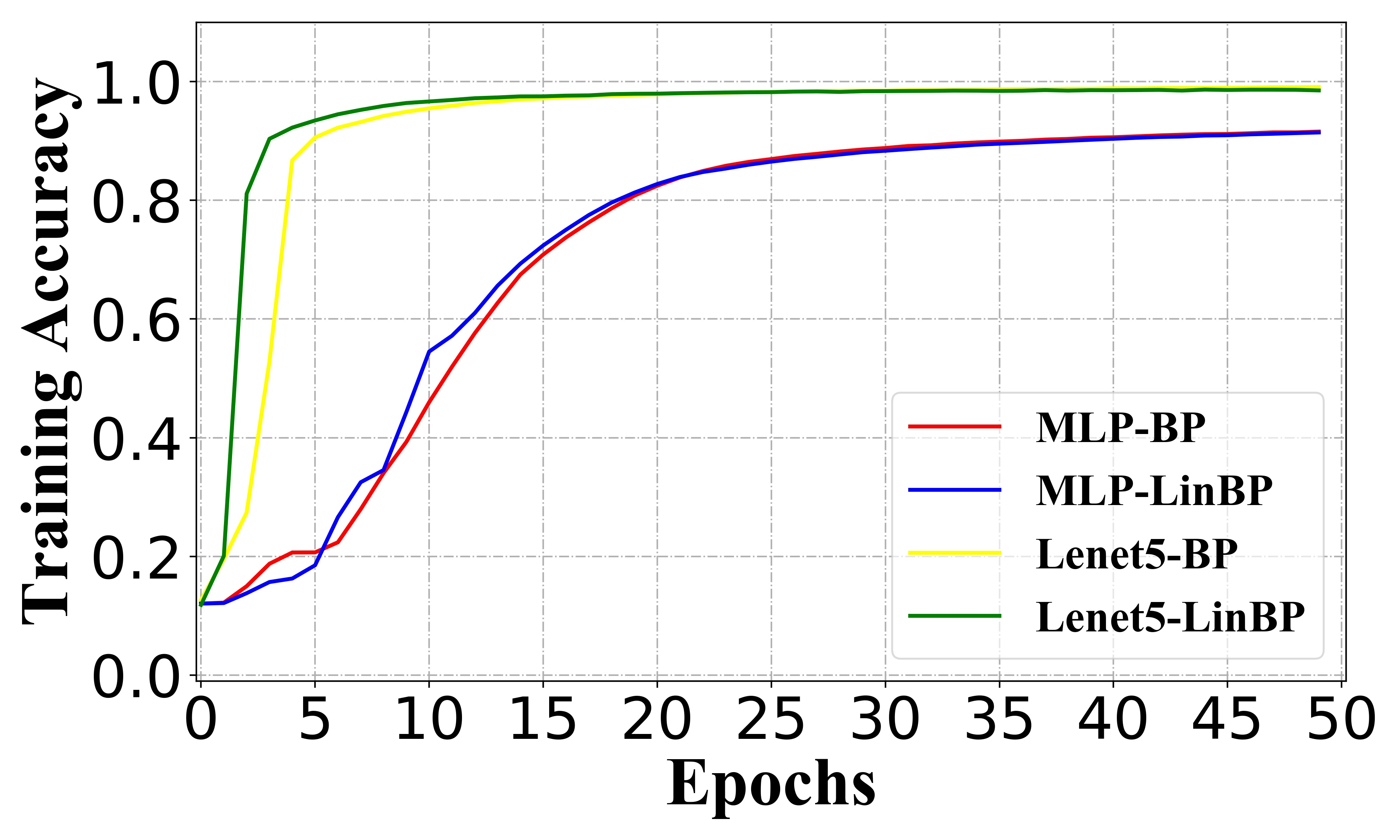}}
	\vskip -0.0in	
	\caption{Compare the training loss and training accuracy of the MLP and LeNet-5 on MNIST, using LinBP or BP.}\label{fig:mnist_train}
	\vskip -0.0in
\end{figure}

\begin{figure}[htb]
	\centering
	\vskip -0.0in
	\subfloat[Test loss]{\label{fig:MNIST_loss_test}	
		\includegraphics[width=0.45\linewidth]{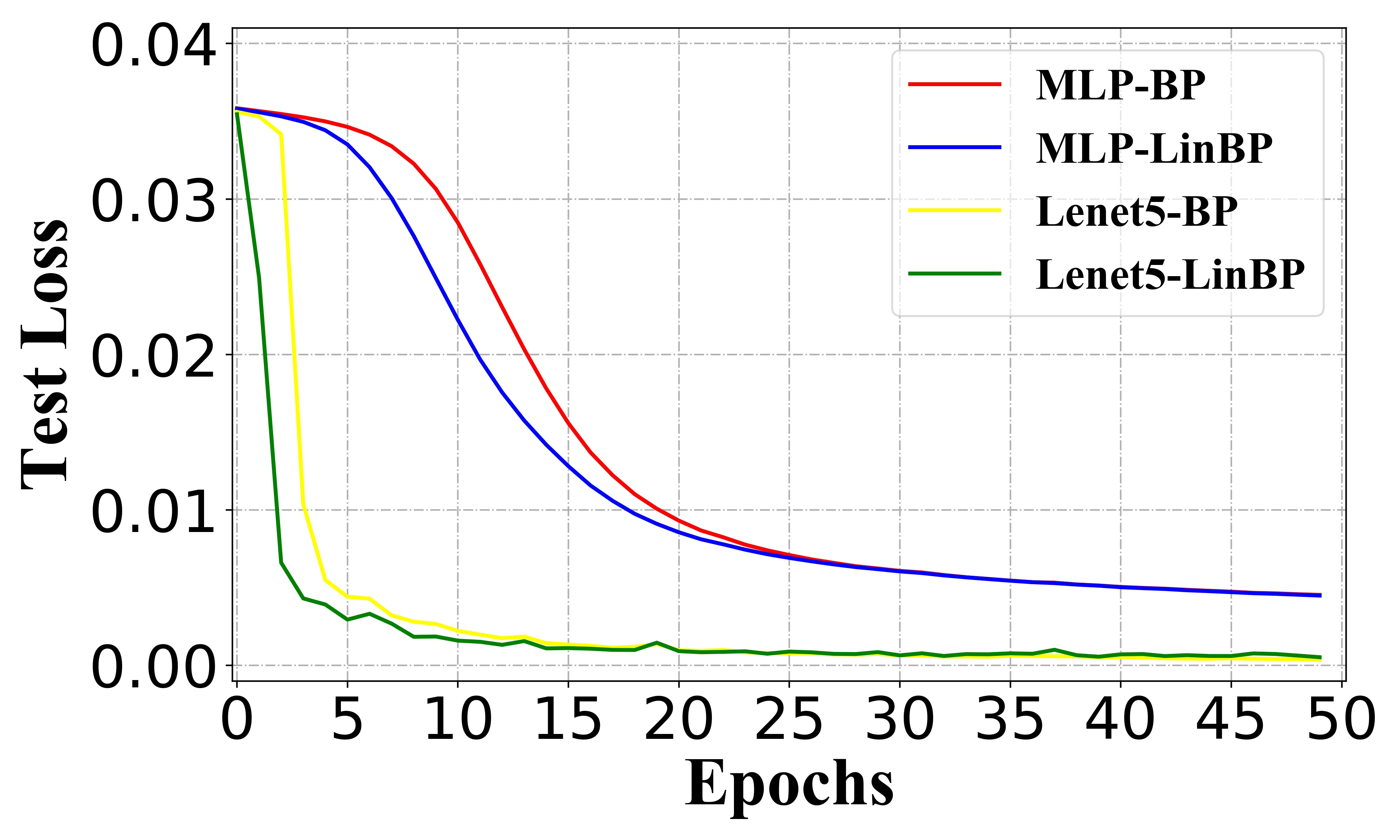}}
	\hskip 0.1in
	\subfloat[Test accuracy]{\label{fig:MNIST_acc_test}	
		\includegraphics[width=0.45\linewidth]{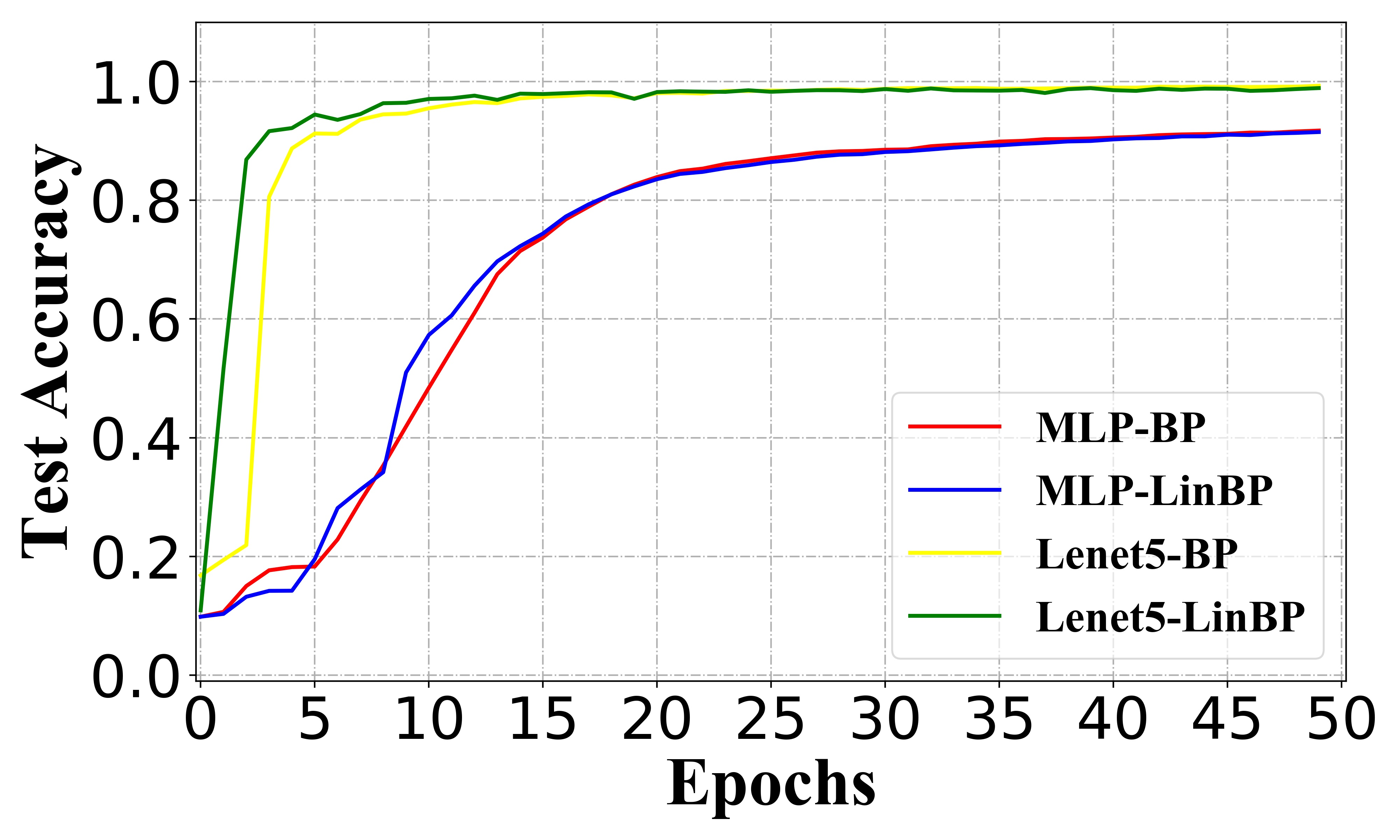}}
	\vskip -0.0in	
	\caption{Compare the test loss and test accuracy of the MLP and LeNet-5 on MNIST, using LinBP or BP.}\label{fig:mnist_test}
 	\vskip -0.0in
\end{figure}

For experiments of DNN training, we first trained and evaluated a simple MLP and LeNet-5 on MNIST. The MLP has four parameterized and learnable layers, and the numbers of its hidden layer units are 400, 200, and 100. We used SGD for optimization and the learning rate was 0.001 and 0.005 for the MLP and LeNet-5, respectively. The training batch size was set to $64$, and the training process lasted for at most $50$ epochs. The training loss and training accuracy are illustrated in Figure~\ref{fig:mnist_train}, and note that we fix the random seed to eliminate unexpected randomness during training. We can easily observe from the training curves in Figure~\ref{fig:mnist_train} that the obtained MLP and LeNet-5 models show lower prediction loss and higher accuracy when incorporating LinBP, especially in the early epochs, which suggests that the incorporation of LinBP can be beneficial to the convergence of SGD. The same observation can be made on the test set of MNIST, and the results are shown in Figure~\ref{fig:mnist_test}.


\begin{figure}[htb]
	\centering
	\subfloat[Training loss]{\label{fig:CIFAR_loss}	
		\includegraphics[width=0.45\linewidth]{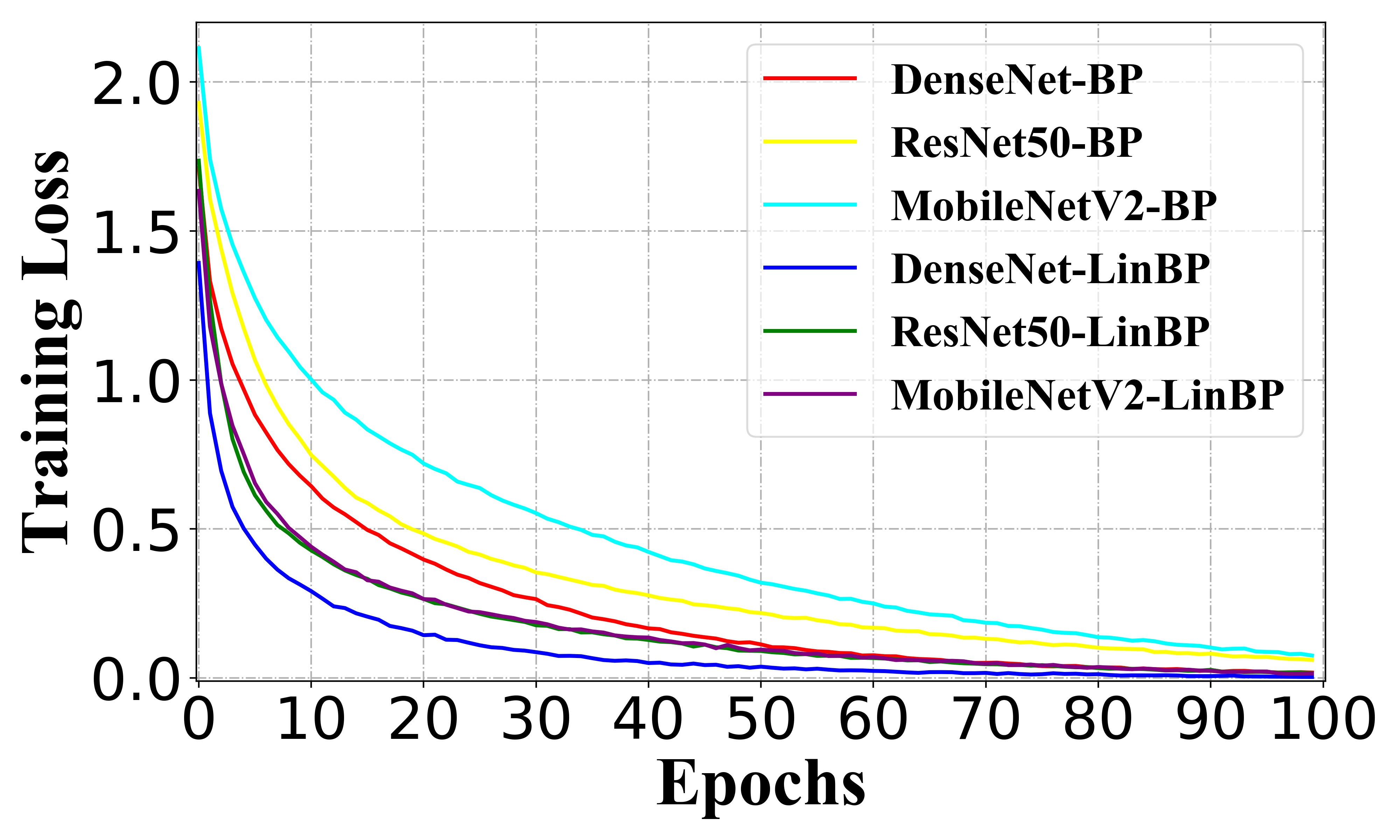}}
	\hskip 0.1in
	\subfloat[Training accuracy]{\label{fig:CIFAR_acc}	
		\includegraphics[width=0.45\linewidth]{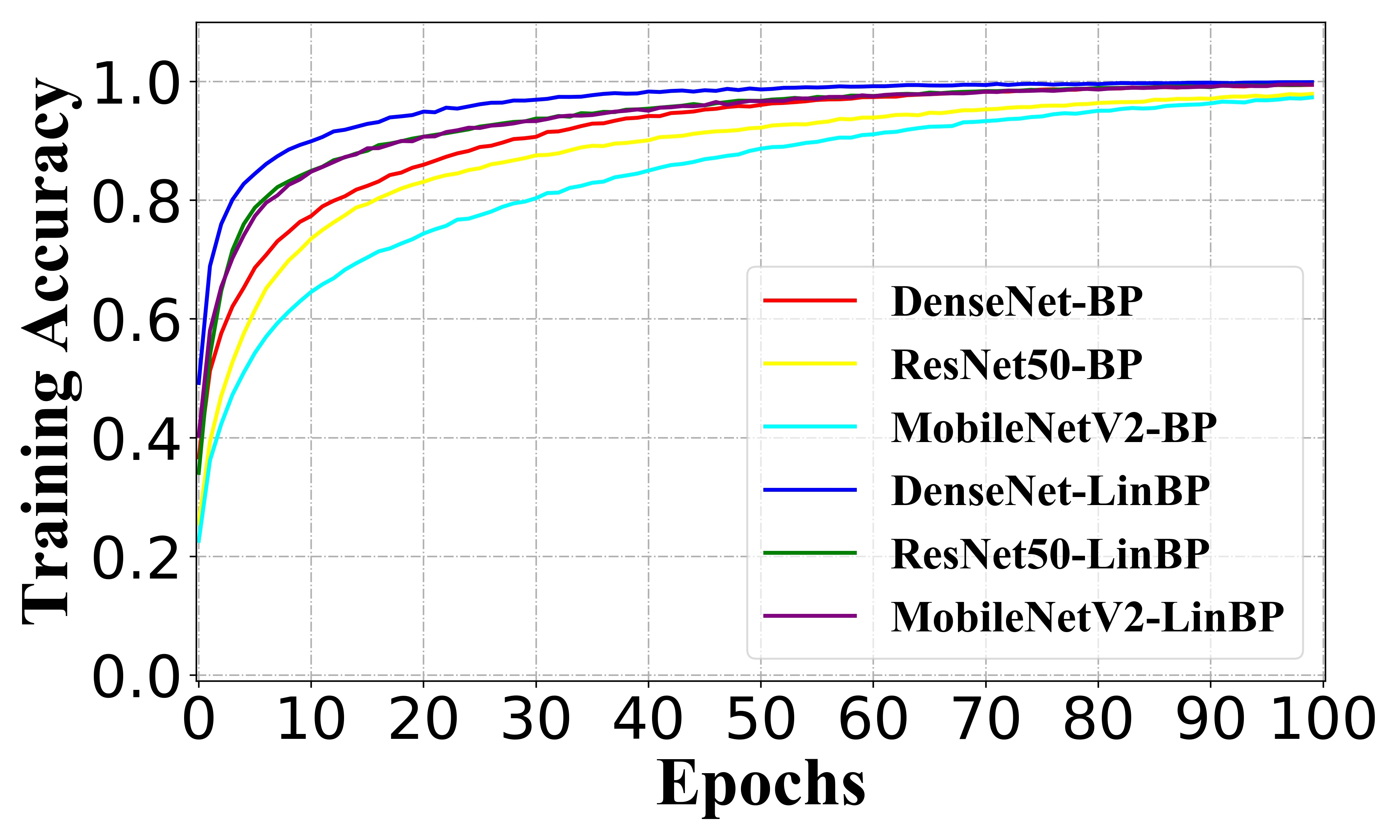}}
	\caption{Training loss and training accuracy of ResNet-50, DenseNet-161, and MobileNetV2 on CIFAR-10, using LinBP or BP.}\label{fig:CIFAR}
\end{figure}

\begin{figure}[htb]
	\centering
	\subfloat[Test loss]{\label{fig:CIFAR_loss_test}	
		\includegraphics[width=0.45\linewidth]{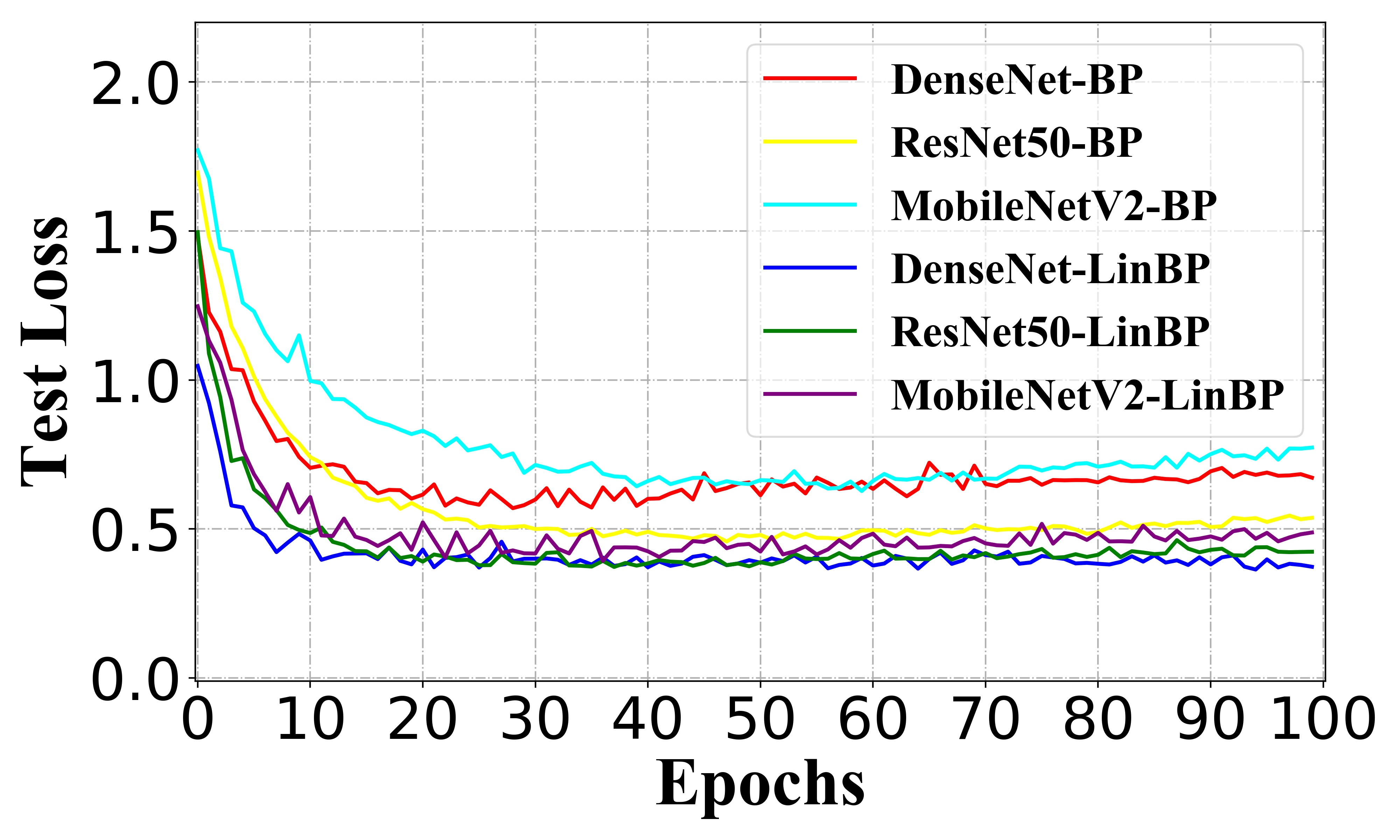}}
	\hskip 0.1in
	\subfloat[Test accuracy]{\label{fig:CIFAR_acc_test}	
		\includegraphics[width=0.45\linewidth]{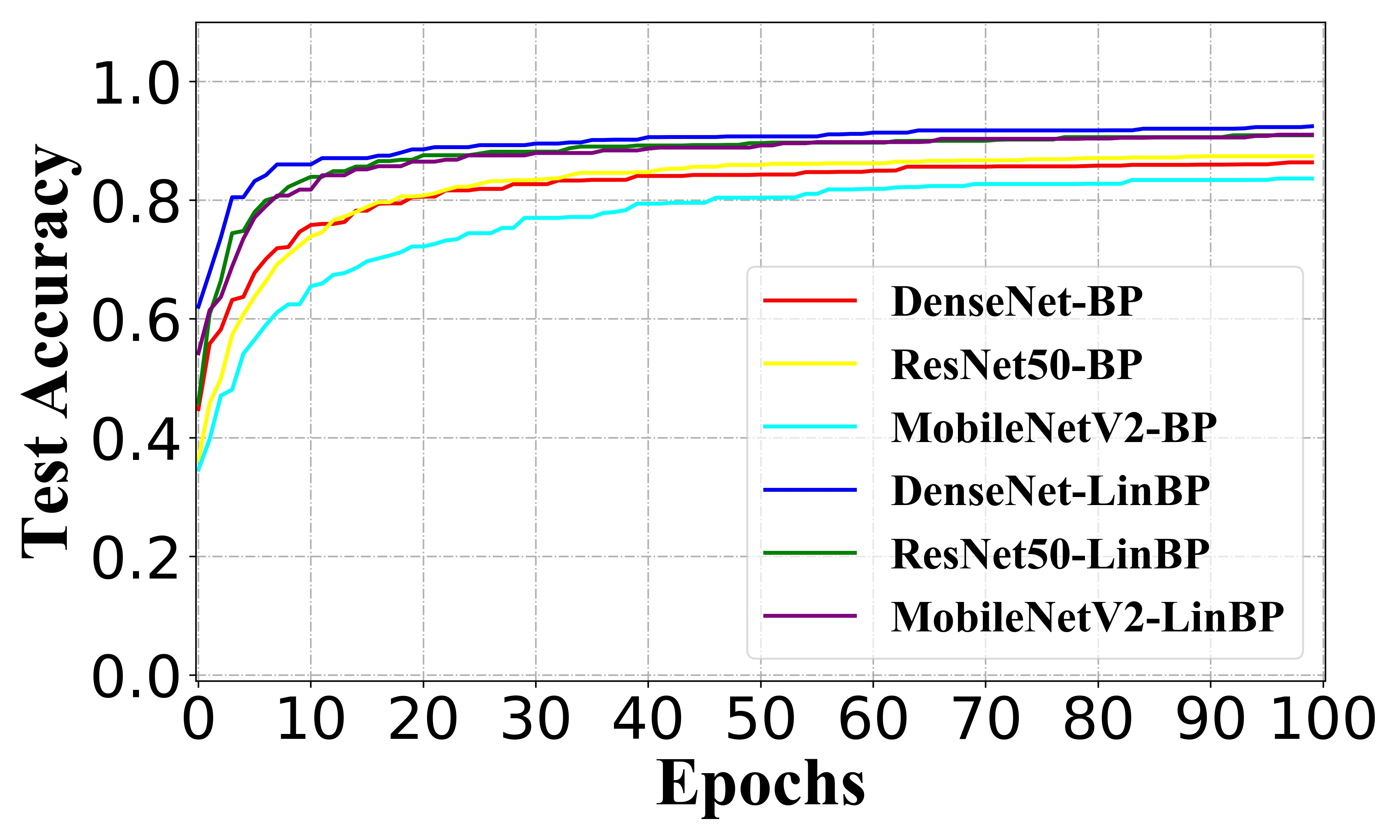}}
	\caption{Test loss and test accuracy of ResNet-50, DenseNet-161, and MobileNetV2 on CIFAR-10, using LinBP or BP.}\label{fig:CIFAR_test}
\end{figure}

We further report our experimental results on CIFAR-10. ResNet-50~\cite{he2016deep}, DenseNet-161~\cite{huang2017densely}, and MobileNetV2~\cite{sandler2018mobilenetv2} were trained and evaluated on the dataset. The architecture of these networks and their detailed settings can be found in their papers. The optimizer was SGD, and the learning rate was $0.002$. We set the batch size to $128$ and trained for 100 epochs. We evaluated the prediction loss and prediction accuracy of trained models using LinBP and BP for comparison. Similar to the experiment on MNIST, we fixed the random seed. The training results in Figure~\ref{fig:CIFAR} and test results in Figure~\ref{fig:CIFAR_test} demonstrate that, equipped with LinBP, the models achieved lower prediction loss and higher prediction accuracy in the same training settings, especially when the training just got started. Nevertheless, as training progressed, the superiority of LinBP became less obvious, and the final performance of LinBP became just comparable to that of the standard BP, which is consistent with our theoretical discussions in Section~\ref{sec:theo_train}.


\begin{figure}[!t]
	\centering
	\vskip -0.0in
	\subfloat[$0.1\times$ learning rate]{\label{fig:MNIST_acc_mid}	
		\includegraphics[width=0.45\linewidth]{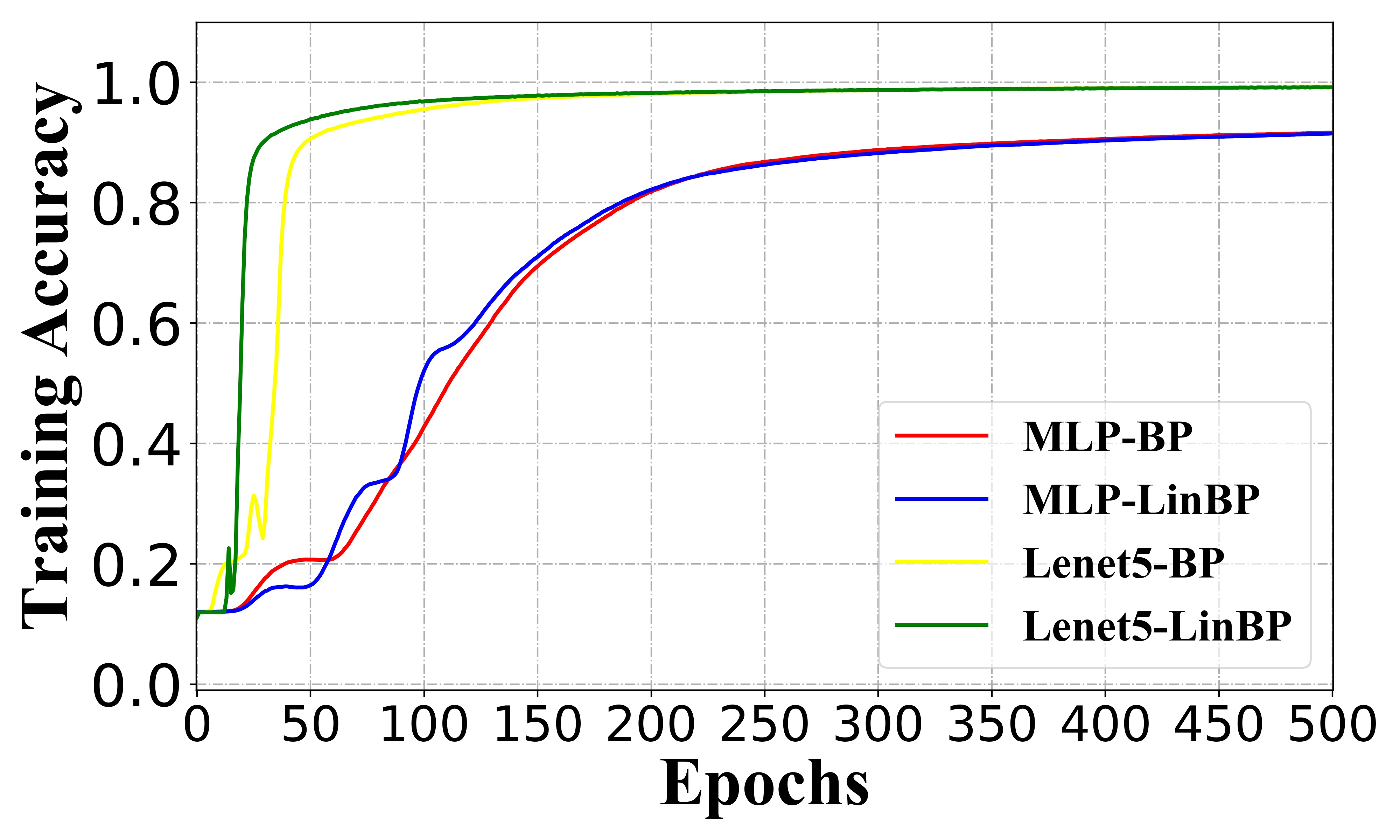}}
	\hskip 0.1in
	\subfloat[$10\times$ learning rate]{\label{fig:MNIST_acc_high}	
		\includegraphics[width=0.45\linewidth]{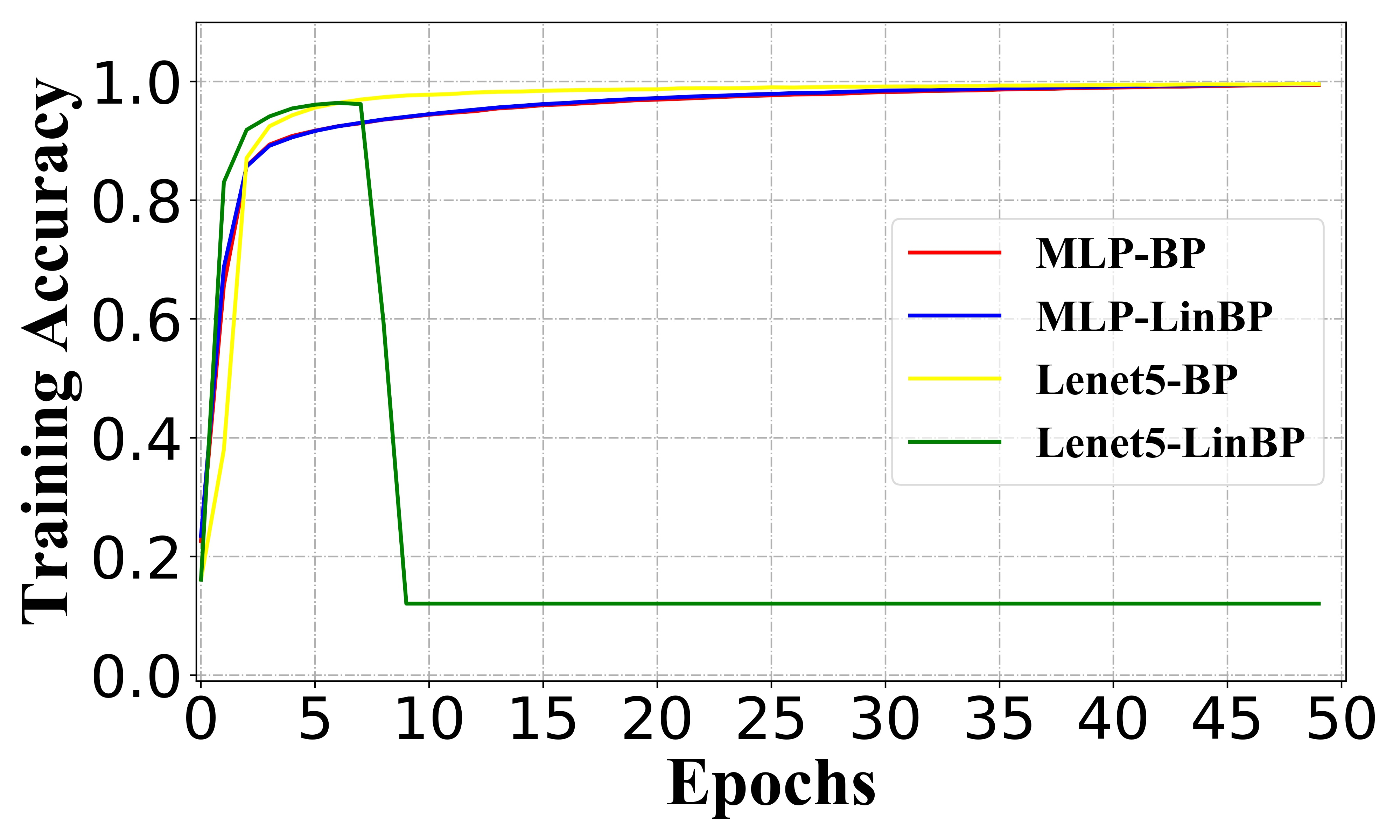}}
	\caption{The influence of setting different learning rate to LinBP and BP for training the MLP and LeNet-5 on MNIST.}\label{fig:learning_rate}
	\vskip -0.1in
\end{figure}

We also noticed that in certain scenarios, optimization using LinBP may fail to converge with very large learning rates. Here we test various learning rate $\eta$ in training MNIST models to observe the stability of LinBP. Recall that the base learning rates for training the MLP and LeNet-5 are $0.001$ and $0.005$, respectively, for obtaining our previous results, and we further tried scaling them by $10\times$ and $0.1\times$ to investigate how the training performance was affected. The training curves for all these settings of learning rates are illustrated in Figure~\ref{fig:learning_rate}. Compared with the results in Figure~\ref{fig:MNIST_acc}, it can be seen that, when increasing the base learning rate by $10\times$, the training performance of LinBP and that of BP are similar on the MLP, but, on LeNet-5, the training failed to converge with LinBP while it could still converge with BP. Also, the same observation was made when training VGG-16 on CIFAR-10 using LinBP and a (relatively large) base learning rate of $0.005$. The results conform that a reasonably small $\eta$ helps stabilize training using LinBP and guarantees its performance. We also tried using normalized gradients during training and LinBP performed no better than BP, just like the results in Section~\ref{sec:exp_train}.

\begin{figure}[htb]
	\centering
	\subfloat[Training accuracy]{\label{fig:CIFAR_loss2}	
		\includegraphics[width=0.45\linewidth]{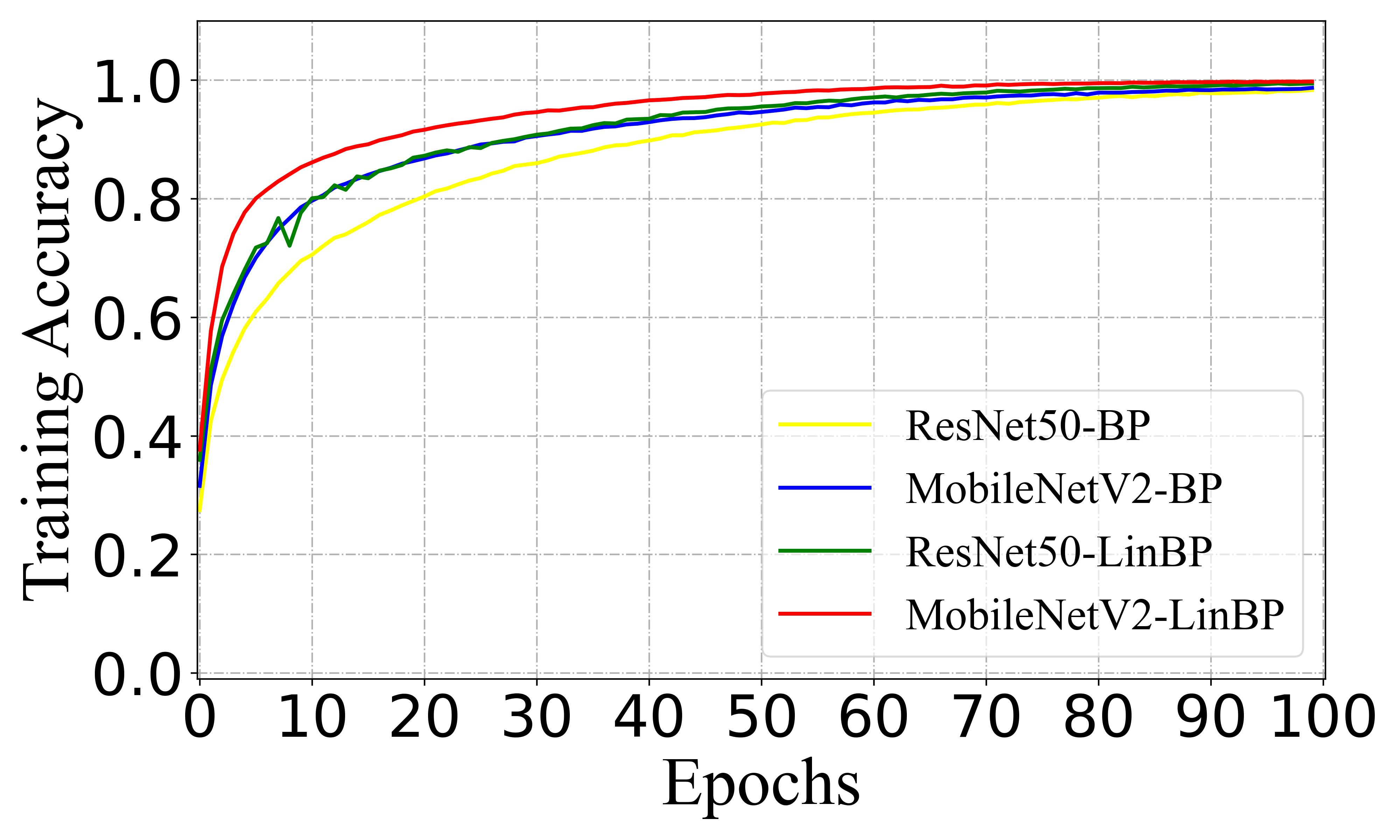}}
	\hskip 0.1in
	\subfloat[Test accuracy]{\label{fig:CIFAR_acc2}	
		\includegraphics[width=0.45\linewidth]{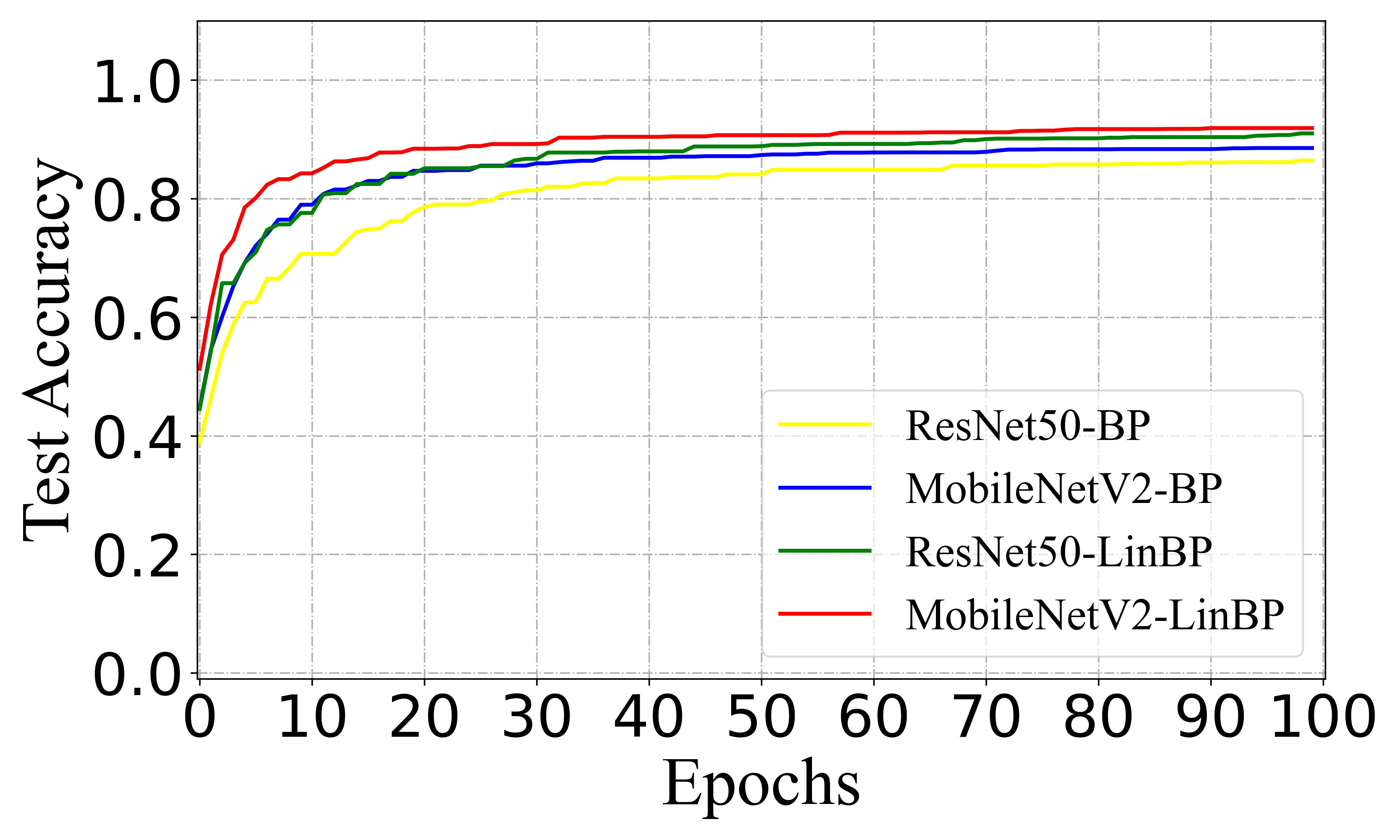}}
	\caption{Training ResNet-50 and MobileNetV2 on CIFAR-10 with the exponential learning rate scheduler, using LinBP or BP.}\label{fig:CIFAR2}
\end{figure}

We used the cosine annealing learning rate scheduler~\cite{loshchilov2016sgdr} for obtaining the results in Figure~\ref{fig:CIFAR} and~\ref{fig:CIFAR_test}. One may also be curious about the training performance with other learning rate scheduler. In this context, we further tested with exponential learning rate scheduler~\cite{li2019exponential} with $\gamma = 0.98$ and find similar results, which are shown in Figure~\ref{fig:CIFAR2}.

\subsubsection{Adversarial training for DNNs}
Since the discovery of adversarial examples, the vulnerability of DNNs has been intensively discussed. Tremendous effort has been devoted to improving the robustness of DNNs. Thus far, a variety of methods have been proposed, in which adversarial training~\cite{goodfellow2015explaining, madry2018deep} has become an indispensable procedure in many application scenarios. In this context, we would like to study how LinBP can be adopted to further enhance the robustness of DNNs. We used the adversarial examples generated by PGD to construct our dataset and we trained our model using simply SGD. There exist multiple strategies of adopting LinBP in adversarial training,~\emph{i.e.}, 1) computing gradients for updating model parameters using LinBP as in the model training experiment and 2) generating adversarial examples using LinBP just like in the adversarial attack experiment. 
We observed that the first strategy is more beneficial than the second strategy when evaluating the robustness of DNNs using BP-generated adversarial examples. 
Due to overfitting, generating adversarial examples using LinBP during adversarial training leads to obviously larger generalization gap between training and test performance and degraded test robustness against BP-generated adversarial examples.

\begin{figure}[htb]
	\centering
	\vskip -0.05in	
	\subfloat[MobileNetV2]{\label{fig:mobilenet_adv_train}
        \includegraphics[width=0.45\linewidth]{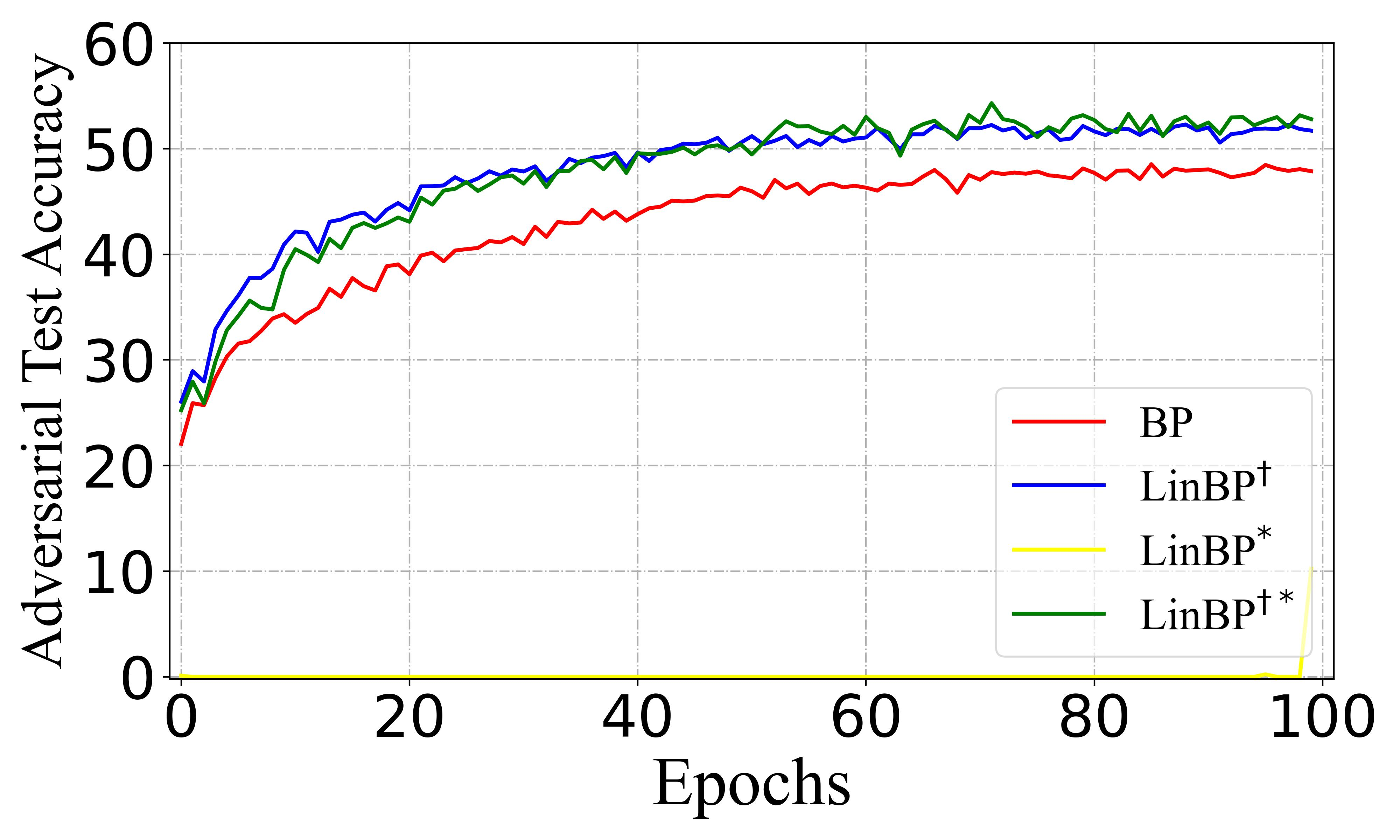}}
        \subfloat[ResNet-50]{\label{fig:resnet_adv_train}
        \includegraphics[width=0.45\linewidth]{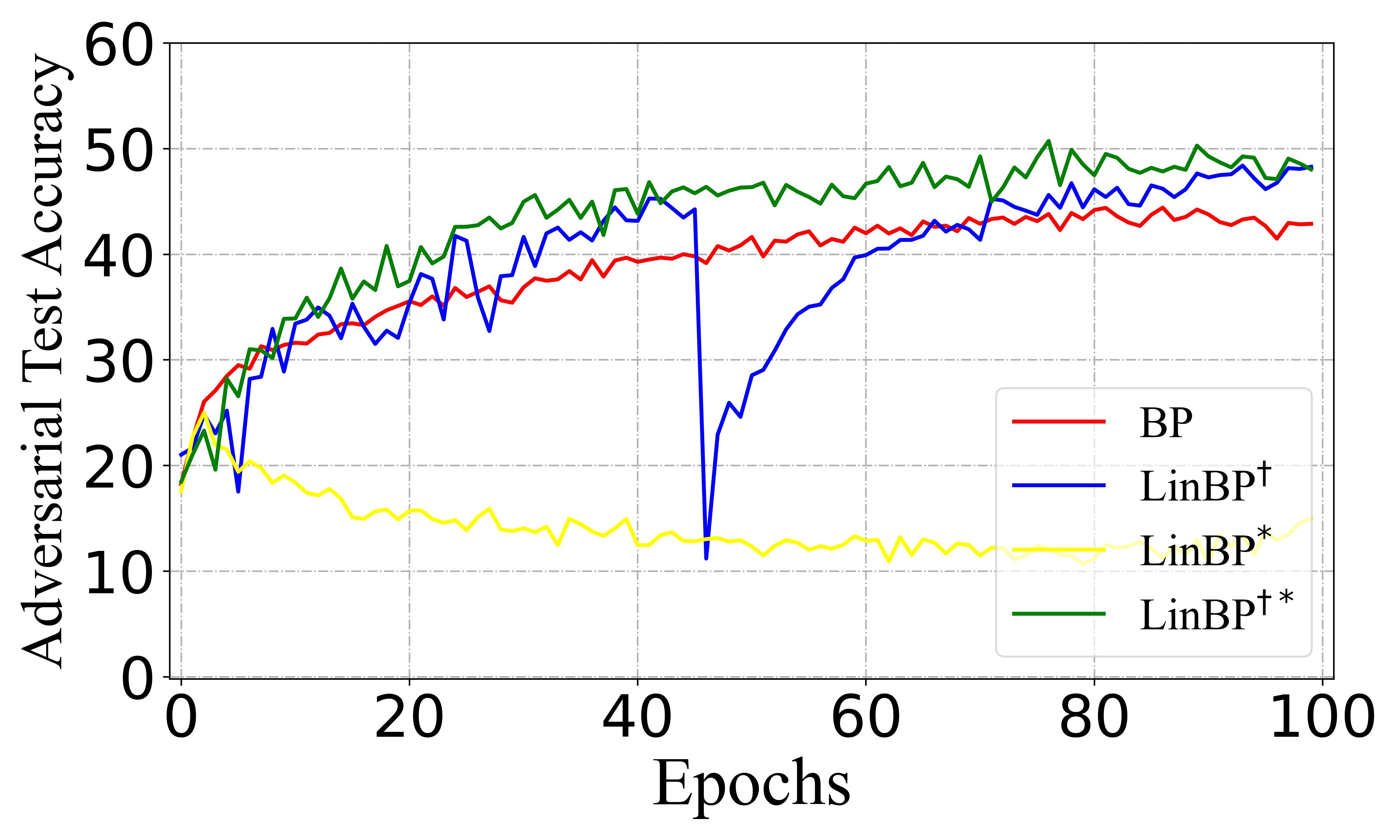}}
	\caption{Compare the PGD robustness of MobileNetV2 and ResNet-50 shielded with different adversarial training methods. We use ``LinBP$^\dagger$'', ``LinBP$^\ast$'', and ``LinBP$^{\dagger\ast}$'' to indicate models trained using the first strategy, the second strategy, and the combination of the two strategies as described in the above paragraph, respectively.}\label{fig:CIFAR_adv_train}
\end{figure}

\begin{figure}[htb]
	\centering
	\vskip -0.05in	
	\subfloat[MobileNetV2]{\label{fig:mobilenet_adv_train_aa}
        \includegraphics[width=0.45\linewidth]{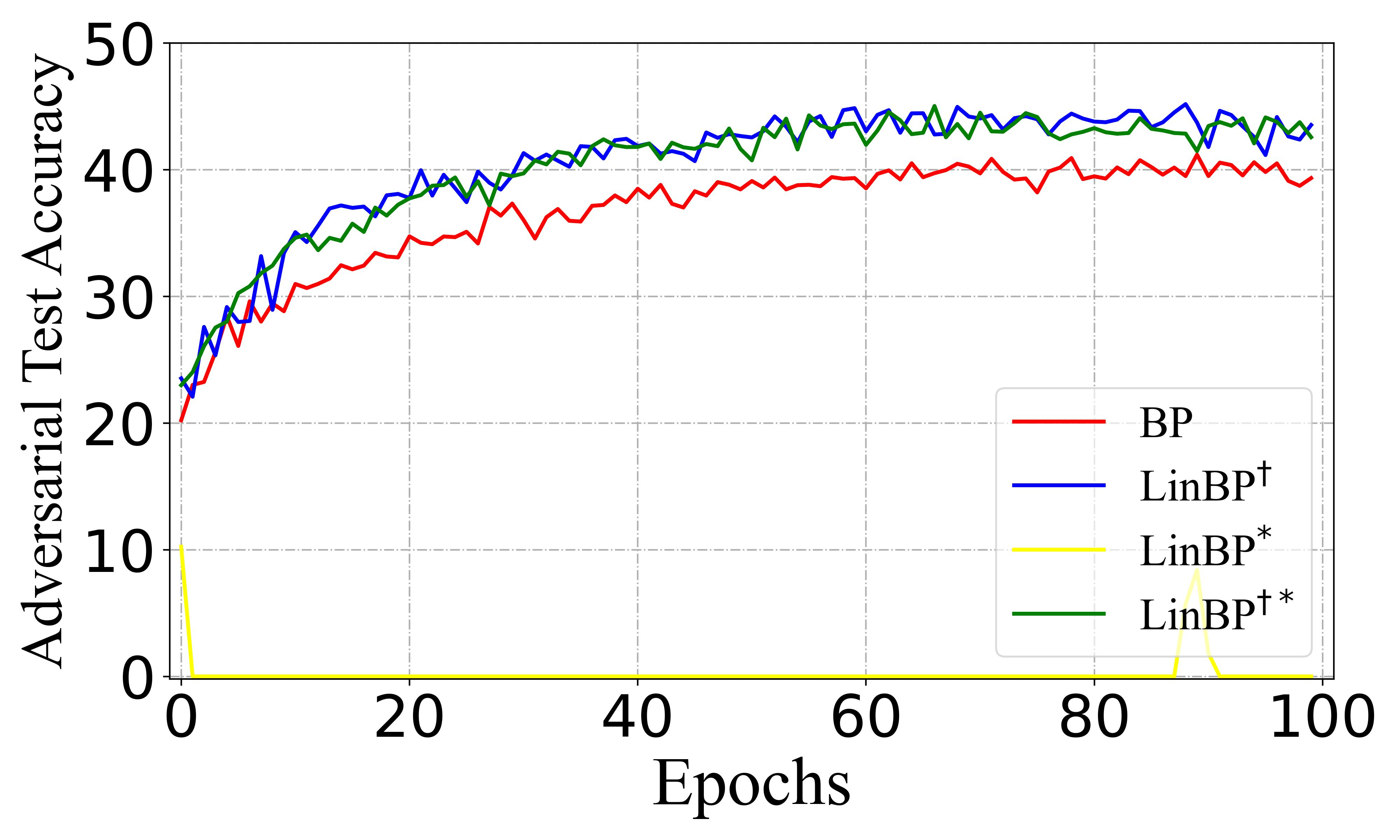}}
        \subfloat[ResNet-50]{\label{fig:resnet_adv_train_aa}
        \includegraphics[width=0.45\linewidth]{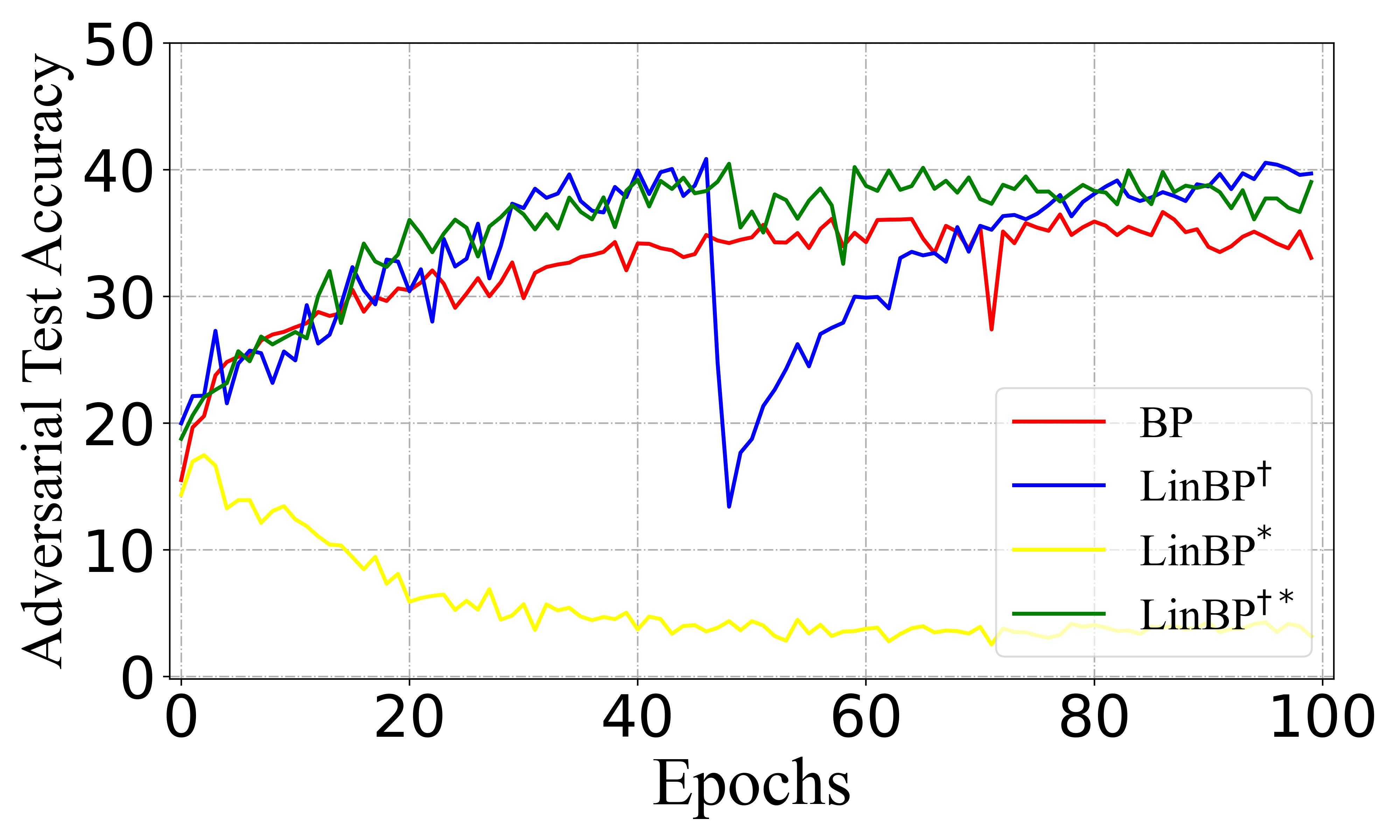}}
	\caption{Compare the AutoAttack robustness of MobileNetV2 and ResNet-50 shielded with different adversarial training methods. }\label{fig:CIFAR_adv_train_aa}
\end{figure}

Though effective in the sense of achieving high robust accuracy after convergence, the first strategy sometimes suffers from unstable performance as the adversarial examples and model parameters are updated asynchronously. Further applying the second strategy in combination with the first strategy stabilize adversarial training and is helpful to achieve reliable performance. See Figures~\ref{fig:CIFAR_adv_train} and~\ref{fig:CIFAR_adv_train_aa} for performance under the PGD attack and AutoAttack, respectively. 
In the figures, we applied the classification accuracy on adversarial examples to evaluate the robustness. The experiment was performed on CIFAR-10 using MobileNetV2 and ResNet-50, where the attack step size was $2/255$, $\epsilon$ was $8/255$, $K=5$, and the training learning rate was 0.01. 

The figures demonstrate that, benefit from it fast convergence, LinBP can be used as an alternative to BP when performing adversarial training for achieving robustness (if used appropriately). %The overfitting phenomenon should be carefully studied, if the LinBP-generated adversarial examples are anticipated to be involved in adversarial training.
%, especially when the training time budget is limited.

\section{Conclusion}\label{sec:conclusion}
In this paper, we have studied the convergence of optimization using LinBP, which skips ReLUs during the backward pass, and compared it to optimization using the standard BP. Theoretical analyses have been carefully performed in two popular application scenarios,~\emph{i.e.}, white-box attack and model training. In addition to the benefits in black-box attacks which has been demonstrated in~\cite{guo2020backpropagating}, we have shown in this paper that LinBP also leads to generating more destructive white-box adversarial examples and obtaining faster model training. Experimental results using synthesized data confirm our theoretical results. Extensive experiments on MNIST and CIFAR-10 further show that the theoretical results hold in more practical settings on a variety of DNN architectures. % Our theoretical and empirical discussions in the paper perform thorough analysis on LinBP, and may become an inspiration for more researches on convergence analysis in neural-network-involved learning tasks.

\section*{Acknowledgments}
This work is funded by the Natural Science Foundation of China (NSFC. No. 62176132) and the Guoqiang Institute of Tsinghua University, with Grant No. 2020GQG0005.

% or


%\appendices


% % use section* for acknowledgment
% \ifCLASSOPTIONcompsoc
%   % The Computer Society usually uses the plural form
%   \section*{Acknowledgments}
% \else
%   % regular IEEE prefers the singular form
% \fi


\ifCLASSOPTIONcaptionsoff
  \newpage
\fi



%\bibliographystyle{IEEEtran}
\bibliographystyle{IEEEtran}
\bibliography{reference}

\begin{thebibliography}{10}
\providecommand{\url}[1]{#1}
\csname url@samestyle\endcsname
\providecommand{\newblock}{\relax}
\providecommand{\bibinfo}[2]{#2}
\providecommand{\BIBentrySTDinterwordspacing}{\spaceskip=0pt\relax}
\providecommand{\BIBentryALTinterwordstretchfactor}{4}
\providecommand{\BIBentryALTinterwordspacing}{\spaceskip=\fontdimen2\font plus
\BIBentryALTinterwordstretchfactor\fontdimen3\font minus
  \fontdimen4\font\relax}
\providecommand{\BIBforeignlanguage}[2]{{%
\expandafter\ifx\csname l@#1\endcsname\relax
\typeout{** WARNING: IEEEtran.bst: No hyphenation pattern has been}%
\typeout{** loaded for the language `#1'. Using the pattern for}%
\typeout{** the default language instead.}%
\else
\language=\csname l@#1\endcsname
\fi
#2}}
\providecommand{\BIBdecl}{\relax}
\BIBdecl

\bibitem{guo2020backpropagating}
Y.~Guo, Q.~Li, and H.~Chen, ``Backpropagating linearly improves transferability
  of adversarial examples,'' in \emph{NeurIPS}, 2020.

\bibitem{simonyan2015very}
K.~Simonyan and A.~Zisserman, ``Very deep convolutional networks for
  large-scale image recognition,'' in \emph{ICLR}, 2015.

\bibitem{he2016deep}
K.~He, X.~Zhang, S.~Ren, and J.~Sun, ``Deep residual learning for image
  recognition,'' in \emph{CVPR}, 2016.

\bibitem{huang2017densely}
G.~Huang, Z.~Liu, L.~Van Der~Maaten, and K.~Q. Weinberger, ``Densely connected
  convolutional networks,'' in \emph{CVPR}, 2017.

\bibitem{vaswani2017attention}
A.~Vaswani, N.~Shazeer, N.~Parmar, J.~Uszkoreit, L.~Jones, A.~N. Gomez,
  L.~Kaiser, and I.~Polosukhin, ``Attention is all you need,'' \emph{arXiv
  preprint arXiv:1706.03762}, 2017.

\bibitem{dosovitskiy2020image}
A.~Dosovitskiy, L.~Beyer, A.~Kolesnikov, D.~Weissenborn, X.~Zhai,
  T.~Unterthiner, M.~Dehghani, M.~Minderer, G.~Heigold, S.~Gelly \emph{et~al.},
  ``An image is worth 16x16 words: Transformers for image recognition at
  scale,'' in \emph{ICLR}, 2021.

\bibitem{kingma2014adam}
D.~P. Kingma and J.~Ba, ``Adam: A method for stochastic optimization,''
  \emph{arXiv preprint arXiv:1412.6980}, 2014.

\bibitem{loshchilov2018decoupled}
I.~Loshchilov and F.~Hutter, ``Decoupled weight decay regularization,'' in
  \emph{ICLR}, 2019.

\bibitem{loshchilov2016sgdr}
------, ``Sgdr: Stochastic gradient descent with warm restarts,'' in
  \emph{ICLR}, 2017.

\bibitem{reddi2019convergence}
S.~J. Reddi, S.~Kale, and S.~Kumar, ``On the convergence of adam and beyond,''
  \emph{arXiv preprint arXiv:1904.09237}, 2019.

\bibitem{sutskever2013importance}
I.~Sutskever, J.~Martens, G.~Dahl, and G.~Hinton, ``On the importance of
  initialization and momentum in deep learning,'' in \emph{ICML}, 2013.

\bibitem{bottou2010large}
L.~Bottou, ``Large-scale machine learning with stochastic gradient descent,''
  in \emph{Proceedings of COMPSTAT'2010}.\hskip 1em plus 0.5em minus
  0.4em\relax Springer, 2010, pp. 177--186.

\bibitem{lecun1988theoretical}
Y.~LeCun, ``A theoretical framework for back-propagation,'' in
  \emph{Proceedings of the 1988 connectionist models summer school}, vol.~1,
  1988, pp. 21--28.

\bibitem{papernot2017practical}
N.~Papernot, P.~McDaniel, I.~Goodfellow, S.~Jha, Z.~B. Celik, and A.~Swami,
  ``Practical black-box attacks against machine learning,'' in \emph{ACM on
  Asia conference on computer and communications security}, 2017.

\bibitem{sandler2018mobilenetv2}
M.~Sandler, A.~Howard, M.~Zhu, A.~Zhmoginov, and L.-C. Chen, ``Mobilenetv2:
  Inverted residuals and linear bottlenecks,'' in \emph{CVPR}, 2018.

\bibitem{zagoruyko2016wide}
S.~Zagoruyko and N.~Komodakis, ``Wide residual networks,'' \emph{arXiv preprint
  arXiv:1605.07146}, 2016.

\bibitem{goodfellow2015explaining}
I.~J. Goodfellow, J.~Shlens, and C.~Szegedy, ``Explaining and harnessing
  adversarial examples,'' in \emph{ICLR}, 2015.

\bibitem{kurakin2017adversarial}
A.~Kurakin, I.~Goodfellow, and S.~Bengio, ``Adversarial machine learning at
  scale,'' in \emph{ICLR}, 2017.

\bibitem{madry2018deep}
A.~Madry, A.~Makelov, L.~Schmidt, D.~Tsipras, and A.~Vladu, ``Towards deep
  learning models resistant to adversarial attacks,'' in \emph{ICLR}, 2018.

\bibitem{athalye2018obfuscated}
A.~Athalye, N.~Carlini, and D.~Wagner, ``Obfuscated gradients give a false
  sense of security: Circumventing defenses to adversarial examples,'' in
  \emph{ICML}, 2018.

\bibitem{moosavi2016deepfool}
S.-M. Moosavi-Dezfooli, A.~Fawzi, and P.~Frossard, ``Deepfool: a simple and
  accurate method to fool deep neural networks,'' in \emph{CVPR}, 2016.

\bibitem{carlini2017towards}
N.~Carlini and D.~Wagner, ``Towards evaluating the robustness of neural
  networks,'' in \emph{IEEE symposium on security and privacy (SP)}, 2017.

\bibitem{krizhevsky2009learning}
A.~Krizhevsky, G.~Hinton \emph{et~al.}, ``Learning multiple layers of features
  from tiny images,'' 2009.

\bibitem{russakovsky2015imagenet}
O.~Russakovsky, J.~Deng, H.~Su, J.~Krause, S.~Satheesh, S.~Ma, Z.~Huang,
  A.~Karpathy, A.~Khosla, M.~Bernstein \emph{et~al.}, ``Imagenet large scale
  visual recognition challenge,'' \emph{IJCV}, 2015.

\bibitem{bengio2013estimating}
Y.~Bengio, N.~L{\'e}onard, and A.~Courville, ``Estimating or propagating
  gradients through stochastic neurons for conditional computation,''
  \emph{arXiv preprint arXiv:1308.3432}, 2013.

\bibitem{du2019gradient}
S.~Du, J.~Lee, H.~Li, L.~Wang, and X.~Zhai, ``Gradient descent finds global
  minima of deep neural networks,'' in \emph{ICML}, 2019.

\bibitem{arora2019fine}
S.~Arora, S.~Du, W.~Hu, Z.~Li, and R.~Wang, ``Fine-grained analysis of
  optimization and generalization for overparameterized two-layer neural
  networks,'' in \emph{ICML}, 2019.

\bibitem{tian2017analytical}
Y.~Tian, ``An analytical formula of population gradient for two-layered relu
  network and its applications in convergence and critical point analysis,'' in
  \emph{ICML}, 2017.

\bibitem{du2019provably}
S.~S. Du, X.~Zhai, B.~Poczos, and A.~Singh, ``Gradient descent provably
  optimizes over-parameterized neural networks,'' in \emph{ICLR}, 2019.

\bibitem{paszke2019pytorch}
A.~Paszke, S.~Gross, F.~Massa, A.~Lerer, J.~Bradbury, G.~Chanan, T.~Killeen,
  Z.~Lin, N.~Gimelshein, L.~Antiga \emph{et~al.}, ``Pytorch: An imperative
  style, high-performance deep learning library,'' in \emph{NeurIPS}, 2019.

\bibitem{zhang2021geometry}
J.~Zhang, J.~Zhu, G.~Niu, B.~Han, M.~Sugiyama, and M.~Kankanhalli,
  ``Geometry-aware instance-reweighted adversarial training,'' in \emph{ICLR},
  2021.

\bibitem{croce2020robustbench}
F.~Croce, M.~Andriushchenko, V.~Sehwag, E.~Debenedetti, N.~Flammarion,
  M.~Chiang, P.~Mittal, and M.~Hein, ``Robustbench: a standardized adversarial
  robustness benchmark,'' \emph{arXiv preprint arXiv:2010.09670}, 2020.

\bibitem{lecun1998gradient}
Y.~LeCun, L.~Bottou, Y.~Bengio, and P.~Haffner, ``Gradient-based learning
  applied to document recognition,'' \emph{Proceedings of the IEEE}, vol.~86,
  no.~11, pp. 2278--2324, 1998.

\bibitem{croce2020reliable}
F.~Croce and M.~Hein, ``Reliable evaluation of adversarial robustness with an
  ensemble of diverse parameter-free attacks,'' in \emph{International
  conference on machine learning}.\hskip 1em plus 0.5em minus 0.4em\relax PMLR,
  2020, pp. 2206--2216.

\bibitem{li2019exponential}
Z.~Li and S.~Arora, ``An exponential learning rate schedule for deep
  learning,'' \emph{arXiv preprint arXiv:1910.07454}, 2019.

\end{thebibliography}


\begin{thebibliography}{1}
\providecommand{\url}[1]{#1}
\csname url@samestyle\endcsname
\providecommand{\newblock}{\relax}
\providecommand{\bibinfo}[2]{#2}
\providecommand{\BIBentrySTDinterwordspacing}{\spaceskip=0pt\relax}
\providecommand{\BIBentryALTinterwordstretchfactor}{4}
\providecommand{\BIBentryALTinterwordspacing}{\spaceskip=\fontdimen2\font plus
\BIBentryALTinterwordstretchfactor\fontdimen3\font minus
  \fontdimen4\font\relax}
\providecommand{\BIBforeignlanguage}[2]{{%
\expandafter\ifx\csname l@#1\endcsname\relax
\typeout{** WARNING: IEEEtran.bst: No hyphenation pattern has been}%
\typeout{** loaded for the language `#1'. Using the pattern for}%
\typeout{** the default language instead.}%
\else
\language=\csname l@#1\endcsname
\fi
#2}}
\providecommand{\BIBdecl}{\relax}
\BIBdecl

\bibitem{tian2017analytical}
Y.~Tian, ``An analytical formula of population gradient for two-layered relu
  network and its applications in convergence and critical point analysis,'' in
  \emph{ICML}, 2017.

\end{thebibliography}

%\begin{thebibliography}{1}
%\end{thebibliography}


\begin{IEEEbiography}[{\includegraphics[width=1in,height=1.25in,clip,keepaspectratio]{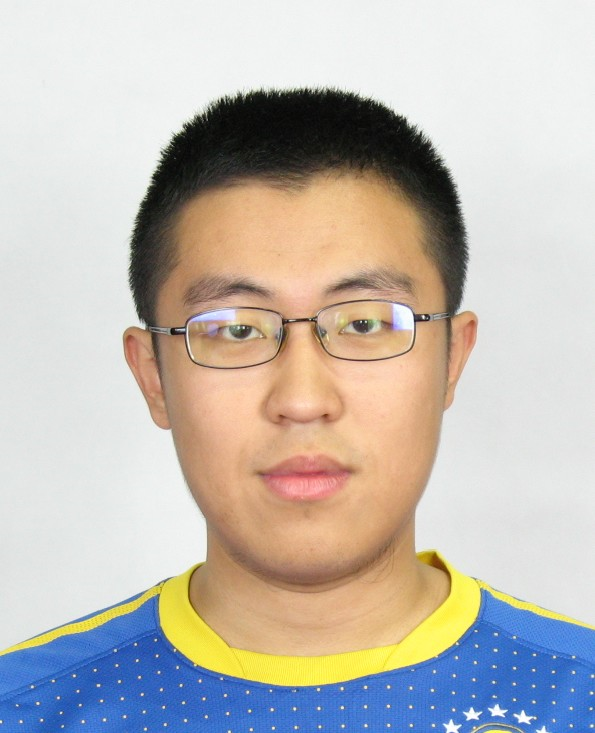}}]{Ziang Li}
received the B.E. degree from Tsinghua University, Beijing, China, in 2019. He is currently working toward the PhD degree with the Department of Automation, Tsinghua University, Beijing, China. His current research interests include pattern recognition and machine learning.
\end{IEEEbiography}

\begin{IEEEbiography}[{\includegraphics[width=1in,height=1.25in,clip,keepaspectratio]{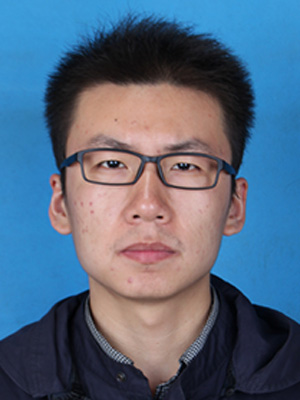}}]{Yiwen Guo}
received the B.E. degree from Wuhan University in 2011, and the Ph.D. degree from Tsinghua University in 2016. He is a research scientist at ByteDance AI Lab, Beijing. Prior to this, he was a staff research scientist at Intel Labs China. His current research interests include computer vision, pattern recognition, and machine learning.
\end{IEEEbiography}

\begin{IEEEbiography}[{\includegraphics[width=1in,height=1.25in,clip,keepaspectratio]{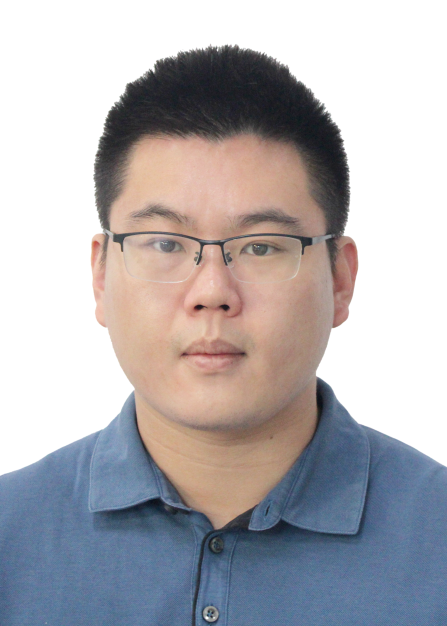}}]{Haodi Liu}
received the B.A. degree from Bard College and B.S. degree from Columbia University in 2018(Dual degree program). And he received the M.S. degree from Columbia University in 2020. Now he is currently working toward the Ph.D. degree with the Department of Automation of Tsinghua University. Prior to this, he was an algorithm engineer at AiBee.Inc. His current research interests include pattern recognition, machine learning and computer vision.
\end{IEEEbiography}

\begin{IEEEbiography}[{\includegraphics[width=1in,height=1.25in,clip,keepaspectratio]{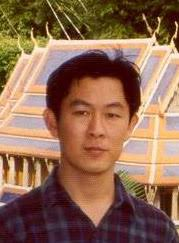}}]{Changshui Zhang}
(M’02-SM’15-F’18) received the B.S. degree in mathematics from Peking University, Beijing, China, in 1986, and the M.S. and Ph.D. degrees in control science and engineering from Tsinghua University, Beijing, in 1989 and 1992, respectively. In 1992, he joined the Department of Automation, Tsinghua University, where he is currently a professor. His current research interests include pattern recognition and machine learning. He has authored more than 200 articles. He is a fellow of the IEEE.
\end{IEEEbiography}

\end{document}

\title{Appendices: A Theoretical View of Linear Backpropagation and Its Convergence}

\author{Ziang~Li*,
        Yiwen~Guo*,
        Haodi~Liu,
        and~Changshui~Zhang,~\IEEEmembership{Fellow,~IEEE}% <-this % stops a space
}

\maketitle

\IEEEdisplaynontitleabstractindextext

% \fi
\IEEEpeerreviewmaketitle



% or


\appendices
\section{Proof of Lemma 1}
We recall the contents of Lemma 1. Assume $D(\mathbf{W},\mathbf{x}) := {\rm diag} (\mathbf{Wx} > 0)$.
\begin{Lemma}
	\label{lemma1_prf}
	Denote $G(\mathbf{e},\mathbf{x}) := \mathbf{W}^TD(\mathbf{W},\mathbf{e})\mathbf{V}^T\mathbf{V}D(\mathbf{W},\mathbf{x})\mathbf{Wx}$, where $\mathbf{e} \in \mathbb{R}^{d_1}$ is a unit vector, $\mathbf{x} \in \mathbb{R}^{d_1}$ is the input data vector, $\mathbf{W}\in\mathbb{R}^{d_2 \times d_1}$ and $\mathbf{V}\in\mathbb{R}^{d_3 \times d_2}$ are weight matrices. If $\mathbf{W}$ and $\mathbf{V}$ are independent and both generated from the standard Gaussian distribution, we have $$\mathbb{E}\left(G(\mathbf{e},\mathbf{x})\right)=\frac{d_2d_3}{2\pi}[(\pi-\Theta)\mathbf{x}+\|\mathbf{x}\|\sin\Theta \mathbf{e}],$$ where $\Theta \in [0,\pi]$ is the angle between $\mathbf{e}$ and $\mathbf{x}$.
\end{Lemma}
\textbf{Proof.} Assume $\mathbf{W}=[\mathbf{w}_1,\cdots,\mathbf{w}_{d_2}]^T$, where $\mathbf{w}_i \in \mathbb{R}^{d_1}$ for $i=1,\cdots,d_2$. 
For $G(\mathbf{e},\mathbf{x})$, we have 
\begin{equation}
    G(\mathbf{e},\mathbf{x}) = \sum_{i:\mathbf{w}_i^T \mathbf{x} \geq 0, \mathbf{w}_i^T \mathbf{e} \geq 0}\sum_{j:\mathbf{w}_j^T \mathbf{x} \geq 0, \mathbf{w}_j^T \mathbf{e} \geq 0}\sum_{d=1}^{d_3}v_{di}v_{dj}\mathbf{w}_i\mathbf{w}_j^T\mathbf{x}.
\end{equation}
We consider the expectation of $G(\mathbf{e},\mathbf{x})$, there is
\begin{equation}
    \label{exp:G}
    \mathbb{E}(G(\mathbf{e},\mathbf{x})) = \sum_{i=1}^{d_2}\sum_{j=1}^{d_2}\mathbb{E}(\sum_{d=1}^{d_3}v_{di}v_{dj})\mathbb{E}(\mathbf{w}_i\mathbf{w}_j^T\cdot\mathbb{I}_{(\mathbf{w}_i^T \mathbf{x} \geq 0, \mathbf{w}_i^T \mathbf{e} \geq 0,\mathbf{w}_j^T \mathbf{x} \geq 0, \mathbf{w}_j^T \mathbf{e} \geq 0)}(\mathbf{x},\mathbf{e})\mathbf{x}),
\end{equation}
where $\mathbb{I}_A(x)$ is the indicator function, \emph{i.e.}, $\mathbb{I}_A(x)$ equals $1$ if $x\in A$ and equals $0$ if $x \notin A$. As the assumption that $\mathbf{W}$ and $\mathbf{V}$ are independent and both generated from the standard Gaussian distribution, we find $\mathbb{E}(\sum_{d=1}^{d_3}v_{di}v_{dj}) = 0$ when $i \neq j$ and $\mathbb{E}(\sum_{d=1}^{d_3}v_{di}v_{dj}) = d_3$ when $i = j$. Therefore, Eq.~(\ref{exp:G}) can be simplified as
\begin{equation}
    \label{exp:G2}
    \mathbb{E}(G(\mathbf{e},\mathbf{x})) = \sum_{i=1}^{d_2}\mathbb{E}(\mathbf{w}_i\mathbf{w}_i^T\cdot\mathbb{I}_{(\mathbf{w}_i^T \mathbf{x} \geq 0, \mathbf{w}_i^T \mathbf{e} \geq 0)}(\mathbf{x},\mathbf{e})\mathbf{x}).
\end{equation}
We then introduce a coordinate system, where $\mathbf{e} = [1,0,\cdots,0]^T$, $\mathbf{x} = \|\mathbf{x}\|[\cos\Theta,\sin\Theta,0,\cdots,0]^T$ are hold. Thus, $\mathbf{w_i} = [r\cos\phi_i,r\sin\phi_i,w_{i,3},\cdots,w_{i,d_1}]^T$. We then can rewrite Eq.~(\ref{exp:G2}) as
\begin{equation}
    \label{exp:G3}
    \begin{aligned}
    \mathbb{E}(G(\mathbf{e},\mathbf{x})) &= d_3\mathbb{E}\sum_{i:\mathbf{w}_i^T \mathbf{x} \geq 0, \mathbf{w}_i^T \mathbf{e} \geq 0}\mathbf{w}_i\mathbf{w}_i^T\mathbf{x}\\
    &=d_3\mathbb{E}\sum_{i:\phi_i\in[-\pi/2+\Theta,\pi/2]}\mathbf{w}_i\mathbf{w}_i^T\mathbf{x}
    \end{aligned}
\end{equation}
We now compute the following equation,
\begin{equation}
\label{R}
\begin{aligned}
R(\phi_0) &= \mathbb{E}\left[\frac{1}{d_2}\sum_{i:\phi_i\in[0,\phi_0]}\mathbf{w}_i\mathbf{w}_i^T\right]=\mathbb{E}\left[\mathbf{w}\mathbf{w}^T|\phi\in[0,\phi_0]\right]\mathbb{P}\left[\phi\in[0,\phi_0]\right]\\
&= \int^{\infty}_{-\infty}\ldots\int^{\phi_0}_{0}\int^{\infty}_{0}\mathbf{w}\mathbf{w}^Tp(r)p(\phi)\prod_{i=3}^{d}p(w_i)rdrd\phi dw_3 \dots dw_{d_1},
\end{aligned}
\end{equation}
where $p(r)=e^{-\frac{r^2}{2}}$ and $p(\phi)=\frac{1}{2\pi}$. Since $\mathbf{W}$ follows gaussian distribution, the off-diagonal and diagonal elements except the first $2\times2$ block are equal to 0 and $\frac{\phi_0}{2\pi}$ respectively. The first $2\times2$ block can be formulated as
\begin{equation}
\label{R22}
\begin{aligned}
R(\phi_0)_{[1:2,1:2]}=&\int^{\phi_0}_{0}\int^{\infty}_{0}\begin{bmatrix}
r\cos\phi\\
r\sin\phi
\end{bmatrix}\begin{bmatrix}
r\cos\phi&r\sin\phi
\end{bmatrix}p(r)p(\phi)rdrd\phi\\
&=\int^{\infty}_{0}\frac{r^3e^{-\frac{r^2}{2}}}{2\pi}dr\int^{\phi_0}_{0}\begin{bmatrix}
\cos^2\phi&\cos\phi\sin\phi\\
\cos\phi\sin\phi&\sin^2\phi\\
\end{bmatrix}d\phi.\\
&=\frac{1}{4\pi}\begin{bmatrix}
2+\sin2\phi_0&1-\cos2\phi_0\\
1-\cos2\phi_0&2-\sin2\phi_0\\
\end{bmatrix}
\end{aligned}
\end{equation}
Further we have
\begin{equation}
\label{R_res}
\begin{aligned}
R(\phi_0) = \frac{1}{2\pi}\mathbf{I}_d+\frac{1}{4\pi}\begin{bmatrix}
\sin2\phi_0&1-\cos2\phi_0&\mathbf{0}\\
1-\cos2\phi_0&-\sin2\phi_0&\mathbf{0}\\
\mathbf{0} & \mathbf{0} & \mathbf{0}
\end{bmatrix}
\end{aligned}
\end{equation}
Then Eq.~(\ref{exp:G3}) can be formulated as
\begin{equation}
\label{F_res}
\begin{aligned}
\mathbb{E}(G(\mathbf{e},\mathbf{x}))&=d_2d_3\left(R(\frac{\pi}{2})-R(-\frac{\pi}{2}+\Theta)\right)x\\
&=d_2d_3\frac{\pi-\Theta}{2\pi}\mathbf{I}_d+\frac{d_2d_3}{4\pi}(\begin{bmatrix}
0&2\\
2&0\\
\end{bmatrix}-\begin{bmatrix}
\sin(2\Theta-\pi)&1-\cos(2\Theta-\pi)\\
1-\cos(2\Theta-\pi)&-\sin(2\Theta-\pi)
\end{bmatrix})\|x\|\begin{bmatrix}
\cos\Theta\\
\sin\Theta\\
\end{bmatrix}\\
&=d_2d_3\frac{\pi-\Theta}{2\pi}x+\frac{d_2d_3\|x\|}{4\pi}\begin{bmatrix}
2\sin\Theta\\
0
\end{bmatrix}\\
&=\frac{d_2d_3}{2\pi}[(\pi-\Theta)\mathbf{x}+\|\mathbf{x}\|\sin\Theta \mathbf{e}]
\end{aligned}
\end{equation}

\section{Lemma for proving Theorem 1 and Theorem 2}
We here propose a lemma before we delve deep into Theorem 1 and Theorem 2. The lemma is described as follows,
\begin{Lemma}
    Let $\alpha_i$ define as $$\alpha_i = (u \cdot \mathbf{x}^\star_i- v \cdot \mathbf{x}_i)\text{sgn}(\mathbf{x}^\star_i-\mathbf{x}_i) = \left\{
	\begin{aligned}
	&u \cdot \mathbf{x}^\star_i- v \cdot \mathbf{x}_i,\quad if \quad \mathbf{x}^\star_i>\mathbf{x}_i,\\
	&v \cdot \mathbf{x}_i - u \cdot \mathbf{x}^\star_i,\quad if \quad \mathbf{x}^\star_i<\mathbf{x}_i,
	\end{aligned}
	\right.$$
	where $u$ and $v$ are constants, $\mathbf{x}^\star_i \sim N(\mu_1,\sigma_1^2)$ and $\mathbf{x}_i \sim N(\mu_2,\sigma_2^2)$. The expectation of $\alpha_i$ can be formulated as $$
	\mathbb{E}(\alpha_i) = 2\gamma (u\cdot\sigma_1^2+v\cdot\sigma_2^2) + (u\cdot\mu_1 - v\cdot\mu_2)(2\mathbf{P}(\mathbf{x}_i < \mathbf{x}^\star_i) - 1),
	$$
	where $\gamma = \frac{1}{\sqrt{2\pi(\sigma_1^2+\sigma_2^2})}e^{-(\mu_1-\mu_2)^2/2(\sigma_1^2+\sigma_2^2)} > 0.$ Further we have $\mathbb{E}(\alpha_i)>0$ when $u>0$, $v>0$ and $\mu_2=0$.\\
\end{Lemma}
\textbf{Proof.}
From the definition of $\alpha_i$, the expectation of $\alpha_i$ can be formulated as
\begin{equation}
\label{e_alpha}
\mathbb{E}(\alpha_i) = \mathbb{E}(u \cdot \mathbf{x}^\star_i- v \cdot \mathbf{x}_i | \mathbf{x}^\star_i>\mathbf{x}_i)\mathbf{P}(\mathbf{x}^\star_i>\mathbf{x}_i) + \mathbb{E}(v \cdot \mathbf{x}_i - u \cdot \mathbf{x}^\star_i| \mathbf{x}^\star_i<\mathbf{x}_i)\mathbf{P}(\mathbf{x}^\star_i<\mathbf{x}_i).
\end{equation}
Further, we have $\mathbf{x}^\star_i - \mathbf{x}_i \sim N(\mu_1-\mu_2,\sigma_1^2+\sigma_2^2)$. Therefore, we can conclude that $\mathbf{P}(\mathbf{x}^\star_i>\mathbf{x}_i) = 1 - F(0)$ and $\mathbf{P}(\mathbf{x}^\star_i<\mathbf{x}_i) = F(0)$, where $F(\cdot)$ denotes the cumulative distribution function for $N(\mu_1-\mu_2,\sigma_1^2+\sigma_2^2)$.

We then solve the conditional cumulative distribution function for $\mathbf{x}^\star_i$ when $\mathbf{x}^\star_i>\mathbf{x}_i$. We define $f_1(\cdot)$, $f_2(\cdot)$ denote the probability density function of $\mathbf{x}^\star_i$ and $\mathbf{x}_i$ and $F_1(\cdot)$, $F_2(\cdot)$ denote the cumulative distribution function of $\mathbf{x}^\star_i$ and $\mathbf{x}_i$. There we have
\begin{equation}
\label{con_cdf}
\begin{aligned}
G_1(x) &= \mathbf{P}(\mathbf{x}^\star_i \leq x | \mathbf{x}^\star_i>\mathbf{x}_i)\\
&= \frac{\mathbf{P}(\mathbf{x}_i < \mathbf{x}^\star_i \leq x)}{\mathbf{P}(\mathbf{x}_i < \mathbf{x}^\star_i)}\\
&= \frac{1}{\mathbf{P}(\mathbf{x}_i < \mathbf{x}^\star_i)}\int_{-\infty}^{x}\mathbf{P}(y < \mathbf{x}^\star_i \leq x)f_2(y)dy\\
&= \frac{1}{\mathbf{P}(\mathbf{x}_i < \mathbf{x}^\star_i)}\int_{-\infty}^{x}(F_1(x)-F_1(y))f_2(y)dy.\\
\end{aligned}
\end{equation}
Therefore, the probability density function can be formulated as
\begin{equation}
\label{con_pdf}
\begin{aligned}
g_1(x) &= \frac{1}{\mathbf{P}(\mathbf{x}_i < \mathbf{x}^\star_i)}\left[(F_1(x)\int_{-\infty}^{x}f_2(y)dy)'-(\int_{-\infty}^{x}(F_1(y)f_2(y)dy)'\right]\\
&= \frac{1}{\mathbf{P}(\mathbf{x}_i < \mathbf{x}^\star_i)}\left[(f_1(x)\int_{-\infty}^{x}f_2(y)dy) + F_1(x)f_2(x)-F_1(x)f_2(x)\right]\\
&=\frac{f_1(x)F_2(x)}{\mathbf{P}(\mathbf{x}_i < \mathbf{x}^\star_i)}.
\end{aligned}
\end{equation}
Similarly, we first solve the conditional cumulative distribution function for $\mathbf{x}_i$ when $\mathbf{x}^\star_i>\mathbf{x}_i$. There we have,
\begin{equation}
\label{con_cdf2}
\begin{aligned}
G_2(x) &= 1 - \mathbf{P}(\mathbf{x}_i > x | \mathbf{x}^\star_i>\mathbf{x}_i)\\
&=1- \frac{\mathbf{P}(x < \mathbf{x}_i < \mathbf{x}^\star_i)}{\mathbf{P}(\mathbf{x}_i < \mathbf{x}^\star_i)}\\
&=1- \frac{1}{\mathbf{P}(\mathbf{x}_i < \mathbf{x}^\star_i)}\int_{x}^{\infty}\mathbf{P}(x < \mathbf{x}_i < y)f_1(y)dy\\
&= 1-\frac{1}{\mathbf{P}(\mathbf{x}_i < \mathbf{x}^\star_i)}\int_{x}^{\infty}(F_2(y)-F_2(x))f_1(y)dy.\\
\end{aligned}
\end{equation}
The probability density function can be formulated as
\begin{equation}
\label{con_pdf2}
\begin{aligned}
g_2(x) &= \frac{1}{\mathbf{P}(\mathbf{x}_i < \mathbf{x}^\star_i)}\left[(F_2(x)\int_{x}^{\infty}f_1(y)dy)'-(\int_{x}^{\infty}(F_2(y)f_1(y)dy)'\right]\\
&= \frac{1}{\mathbf{P}(\mathbf{x}_i < \mathbf{x}^\star_i)}\left[(f_2(x)\int_{x}^{\infty}f_1(y)dy) - F_2(x)f_1(x)+F_2(x)f_1(x)\right]\\
&=\frac{f_2(x)(1-F_1(x))}{\mathbf{P}(\mathbf{x}_i < \mathbf{x}^\star_i)}.
\end{aligned}
\end{equation}
We assume $\phi(\cdot)$ and $\Phi(\cdot)$ represent the probability density function and cumulative distribution function for standard gaussian distribution, respectively. There we have
\begin{equation}
\label{exp1}
\begin{aligned}
\mathbb{E}(\mathbf{x}^\star_i|\mathbf{x}^\star_i>\mathbf{x}_i) &= \int_{-\infty}^{\infty}xg_1(x)dx\\
&= \frac{1}{\mathbf{P}(\mathbf{x}_i < \mathbf{x}^\star_i)}\int_{-\infty}^{\infty}\frac{x}{\sigma_1}\phi(\frac{x-\mu_1}{\sigma_1})\Phi(\frac{x-\mu_2}{\sigma_2})dx\\
&=\frac{1}{\mathbf{P}(\mathbf{x}_i < \mathbf{x}^\star_i)}\int_{-\infty}^{\infty}\frac{x}{\sqrt{2\pi}\sigma_1}e^{-(x-\mu_1)^2/2\sigma_1^2}\Phi(\frac{x-\mu_2}{\sigma_2})dx\\
&=\frac{1}{\mathbf{P}(\mathbf{x}_i < \mathbf{x}^\star_i)}\int_{-\infty}^{\infty}\left[\frac{x-\mu_1}{\sqrt{2\pi}\sigma_1}+\frac{\mu_1}{\sqrt{2\pi}\sigma_1}\right]e^{-(x-\mu_1)^2/2\sigma_1^2}\Phi(\frac{x-\mu_2}{\sigma_2})dx
\end{aligned}
\end{equation}
The integral can be devided into two part. The first part can be formulated as
\begin{equation}
\label{exp11}
\begin{aligned}
\int_{-\infty}^{\infty}\frac{x-\mu_1}{\sqrt{2\pi}\sigma_1}e^{-(x-\mu_1)^2/2\sigma_1^2}\Phi(\frac{x-\mu_2}{\sigma_2})dx &= \frac{\sigma_1}{\sqrt{2\pi}}\int_{-\infty}^{\infty}\Phi(\frac{x-\mu_2}{\sigma_2})\frac{d(-e^{-(x-\mu_1)^2/2\sigma_1^2})}{dx}dx\\
&=\frac{\sigma_1}{\sqrt{2\pi}}\int_{-\infty}^{\infty}e^{-(x-\mu_1)^2/2\sigma_1^2}\frac{d\Phi(\frac{x-\mu_2}{\sigma_2})}{dx}dx\\
&=\frac{\sigma_1}{2\sigma_2\pi}\int_{-\infty}^{\infty}e^{-(x-\mu_1)^2/2\sigma_1^2-(x-\mu_2)^2/2\sigma_2^2}dx\\
&=\frac{\sigma_1}{2\sigma_2\pi}e^{-(\mu_1-\mu_2)^2/2(\sigma_1^2+\sigma_2^2)}\frac{\sigma_1\sigma_2\sqrt{2\pi}}{\sqrt{\sigma_1^2+\sigma_2^2}}\\
&=\frac{\sigma_1^2}{\sqrt{2\pi(\sigma_1^2+\sigma_2^2})}e^{-(\mu_1-\mu_2)^2/2(\sigma_1^2+\sigma_2^2)}
\end{aligned}
\end{equation}
Step 3 equals Step 4 because there is $\int_{-\infty}^{\infty}e^{-(ax^2+bx+c)}dx = e^{(b^2-4ac)/4a}\sqrt{\frac{\pi}{a}}$.
The second part can be formulated as
\begin{equation}
\label{exp12}
\begin{aligned}
\int_{-\infty}^{\infty}\frac{\mu_1}{\sqrt{2\pi}\sigma_1}e^{-(x-\mu_1)^2/2\sigma_1^2}\Phi(\frac{x-\mu_2}{\sigma_2})dx = \mu_1\int_{-\infty}^{\infty}f_1(x)F_2(x)dx
\end{aligned}
\end{equation}
From Eq.~(\ref{con_pdf}) we have
\begin{equation}
\int_{-\infty}^{\infty}g_1(x)dx = \frac{\int_{-\infty}^{\infty}f_1(x)F_2(x)dx}{\mathbf{P}(\mathbf{x}_i < \mathbf{x}^\star_i)}=1
\end{equation}
Therefore we have
\begin{equation}
\label{exp13}
\begin{aligned}
\int_{-\infty}^{\infty}\frac{\mu_1}{\sqrt{2\pi}\sigma_1}e^{-(x-\mu_1)^2/2\sigma_1^2}\Phi(\frac{x-\mu_2}{\sigma_2})dx = \mu_1\int_{-\infty}^{\infty}f_1(x)F_2(x)dx = \mu_1\mathbf{P}(\mathbf{x}_i < \mathbf{x}^\star_i)
\end{aligned}
\end{equation}
As the result in Eq.~(\ref{exp11}) and Eq.~(\ref{exp13}), Eq.~(\ref{exp1}) can be formulated as
\begin{equation}
\label{e1}
\begin{aligned}
\mathbb{E}(\mathbf{x}^\star_i|\mathbf{x}^\star_i>\mathbf{x}_i) &= \frac{\frac{\sigma_1^2}{\sqrt{2\pi(\sigma_1^2+\sigma_2^2})}e^{-(\mu_1-\mu_2)^2/2(\sigma_1^2+\sigma_2^2)} }{\mathbf{P}(\mathbf{x}_i < \mathbf{x}^\star_i)}+ \mu_1\\
&= \frac{\gamma\sigma_1^2}{\mathbf{P}(\mathbf{x}_i < \mathbf{x}^\star_i)}  + \mu_1,
\end{aligned}
\end{equation}
where $\gamma = \frac{1}{\sqrt{2\pi(\sigma_1^2+\sigma_2^2})}e^{-(\mu_1-\mu_2)^2/2(\sigma_1^2+\sigma_2^2)} > 0$.

Similarly, we have
\begin{equation}
\label{exp3}
\begin{aligned}
\mathbb{E}(\mathbf{x}_i|\mathbf{x}^\star_i>\mathbf{x}_i) &= \int_{-\infty}^{\infty}xg_2(x)dx\\
&= \frac{1}{\mathbf{P}(\mathbf{x}_i < \mathbf{x}^\star_i)}\int_{-\infty}^{\infty}\frac{x}{\sigma_2}\phi(\frac{x-\mu_2}{\sigma_2})(1-\Phi(\frac{x-\mu_1}{\sigma_1}))dx\\
&=\frac{1}{\mathbf{P}(\mathbf{x}_i < \mathbf{x}^\star_i)}\int_{-\infty}^{\infty}\frac{x}{\sqrt{2\pi}\sigma_2}e^{-(x-\mu_2)^2/2\sigma_2^2}(1-\Phi(\frac{x-\mu_1}{\sigma_1}))dx\\
&=\frac{1}{\mathbf{P}(\mathbf{x}_i < \mathbf{x}^\star_i)}\int_{-\infty}^{\infty}\left[\frac{x-\mu_2}{\sqrt{2\pi}\sigma_2}+\frac{\mu_2}{\sqrt{2\pi}\sigma_2}\right]e^{-(x-\mu_2)^2/2\sigma_2^2}(1-\Phi(\frac{x-\mu_1}{\sigma_1}))dx
\end{aligned}
\end{equation}
The integral can also be devided into two part. The first part can be formulated as
\begin{equation}
\label{exp21}
\begin{aligned}
&\int_{-\infty}^{\infty}\frac{x-\mu_2}{\sqrt{2\pi}\sigma_2}e^{-(x-\mu_2)^2/2\sigma_2^2}(1-\Phi(\frac{x-\mu_1}{\sigma_1}))dx \\
&=\frac{\sigma_2}{\sqrt{2\pi}}\int_{-\infty}^{\infty}[1-\Phi(\frac{x-\mu_1}{\sigma_1})]\frac{d(-e^{-(x-\mu_2)^2/2\sigma_2^2})}{dx}dx\\
&=\frac{\sigma_2}{\sqrt{2\pi}}\int_{-\infty}^{\infty}e^{-(x-\mu_2)^2/2\sigma_2^2}\frac{d(1-\Phi(\frac{x-\mu_1}{\sigma_1}))}{dx}dx\\
&=\frac{-\sigma_2}{2\sigma_1\pi}\int_{-\infty}^{\infty}e^{-(x-\mu_1)^2/2\sigma_1^2-(x-\mu_2)^2/2\sigma_2^2}dx\\
&=\frac{-\sigma_2}{2\sigma_1\pi}e^{-(\mu_1-\mu_2)^2/2(\sigma_1^2+\sigma_2^2)}\frac{\sigma_1\sigma_2\sqrt{2\pi}}{\sqrt{\sigma_1^2+\sigma_2^2}}\\
&=\frac{-\sigma_2^2}{\sqrt{2\pi(\sigma_1^2+\sigma_2^2})}e^{-(\mu_1-\mu_2)^2/2(\sigma_1^2+\sigma_2^2)}
\end{aligned}
\end{equation}
The second part can be formulated as
\begin{equation}
\label{exp22}
\begin{aligned}
\int_{-\infty}^{\infty}\frac{\mu_2}{\sqrt{2\pi}\sigma_2}e^{-(x-\mu_2)^2/2\sigma_2^2}(1-\Phi(\frac{x-\mu_1}{\sigma_1}))dx = \mu_2\int_{-\infty}^{\infty}f_2(x)(1-F_1(x))dx
\end{aligned}
\end{equation}
From Eq.~(\ref{con_pdf2}) we have
\begin{equation}
\int_{-\infty}^{\infty}g_2(x)dx = \frac{\int_{-\infty}^{\infty}f_2(x)(1-F_1(x))dx}{\mathbf{P}(\mathbf{x}_i < \mathbf{x}^\star_i)}=1
\end{equation}
Therefore we have
\begin{equation}
\label{exp23}
\begin{aligned}
\int_{-\infty}^{\infty}\frac{\mu_2}{\sqrt{2\pi}\sigma_2}e^{-(x-\mu_2)^2/2\sigma_2^2}(1-\Phi(\frac{x-\mu_1}{\sigma_1}))dx &= \mu_2\int_{-\infty}^{\infty}f_2(x)(1-F_1(x))dx\\&=\mu_2\mathbf{P}(w_i < w^\star_i)
\end{aligned}
\end{equation}
Therefore, Eq.~(\ref{exp3}) can be formulated as
\begin{equation}
\begin{aligned}
\mathbb{E}(\mathbf{x}_i|\mathbf{x}^\star_i>\mathbf{x}_i)&=\frac{\frac{-\sigma_2^2}{\sqrt{2\pi(\sigma_1^2+\sigma_2^2})}e^{-(\mu_1-\mu_2)^2/2(\sigma_1^2+\sigma_2^2)}}{\mathbf{P}(\mathbf{x}_i < \mathbf{x}^\star_i)} + \mu_2\\
&= \frac{-\gamma\sigma_2^2}{\mathbf{P}(\mathbf{x}_i < \mathbf{x}^\star_i)}  + \mu_2.
\end{aligned}
\end{equation}
Due to symmetry, we also have
$$\mathbb{E}(\mathbf{x}_i|\mathbf{x}^\star_i<\mathbf{x}_i) = \frac{\gamma\sigma_2^2}{\mathbf{P}(\mathbf{x}_i > \mathbf{x}^\star_i)} + \mu_2,$$
and
$$\mathbb{E}(\mathbf{x}^\star_i|\mathbf{x}^\star_i<\mathbf{x}_i) = \frac{-\gamma\sigma_1^2}{\mathbf{P}(\mathbf{x}_i > \mathbf{x}^\star_i)} + \mu_1.$$
Therefore, Eq.~(\ref{e_alpha}) can be formulated as
\begin{equation}
\label{e_}
\begin{aligned}
\mathbb{E}(\alpha_i) = &\gamma (u\cdot\sigma_1^2+v\cdot\sigma_2^2) + (u\cdot\mu_1 - v\cdot\mu_2)\mathbf{P}(\mathbf{x}_i < \mathbf{x}^\star_i)\\&+ \gamma (v\cdot\sigma_2^2+u\cdot\sigma_1^2) + (v\cdot\mu_2-u\cdot\mu_1)\mathbf{P}(\mathbf{x}_i > \mathbf{x}^\star_i)\\
=&2\gamma (u\cdot\sigma_1^2+v\cdot\sigma_2^2) + (u\cdot\mu_1 - v\cdot\mu_2)(2\mathbf{P}(\mathbf{x}_i < \mathbf{x}^\star_i) - 1),
\end{aligned}
\end{equation}
where $\gamma = \frac{1}{\sqrt{2\pi(\sigma_1^2+\sigma_2^2})}e^{-(\mu_1-\mu_2)^2/2(\sigma_1^2+\sigma_2^2)} > 0.$ When $\mu_2=0$, we have
\begin{equation}
\label{ee_}
\begin{aligned}
\mathbb{E}(\alpha_i) = 2\gamma (u\cdot\sigma_1^2+v\cdot\sigma_2^2) + u\mu_1(2\mathbf{P}(\mathbf{x}_i < \mathbf{x}^\star_i) - 1).
\end{aligned}
\end{equation}
If $u>0$ and $v>0$, we have $\gamma (u\cdot\sigma_1^2+v\cdot\sigma_2^2)>0$, and
\begin{equation}
\label{sign_1}
u\mu_1(2\mathbf{P}(\mathbf{x}_i < \mathbf{x}^\star_i) - 1)=u\mu_1(1-2F(0)),
\end{equation}
where $F$ is the cumulative distribution function for $\mathbf{x}^\star_i-\mathbf{x}_i$, and follows the gaussian distribution $N(\mu_1,\sigma_1^2+\sigma_2^2)$. If $\mu_1>0$, we have $F(0)<0.5$, and Eq.~(\ref{sign_1}) is greater than 0. If $\mu_1<0$, we have $F(0)>0.5$, and Eq.~(\ref{sign_1}) is also greater than 0. Therefore, we have $u\mu_1(1-2F(0))\geq 0$. In sum, we have $\mathbb{E}(\alpha_i)>0$ when $u>0$, $v>0$ and $\mu_2=0$.

\section{Proof of Theorem 1}
We first recall the framework of the two-layer network, which can be formulated as
\begin{equation}
\label{attack_forward}
g(\mathbf{W},\mathbf{V},\mathbf{x}) = \mathbf{V}\sigma(\mathbf{W}\mathbf{x}),
\end{equation}
where $\mathbf{x}\in\mathbb{R}^{d_1}$ is the input data vector, $\mathbf{W}\in\mathbb{R}^{d_2 \times d_1}$ and $\mathbf{V}\in\mathbb{R}^{d_3 \times d_2}$ are weight matrices, $\sigma(\cdot)$ is the ReLU function. In the teacher-student frameworks, we assume the teacher network possesses the optimal adversarial example $\mathbf{x}^\star$, and the student network learns the adversarial example $\mathbf{x}$ from the teacher network, in which the loss function is set as
\begin{equation}
\label{loss_attack}
\mathcal{L}(\mathbf{x}) = \frac{1}{2}\|g(\mathbf{W},\mathbf{V},\mathbf{x})-g(\mathbf{W},\mathbf{V},\mathbf{x}^\star)\|_2^2.
\end{equation} 
The update rules is formulated as
\begin{equation}
\label{update_x}
\mathbf{x}^{(t+1)}=\text{Clip}(\mathbf{x}^{(t)} - \eta \nabla_{\mathbf{x}^{(t)}}\mathcal{L}(\mathbf{x}^{(t)})),
\end{equation}
where $\text{Clip}(\cdot) = \min(\mathbf x + \epsilon\mathbf 1,\max(\mathbf x - \epsilon\mathbf 1,\cdot))$ and we use the update of standard BP and LinBP, \emph{i.e.}, $\nabla_{\mathbf{x}}\mathcal{L}(\mathbf{x})$ and $\tilde{\nabla}_{\mathbf{x}}\mathcal{L}(\mathbf{x})$, to obtain $\{\mathbf{x}^{(t)}\}$ and $\{\tilde{\mathbf{x}}^{(t)}\}$, respectively. The expectation of the gradient obtained by BP and LinBP can be computed from Lemma 1, \emph{i.e.}, 
\begin{equation}
\label{ex4}
\mathbb{E}[\nabla_{\mathbf{x}}\mathcal{L}(\mathbf{x})] = G(\mathbf{x}/\|\mathbf{x}\|,\mathbf{x}) - G(\mathbf{x}/\|\mathbf{x}\|,\mathbf{x}^\star)=\frac{d_2d_3}{2}(\mathbf{x}-\mathbf{x}^\star)+\frac{d_2d_3}{2\pi}\left(\Theta \mathbf{x}^\star-\frac{\|\mathbf{x}^\star\|}{\|\mathbf{x}\|}\sin\Theta \mathbf{x}\right),
\end{equation}
and 
\begin{equation}
\label{ex3}
\mathbb{E}[\tilde{\nabla}_{\mathbf{x}}\mathcal{L}(\mathbf{x})] = G(\mathbf{x}/\|\mathbf{x}\|,\mathbf{x}) - G(\mathbf{x}^\star/\|\mathbf{x}^\star\|,\mathbf{x}^\star)
=\frac{d_2d_3}{2}(\mathbf{x}-\mathbf{x}^\star),
\end{equation}
respectively. Note $\Theta \in [0,\pi]$ is the angle between $\mathbf{x}$ and $\mathbf{x^\star}$. Then we begin our proof.
\begin{Theorem}
	For the two-layer teacher-student network formulated as Eq.~(\ref{attack_forward}), the adversarial attack sets Eq.~(\ref{loss_attack}) and Eq.~(\ref{update_x}) as the loss function and the update rule, respectively. Assume that $\mathbf{W}$ and $\mathbf{V}$ follow are independent and both generated from the standard Gaussian distribution, $\mathbf{x}^\star \sim N(\mu_1,\sigma_1^2)$, $\mathbf{x}^{(0)} \sim N(0,\sigma_2^2)$, and $\eta$ is reasonably small~\footnote{See Eq.~(\ref{eq:constraint_2_}) for more details of the constraint.}. Let $\mathbf{x}^{(t)}$ and $\tilde{\mathbf{x}}^{(t)}$ be the adversarial examples generated in the $t$-th iteration of attack using BP and LinBP, respectively, then we have $$\mathbb{E}\|\mathbf{x}^\star - \tilde{\mathbf{x}}^{(t)}\|_1 \leq \mathbb{E}\|\mathbf{x}^\star - \mathbf{x}^{(t)}\|_1.$$
\end{Theorem}
\textbf{Proof.} 
The update rules for BP and LinBP are formulated as
\begin{equation}
\label{attack_gra}
\mathbf{x}^{(t+1)}=\text{Clip}(\mathbf{x}^{(t)} - \eta \nabla_{\mathbf{x}^{(t)}}\mathcal{L}(\mathbf{x}^{(t)})),
\end{equation}
and
\begin{equation}
\label{attack_gra2}
\tilde{\mathbf{x}}^{(t+1)}=\text{Clip}(\tilde{\mathbf{x}}^{(t)} - \eta \tilde{\nabla}_{\tilde{\mathbf{x}}}^{(t)}\mathcal{L}(\tilde{\mathbf{x}}^{(t)})),
\end{equation}
respectively. We first analyse the property of the Eq.~(\ref{attack_gra2}). Assume $\tilde{\mathbf{x}}^{(v)}$ is the first $\{\tilde{\mathbf{x}}^{(t)}\}$ to satisfy $|\tilde{\mathbf{x}}^{(v)}_i-\mathbf{x}^{(0)}_i|=\epsilon,$ which also means $|\tilde{\mathbf{x}}^{(v-1)}_i-\mathbf{x}^{(0)}_i|<\epsilon$. If $\tilde{\mathbf{x}}^{(v)}_i=\mathbf{x}^{(0)}_i+\epsilon$, from Eq.~(\ref{attack_gra2}), we have
\begin{equation}
\begin{aligned}\tilde{\mathbf{x}}^{(v)}_i&=\text{Clip}(\tilde{\mathbf{x}}^{(v-1)}_i - \frac{\eta d_2d_3}{2}(\tilde{\mathbf{x}}^{(v-1)}_i - \tilde{\mathbf{x}}^\star_i))\\
&=\text{Clip}((1 - \frac{\eta d_2d_3}{2})\tilde{\mathbf{x}}^{(v-1)}_i + \frac{\eta d_2d_3}{2}\tilde{\mathbf{x}}^\star_i)\\
&= \mathbf{x}^{(0)}_i+\epsilon.
\end{aligned}
\end{equation}
As $|\tilde{\mathbf{x}}^{(v-1)}_i-\mathbf{x}^{(0)}_i|<\epsilon$, we have $\tilde{\mathbf{x}}^\star_i>\mathbf{x}^{(0)}_i+\epsilon$. Therefore, for the $v+1$-th step, we have
\begin{equation}
\begin{aligned}\tilde{\mathbf{x}}^{(v+1)}_i&=\text{Clip}(\tilde{\mathbf{x}}^{(v)}_i - \frac{\eta d_2d_3}{2}(\tilde{\mathbf{x}}^{(v)}_i - \tilde{\mathbf{x}}^\star_i))\\
&=\text{Clip}((1 - \frac{\eta d_2d_3}{2})\tilde{\mathbf{x}}^{(v)}_i + \frac{\eta d_2d_3}{2}\tilde{\mathbf{x}}^\star_i)\\
&= \text{Clip}((1 - \frac{\eta d_2d_3}{2})(\mathbf{x}^{(0)}_i+\epsilon) + \frac{\eta d_2d_3}{2}\tilde{\mathbf{x}}^\star_i)\\
&= \mathbf{x}^{(0)}_i+\epsilon.
\end{aligned}
\end{equation}
Note that the final step is established because $(1 - \frac{\eta d_2d_3}{2})(\mathbf{x}^{(0)}_i+\epsilon) + \frac{\eta d_2d_3}{2}\tilde{\mathbf{x}}^\star_i > \mathbf{x}^{(0)}_i+\epsilon$. Further we have $\tilde{\mathbf{x}}^{(t)}_i=\mathbf{x}^{(0)}_i+\epsilon$, for $\forall t>v$. If $\tilde{\mathbf{x}}^{(v)}_i=\mathbf{x}^{(0)}_i-\epsilon$, from Eq.~(\ref{attack_gra2}), we have
\begin{equation}
\begin{aligned}\tilde{\mathbf{x}}^{(v)}_i&=\text{Clip}((1 - \frac{\eta d_2d_3}{2})\tilde{\mathbf{x}}^{(v-1)}_i + \frac{\eta d_2d_3}{2}\tilde{\mathbf{x}}^\star_i)\\
&= \mathbf{x}^{(0)}_i-\epsilon.
\end{aligned}
\end{equation}
As $|\tilde{\mathbf{x}}^{(v-1)}_i-\mathbf{x}^{(0)}_i|<\epsilon$, we have $\tilde{\mathbf{x}}^\star<\mathbf{x}^{(0)}-\epsilon$. Therefore, for the $v+1$-th step, we have
\begin{equation}
\begin{aligned}\tilde{\mathbf{x}}^{(v+1)}_i&=\text{Clip}((1 - \frac{\eta d_2d_3}{2})\tilde{\mathbf{x}}^{(v)}_i + \frac{\eta d_2d_3}{2}\tilde{\mathbf{x}}^\star_i)\\
&= \text{Clip}((1 - \frac{\eta d_2d_3}{2})(\mathbf{x}^{(0)}_i-\epsilon) + \frac{\eta d_2d_3}{2}\tilde{\mathbf{x}}^\star_i)\\
&= \mathbf{x}^{(0)}_i-\epsilon.
\end{aligned}
\end{equation}
Then we have for $\forall t>v$, $\tilde{\mathbf{x}}^{(t)}_i=\mathbf{x}^{(0)}_i-\epsilon$. In sum, for $\forall t>v$, we have $|\tilde{\mathbf{x}}^{(t)}_i-\mathbf{x}^{(0)}_i|=\epsilon$. Similar conclusion can be for made for Eq.~(\ref{attack_gra}), further we can find that if a $\mathbf{x}^{(t)}$ (or $\tilde{\mathbf{x}}^{(t)}$) achieve the bound of $\text{Clip}$, the following steps will keep the bounded value. We then let $\mathbf{p}_j=\theta_j\mathbf{x}^\star-\frac{\|\mathbf{x}^\star\|}{\|\mathbf{x}^{(j)}\|}\sin\theta_j\mathbf{x}^{(j)}$, the $l_1$ distance bewteen $\mathbf{x}^{(t)}$ ($\tilde{\mathbf{x}}^{(t)}$) and $\mathbf{x}^\star$ can be formulated as
\begin{equation}
\label{norm5}
\begin{aligned}
	\mathbb{E}\|\mathbf{x}^\star - \mathbf{x}^{(t+1)}\|_1 &= \mathbb{E}\|\mathbf{x}^\star - \text{Clip}(\mathbf{x}^{(t)} - \eta \nabla_{\mathbf{x}^{(t)}}\mathcal{L}(\mathbf{x}^{(t)}))\|_1\\
	&=\mathbb{E}\|H\left((1-\frac{\eta d_2d_3}{2})(\mathbf{x}^\star - \mathbf{x}^{(t)}) + \frac{\eta d_2d_3}{2\pi}\mathbf{p}_t\right)\|_1\\
	&=\mathbb{E}\|H\left((1-\frac{\eta d_2d_3}{2})^{t+1}(\mathbf{x}^\star - \mathbf{x}^{(0)}) + \sum\limits_{j=0}^{t}\frac{\eta d_2d_3}{2\pi}(1-\frac{\eta d_2d_3}{2})^{t-j}\mathbf{p}_{j}\right)\|_1,
\end{aligned} 
\end{equation}
and
\begin{equation}
\label{norm6}
\begin{aligned}
\mathbb{E}\|\mathbf{x}^\star - \tilde{\mathbf{x}}^{(t+1)}\|_1 &= \mathbb{E}\|\mathbf{x}^\star - \text{Clip}(\tilde{\mathbf{x}}^{(t)} - \eta \tilde{\nabla}_{\tilde{\mathbf{x}}}^{(t)}\mathcal{L}(\tilde{\mathbf{x}}^{(t)}))\|_1\\
&=\mathbb{E}\|H\left((1-\frac{\eta d_2d_3}{2})^{t+1}(\mathbf{x}^\star - \tilde{\mathbf{x}}^{(0)})\right)\|_1,
\end{aligned} 
\end{equation}
where $H(\cdot) = \min(\mathbf{x}^\star-\mathbf{x}^{(0)} + \epsilon\mathbf 1,\max(\mathbf{x}^\star-\mathbf{x}^{(0)} - \epsilon\mathbf 1,\cdot))$. From Eq.~(\ref{norm5}) and Eq.~(\ref{norm6}), further we have
\begin{equation}
\label{norm7}
\begin{aligned}
\mathbb{E}\|\mathbf{x}^\star - \mathbf{x}^{(t+1)}\|_1 &=\mathbb{E}\|H\left((1-\frac{\eta d_2d_3}{2})^{t+1}(\mathbf{x}^\star - \mathbf{x}^{(0)}) + \sum\limits_{j=0}^{t}\frac{\eta d_2d_3}{2\pi}(1-\frac{\eta d_2d_3}{2})^{t-j}\mathbf{p}_{j}\right)\|_1\\
&=\sum\limits_{i=0}^{d}\mathbb{E}|H\left((1-\frac{\eta d_2d_3}{2})^{t+1}(\mathbf{x}^\star_i - \mathbf{x}^{(0)}_i) + \sum\limits_{j=0}^{t}\frac{\eta d_2d_3}{2\pi}(1-\frac{\eta d_2d_3}{2})^{t-j}\mathbf{p}_{ji}\right)_i|,
\end{aligned} 
\end{equation}
and
\begin{equation}
\label{norm8}
\begin{aligned}
\mathbb{E}\|\mathbf{x}^\star - \tilde{\mathbf{x}}^{(t+1)}\|_1 &=\mathbb{E}\|H\left((1-\frac{\eta d_2d_3}{2})^{t+1}(\mathbf{x}^\star - \tilde{\mathbf{x}}^{(0)})\right)\|_1 \\
&=\sum\limits_{i=0}^{d}\mathbb{E}|H\left((1-\frac{\eta d_2d_3}{2})^{t+1}(\mathbf{x}_i^\star - \tilde{\mathbf{x}}_i^{(0)})\right)_i|.
\end{aligned}
\end{equation}
Note that $\mathbf{x}^{(0)}$ = $\tilde{\mathbf{x}}^{(0)}$ in the theorem. If $|\mathbf{x}^\star_i - \mathbf{x}^{(0)}_i| < \epsilon$, then for $\forall t$, $|\mathbf{x}^\star_i - \mathbf{x}^{(0)}_i| < \epsilon$ and $|\mathbf{x}^\star_i - \tilde{\mathbf{x}}^{(0)}_i| < \epsilon$, where the $\text{Clip}$ function can be removed in this case. Eq.~(\ref{norm7}) and Eq.~(\ref{norm8}) can be formulated as
\begin{equation}
\label{norm9}
\begin{aligned}
\mathbb{E}\|\mathbf{x}^\star - \mathbf{x}^{(t+1)}\|_1  &=\sum\limits_{i=0}^{d}\mathbb{E}|(1-\frac{\eta d_2d_3}{2})^{t+1}(\mathbf{x}^\star_i - \mathbf{x}^{(0)}_i) + \sum\limits_{j=0}^{t}\frac{\eta d_2d_3}{2\pi}(1-\frac{\eta d_2d_3}{2})^{t-j}\mathbf{p}_{ji}|\\
&=\sum\limits_{i=0}^{d}\mathbb{E}\left((1-\frac{\eta d_2d_3}{2})^{t+1}(\mathbf{x}^\star_i - \mathbf{x}^{(0)}_i) + \sum\limits_{j=0}^{t}\frac{\eta d_2d_3}{2\pi}(1-\frac{\eta d_2d_3}{2})^{t-j}\mathbf{p}_{ji}\right)\text{sgn}(\mathbf{x}^\star_i - \mathbf{x}^{(0)}_i),
\end{aligned} 
\end{equation}
and
\begin{equation}
\label{norm10}
\begin{aligned}
\mathbb{E}\|\mathbf{x}^\star - \tilde{\mathbf{x}}^{(t+1)}\|_1  &=\sum\limits_{i=0}^{d}\mathbb{E}\left((1-\frac{\eta d_2d_3}{2})^{t+1}(\mathbf{x}_i^\star - \tilde{\mathbf{x}}_i^{(0)})\right)\text{sgn}(\mathbf{x}^\star_i - \mathbf{x}^{(0)}_i).
\end{aligned}
\end{equation}
If $|\mathbf{x}^\star_i - \mathbf{x}^{(0)}_i| \geq \epsilon$, the sign of $H(\cdot)_i$ is determined by the sign of $\mathbf{x}^\star_i - \mathbf{x}^{(0)}_i$. Therefore Eq.~(\ref{norm7}) and Eq.~(\ref{norm8}) can be formulated as
\begin{equation}
\label{norm9_2}
\begin{aligned}
\mathbb{E}\|\mathbf{x}^\star - \mathbf{x}^{(t+1)}\|_1  &=\sum\limits_{i=0}^{d}\mathbb{E}|(1-\frac{\eta d_2d_3}{2})^{t+1}(\mathbf{x}^\star_i - \mathbf{x}^{(0)}_i) + \sum\limits_{j=0}^{t}\frac{\eta d_2d_3}{2\pi}(1-\frac{\eta d_2d_3}{2})^{t-j}\mathbf{p}_{ji}|\\
&=\sum\limits_{i=0}^{d}\mathbb{E}H\left((1-\frac{\eta d_2d_3}{2})^{t+1}(\mathbf{x}^\star_i - \mathbf{x}^{(0)}_i) + \sum\limits_{j=0}^{t}\frac{\eta d_2d_3}{2\pi}(1-\frac{\eta d_2d_3}{2})^{t-j}\mathbf{p}_{ji}\right)_i\text{sgn}(\mathbf{x}^\star_i - \mathbf{x}^{(0)}_i),
\end{aligned} 
\end{equation}
and
\begin{equation}
\label{norm10_2}
\begin{aligned}
\mathbb{E}\|\mathbf{x}^\star - \tilde{\mathbf{x}}^{(t+1)}\|_1  &=\sum\limits_{i=0}^{d}\mathbb{E}H\left((1-\frac{\eta N}{2})^{t+1}(\mathbf{x}_i^\star - \tilde{\mathbf{x}}_i^{(0)})\right)_i\text{sgn}(\mathbf{x}^\star_i - \mathbf{x}^{(0)}_i).
\end{aligned}
\end{equation}
Recall the assumption that $\eta$ is sufficiently small, to be exact, it should satisfy the following constraints,
\begin{equation}
\label{eq:constraint_2_}
|\sum\limits_{j=0}^{m-1}\frac{\eta d_2d_3}{2\pi}(1-\frac{\eta d_2d_3}{2})^{m-1-j}\mathbf{p}_{ji}| < |(1-\frac{\eta d_2d_3}{2})^{m}(\mathbf{x}_i^{\star}-\mathbf{x}^{(0)}_i)|,
\end{equation}
for $m=1,\ldots,t$ and $i=1,\ldots,d$. Under the constraints, we find the sign of $\mathbf{x}_i^{\star}-\mathbf{x}^{(0)}_i$ determine the sign of $\left((1-\frac{\eta d_2d_3}{2})^{t-j}\mathbf{p}_{ji}\right)_i$. Therefore, when we compare Eq.~(\ref{norm9}) and Eq.~(\ref{norm10}), Eq.~(\ref{norm9_2}) and Eq.~(\ref{norm10_2}), we actually compare the sign of $\mathbb{E}(\mathbf{p}_{ji}\text{sgn}(\mathbf{w}^\star_i-\mathbf{w}^{(0)}_i)$. We first solve the expectation of $\textbf{x}^{(t)}$. There is
\begin{equation}
\label{e_x}
\begin{aligned}
\mathbb{E}\mathbf{x}^{(t)}&=\mathbb{E}\left((1-\frac{\eta d_2d_3}{2})\textbf{x}^{(t-1)}+\frac{\eta d_2d_3}{2}\mathbf{x}^\star-\frac{\eta d_2d_3}{2\pi}\textbf{p}_t\right)\\
&=\mathbb{E}\left((1-\frac{\eta d_2d_3}{2})^{t}\textbf{x}^{(0)}+(1-(1-\frac{\eta d_2d_3}{2})^{t})\mathbf{x}^\star-\frac{\eta d_2d_3}{2\pi}\sum_{j=0}^{t-1}\left((1-\frac{\eta d_2d_3}{2})^{(t-1-j)}\textbf{p}_j\right)\right).
\end{aligned}
\end{equation}
We then solve the following equation,
\begin{equation}
\label{T1_sign}
\begin{aligned}
\mathbb{E}\left(\mathbf{p}_{si}\text{sgn}(\mathbf{x}^\star_i-\mathbf{x}^{(0)}_i)\right)=&\mathbb{E}\left((\theta_s\mathbf{x}^\star_i-\frac{\|\mathbf{x}^\star\|}{\|\mathbf{x}^{(s)}\|}\sin\theta_s\mathbf{x}^{(s)}_i)\text{sgn}(\mathbf{x}^\star_i-\mathbf{x}^{(0)}_i)\right)\\
=&\mathbb{E}\left(\left(\theta_s-\frac{\|\mathbf{x}^\star\|}{\|\mathbf{x}^{(s)}\|}\sin\theta_s(1-(1-\frac{\eta d_2d_3}{2})^{s})\right)\mathbf{x}^\star_i-\frac{\|\mathbf{x}^\star\|}{\|\mathbf{x}^{(s)}\|}\sin\theta_s(1-\frac{\eta d_2d_3}{2})^{s}\mathbf{x}^{(0)}_i\right)\text{sgn}(\mathbf{x}^\star_i-\mathbf{x}^{(0)}_i)\\
&+\mathbb{E}\left(\frac{\eta d_2d_3}{2\pi}\sum_{j=0}^{s-1}\left((1-\frac{\eta d_2d_3}{2})^{(s-1-j)}\textbf{p}_{ji}\right)\right)\text{sgn}(\mathbf{x}^\star_i-\mathbf{x}^{(0)}_i).
\end{aligned}
\end{equation}
From Eq.~(\ref{eq:constraint_2_}), we have $\theta_s-\frac{\|\mathbf{x}^\star\|}{\|\mathbf{x}^{(s)}\|}\sin\theta_s>0$. Recall that $\mathbf{x}^\star \sim N(\mu_1,\sigma_1^2)$ and $\mathbf{x}^{(0)} \sim N(0,\sigma_2^2)$, using the Lemma 2, we first have
\begin{equation}
\mathbb{E}\left(\mathbf{p}_{0i}\text{sgn}(\mathbf{x}^\star_i-\mathbf{x}^{(0)}_i)\right)= \mathbb{E}\left(\theta_j\mathbf{x}^\star-\frac{\|\mathbf{x}^\star\|}{\|\mathbf{x}^{(0)}\|}\sin\theta_j\mathbf{x}^{(0)}\right)\text{sgn}(\mathbf{x}^\star_i-\mathbf{x}^{(0)}_i) > 0.
\end{equation}
Using the mathematical induction, we have
\begin{equation}
\label{eq:p>0}
\mathbb{E}\left(\mathbf{p}_{ji}\text{sgn}(\mathbf{w}^\star_i-\mathbf{w}^{(0)}_i)\right)>0.
\end{equation}
With Eq.~(\ref{eq:p>0}), we find Eq.~(\ref{norm9}) and Eq.~(\ref{norm9_2}) are greater than Eq.~(\ref{norm10}) and Eq.~(\ref{norm10_2}), respectively. That is to say, we have $$\mathbb{E}\|\mathbf{x}^\star - \tilde{\mathbf{x}}^{(t+1)}\|_1\leq \mathbb{E}\|\mathbf{x}^\star - \mathbf{x}^{(t+1)}\|_1.$$

\section{Proof of Theorem 2}
The one-layer network is formulated as
\begin{equation}
\label{ori}
h(\mathbf{x},\mathbf{w}) = \sigma(\mathbf{x}^T\mathbf{w}),
\end{equation}
where $\mathbf{x}\in\mathbb{R}^d$ is the input vector, $\mathbf{w}\in\mathbb{R}^d$ is the weight vector, and $\sigma(\cdot)$ is the ReLU function. Therefore, $h(\mathbf X, \mathbf w)=\sigma(\mathbf X\mathbf w)$ where $\mathbf{X} = [\mathbf{x}_1^T;...;\mathbf{x}_N^T]$ is the input data matrix and $\mathbf{x}_k\in\mathbb{R}^d$ is the $k$-th training instance, for $k=1,...,N$. We assume the teacher network has the optimal weight, and the loss function can be formulated as 
\begin{equation}
\label{loss}
\mathcal{L}(\mathbf{w}) = \frac{1}{2}\|h(\mathbf{X},\mathbf{w})-h(\mathbf{X},\mathbf{w}^\star)\|_2^2,
\end{equation}
where $\mathbf{w}$ and $\mathbf{w}^\star$ are the weight vector for the student network and teacher network, respectively. While training the simple one-layer network with SGD, The update rule for training one-layer network with SGD can be formulated as 
\begin{equation}
\label{update}
\mathbf{w}^{(t+1)}=\mathbf{w}^{(t)} - \eta \nabla_{\mathbf{w^{(t)}}}\mathcal{L}(\mathbf{w^{(t)}}),
\end{equation}
where we use the gradients of standard BP and LinBP, \emph{i.e.}, $\nabla_{\mathbf{w}}\mathcal{L}(\mathbf{w})$ and $\tilde{\nabla}_{\mathbf{w}}\mathcal{L}(\mathbf{w})$, to obtain $\{\mathbf{w}^{(t)}\}$ and $\{\tilde{\mathbf{w}}^{(t)}\}$, respectively.
\begin{Theorem}
	For the one-layer teacher-student network formulated as Eq.~(\ref{ori}), the training task sets Eq.~(\ref{loss}) and Eq.~(\ref{update}) as the loss function and the update rule, respectively.
	Assume that $\mathbf{X}$ is generated from standard Gaussian distribution, $\mathbf{w}^\star \sim N(\mu_1,\sigma_1^2)$, $\mathbf{w}^{(0)} \sim N(0,\sigma_2^2)$, and $\eta$ is reasonably small\footnote{See Eq.~(\ref{eq:constraint_1_}) for a precious formulation of the constraint.}. 
	Let $\mathbf{w}^{(t)}$ and $\tilde{\mathbf{w}}^{(t)}$ be the weight vectors obtained in the $t$-th iteration of training using standard BP and LinBP respectively. Then we have $$\mathbb{E}\|\mathbf{w}^\star - \tilde{\mathbf{w}}^{(t)}\|_1 \leq \mathbb{E}\|\mathbf{w}^\star - \mathbf{w}^{(t)}\|_1.$$
\end{Theorem}
\textbf{Proof.} We first recall the partial gradient to $\mathbf{W}$ for BP and LinBP, which are formulated as
\begin{equation}
\label{gradient_proof}
\nabla_{\mathbf{w}}\mathcal{L}(\mathbf{w}) = \mathbf{X}^T\mathbf{D}(\mathbf{X},\mathbf{w})(\mathbf{D}(\mathbf{X},\mathbf{w})\mathbf{Xw} - \mathbf{D}(\mathbf{X},\mathbf{w}^\star) \mathbf{X} \mathbf{w}^{\star}),
\end{equation}
and
\begin{equation}
\label{gradient2_proof}
\tilde{\nabla}_{\mathbf{w}}\mathcal{L}(\mathbf{w}) = \mathbf{X}^T(\mathbf{D}(\mathbf{X},\mathbf{w})\mathbf{Xw} - \mathbf{D}(\mathbf{X},\mathbf{w}^\star) \mathbf{X} \mathbf{w}^{\star}),
\end{equation}
respectively. From Theorem 1 in \cite{tian2017analytical}, we can calculate the expectation of Eq.~(\ref{gradient_proof}) and Eq.~(\ref{gradient2_proof}) as
\begin{equation}
\label{ex1}
\begin{aligned}
\mathbb{E}[\nabla_{\mathbf{w}}\mathcal{L}(\mathbf{w})] &=\frac{N}{2}(\mathbf{w}-\mathbf{w}^\star)+\frac{N}{2\pi}\left(\theta \mathbf{w}^\star-\frac{\|\mathbf{w}^\star\|}{\|\mathbf{w}\|}\sin\theta \mathbf{w}\right),
\end{aligned}
\end{equation}
and
\begin{equation}
\label{ex2}
\mathbb{E}[\tilde{\nabla}_{\mathbf{w}}\mathcal{L}(\mathbf{w})] =\frac{N}{2}(\mathbf{w}-\mathbf{w}^\star),
\end{equation}
where $\theta \in [0,\pi]$ is the angle between $\mathbf{w}$ and $\mathbf{w^\star}$.
The expectation of $l_1$ distance between $\mathbf{w}^{(t+1)}$ ($\tilde{\mathbf{w}}^{(t+1)}$) and $\textbf{w}^\star$ can be formulated as
\begin{equation}
\label{norm1}
\begin{aligned}
\mathbb{E}\|\mathbf{w}^\star - \mathbf{w}^{(t+1)}\|_1 &= \mathbb{E}\|\mathbf{w}^\star - \mathbf{w}^{(t)} + \eta \nabla_{\mathbf{w}^{(t)}}\mathcal{L}(\mathbf{w}^{(t)})\|_1\\
&=\mathbb{E}\|(1-\frac{\eta N}{2})(\mathbf{w}^\star - \mathbf{w}^{(t)}) + \frac{\eta N}{2\pi}\left(\theta \mathbf{w}^\star-\frac{\|\mathbf{w}^\star\|}{\|\mathbf{w}^{(t)}\|}\sin\theta \mathbf{w}^{(t)}\right)\|_1,
\end{aligned} 
\end{equation}
and 
\begin{equation}
\label{norm2}
\begin{aligned}
\mathbb{E}\|\mathbf{w}^\star - \tilde{\mathbf{w}}^{(t+1)}\|_1 &= \mathbb{E}\|\mathbf{w}^\star - \tilde{\mathbf{w}}^{(t)} + \eta \tilde{\nabla}_{\mathbf{w}^{(t)}}\mathcal{L}(\mathbf{w}^{(t)})\|_1\\
&=\mathbb{E}\|(1-\frac{\eta N}{2})(\mathbf{w}^\star - \tilde{\mathbf{w}}^{(t)})\|_1.
\end{aligned}
\end{equation}
We assume $\mathbf{q}_j=\theta_j\mathbf{w}^\star-\frac{\|\mathbf{w}^\star\|}{\|\mathbf{w}^{(j)}\|}\sin\theta_j\mathbf{w}^{(j)}$. Therefore, Eq.~(\ref{norm1}) can be formulated as
\begin{equation}
\label{norm1_1}
\begin{aligned}
\mathbb{E}\|\mathbf{w}^\star - \mathbf{w}^{(t+1)}\|_1 &=\mathbb{E}\|(1-\frac{\eta N}{2})(\mathbf{w}^\star - \mathbf{w}^{(t)}) + \frac{\eta N}{2\pi}\mathbf{q}_t\|_1\\
&= \mathbb{E}\|(1-\frac{\eta N}{2})^2(\mathbf{w}^\star - \mathbf{w}^{(t-1)}) + (1-\frac{\eta N}{2})\frac{\eta N}{2\pi}\mathbf{q}_{t-1} + \frac{\eta N}{2\pi}\mathbf{q}_t\|_1\\
&= \mathbb{E}\|(1-\frac{\eta N}{2})^{t+1}(\mathbf{w}^\star - \mathbf{w}^{(0)}) + \sum\limits_{j=0}^{t}\frac{\eta N}{2\pi}(1-\frac{\eta N}{2})^{t-j}\mathbf{q}_{j}\|_1.
\end{aligned}
\end{equation}
And Eq.~(\ref{norm2}) can be formulated as
\begin{equation}
\label{norm2_1}
\begin{aligned}
\mathbb{E}\|\mathbf{w}^\star - \tilde{\mathbf{w}}^{(t+1)}\|_1 &=\mathbb{E}\|(1-\frac{\eta N}{2})(\mathbf{w}^\star - \tilde{\mathbf{w}}^{(t)})\|_1\\
&= \mathbb{E}\|(1-\frac{\eta N}{2})^{t+1}(\mathbf{w}^\star - \tilde{\mathbf{w}}^{(0)})\|_1.
\end{aligned}
\end{equation}
Note that $\tilde{\mathbf{w}}^{(0)}= \mathbf{w}^{(0)}$. As mentioned in Theorem 1, we assume $\eta$ is sufficiently small, to be exact, it should satisfy the following constraints,
\begin{equation}
\label{eq:constraint_1_}
|\sum\limits_{j=0}^{m}\frac{\eta N}{2\pi}(1-\frac{\eta N}{2})^{m-j}\mathbf{q}_{ji}| < |(1-\frac{\eta N}{2})^{m+1}(\mathbf{w}_i^{\star}-\mathbf{w}^{(0)}_i)|,
\end{equation}
for $m=1,\ldots,t$ and $i=1,\ldots,d$. Under the constraints, $\mathbf{w}^\star_i-\mathbf{w}^{(0)}_i$ determines the sign of $\mathbf{w}^\star_i-\mathbf{w}^{(t+1)}_i$ in Eq.~(\ref{norm1_1}), then Eq.~(\ref{norm1_1}) and Eq.~(\ref{norm2_1}) can be calculated as
\begin{equation}
\label{norm3_1}
\begin{aligned}
&\mathbb{E}\|\mathbf{w}^\star - \mathbf{w}^{(t+1)}\|_1 \\
&= \sum\limits_{i=0}^{d}\mathbb{E}\left((1-\frac{\eta N}{2})^{t+1}(\mathbf{w}^\star_i - \mathbf{w}_i^{(0)}) + \sum\limits_{j=0}^{t}\frac{\eta N}{2\pi}(1-\frac{\eta N}{2})^{t-j}\mathbf{q}_{ji}\right)\text{sgn}(\mathbf{w}^\star_i-\mathbf{w}^{(0)}_i)\\
&=\sum\limits_{i=0}^{d}\mathbb{E}\left((1-\frac{\eta N}{2})^{t+1}|\mathbf{w}^\star_i - \mathbf{w}_i^{(0)}| + \sum\limits_{j=0}^{t}\frac{\eta N}{2\pi}(1-\frac{\eta N}{2})^{t-j}\mathbf{q}_{ji}\text{sgn}(\mathbf{w}^\star_i-\mathbf{w}^{(0)}_i)\right),
\end{aligned}
\end{equation}
and
\begin{equation}
\label{norm3_2}
\begin{aligned}
\mathbb{E}\|\mathbf{w}^\star - \tilde{\mathbf{w}}^{(t+1)}\|_1 &= \sum\limits_{i=0}^{d}\mathbb{E}\left((1-\frac{\eta N}{2})^{t+1}(\mathbf{w}^\star_i - \tilde{\mathbf{w}}^{(0)}_i)\text{sgn}(\mathbf{w}^\star_i-\tilde{\mathbf{w}}^{(0)}_i)\right)\\
&=\sum\limits_{i=0}^{d}\mathbb{E}\left((1-\frac{\eta N}{2})^{t+1}(\mathbf{w}^\star_i - \mathbf{w}^{(0)}_i)\text{sgn}(\mathbf{w}^\star_i-\mathbf{w}^{(0)}_i)\right)\\
&=\sum\limits_{i=0}^{d}\mathbb{E}\left((1-\frac{\eta N}{2})^{t+1}|\mathbf{w}^\star_i - \mathbf{w}^{(0)}_i|\right).
\end{aligned}
\end{equation}
Similar to Theorem 2, using Lemma 2, we have
\begin{equation}
\label{eq:q>0}
\mathbb{E}\left(\mathbf{q}_{ji}\text{sgn}(\mathbf{w}^\star_i-\mathbf{w}^{(0)}_i)\right)>0.
\end{equation}
With Eq.~(\ref{eq:q>0}), we find Eq.~(\ref{norm3_1}) is greater than Eq.~(\ref{norm3_2}), i.e.,
$$\mathbb{E}\|\mathbf{w}^\star - \tilde{\mathbf{w}}^{(t+1)}\|_1 \leq \mathbb{E}\|\mathbf{w}^\star - \mathbf{w}^{(t+1)}\|_1.$$


\ifCLASSOPTIONcaptionsoff
  \newpage
\fi

\bibliographystyle{IEEEtran}
\bibliography{reference}